\documentclass{article}

\PassOptionsToPackage{numbers, compress}{natbib}
\usepackage[main,final]{neurips_2026}

\usepackage[utf8]{inputenc} 
\usepackage[T1]{fontenc}    
\usepackage{hyperref}       
\usepackage{url}            
\usepackage{booktabs}       
\usepackage{amsfonts}       
\usepackage{nicefrac}       
\usepackage{microtype}      
\usepackage{xcolor}         

\usepackage{amsmath}
\usepackage{amssymb}
\usepackage{mathtools}
\usepackage{amsthm}

\usepackage{caption}
\usepackage{subcaption}

\usepackage{xcolor}
\usepackage{mdframed}
\usepackage{fancyvrb}

\usepackage{algorithm}
\usepackage{algpseudocode}

\usepackage{array, colortbl}
\usepackage{tikz}
\usepackage{makecell} 

\usepackage{enumitem}

\usepackage{wrapfig}

\usepackage[capitalize,noabbrev]{cleveref}

\theoremstyle{plain}
\newtheorem{theorem}{Theorem}[section]

\newtheorem{lemma}[theorem]{Lemma}
\newtheorem{corollary}[theorem]{Corollary}
\theoremstyle{definition}

\newtheorem{assumption}[theorem]{Assumption}
\theoremstyle{remark}

\surroundwithmdframed[backgroundcolor=gray!10,linecolor=gray!70]{theorem}
\surroundwithmdframed[backgroundcolor=gray!10,linecolor=gray!70]{lemma}
\surroundwithmdframed[backgroundcolor=gray!10,linecolor=gray!70]{corollary}
\surroundwithmdframed[backgroundcolor=gray!10,linecolor=gray!70]{remark}

\usepackage{amsmath,amsfonts,bm}

\def\eqref#1{equation~\ref{#1}}

\def\1{\bm{1}}

\DeclareMathAlphabet{\mathsfit}{\encodingdefault}{\sfdefault}{m}{sl}
\SetMathAlphabet{\mathsfit}{bold}{\encodingdefault}{\sfdefault}{bx}{n}

\def\gA{{\mathcal{A}}}

\def\gC{{\mathcal{C}}}

\def\gF{{\mathcal{F}}}

\def\gM{{\mathcal{M}}}
\def\gN{{\mathcal{N}}}

\def\sR{{\mathbb{R}}}

\newcommand{\E}{\mathbb{E}}

\newcommand{\PMpred}{V^{(P)}}
\newcommand{\TMpred}{V^{(T)}}
\newcommand{\PMcontrol}{Q^{(P)}}
\newcommand{\TMcontrol}{Q^{(T)}}
\newcommand{\vpt}{V^{(PT)}}
\newcommand{\QPT}{Q^{(PT)}}
\newcommand{\PMlr}{\overline{\alpha}}
\newcommand{\TMlr}{\alpha}
\newcommand{\PMfeat}{\phi}
\newcommand{\TMfeat}{z}
\newcommand{\PMFeatMatrix}{\Phi}
\newcommand{\TMFeatMatrix}{Z}

\newcommand{\p}{\mathcal{P}}
\newcommand{\s}{\mathcal{S}}
\newcommand{\m}{\mathcal{M}}
\newcommand{\rew}{\mathcal{R}}
\newcommand{\act}{\mathcal{A}}
\newcommand{\pol}{\pi}

\newcommand{\w}{{\bf w}}

\newcommand{\PMparams}{\theta}
\newcommand{\TMparams}{\w}

\newcommand{\Ep}{\E_{\pol}}
\newcommand{\keyA}{\mathbf{A}}
\newcommand{\keyC}{\mathbf{C}}
\newcommand{\gb}{b}
\newcommand{\keyb}{\mathbf{b}}
\newcommand{\dis}{d_\pol}
\newcommand{\task}{\tau}
\newcommand{\ptask}{\p_{\task}}
\newcommand{\ppoltask}{\p_{\pol, \task}}
\newcommand{\rpoltask}{\rew_{\pol, \task}}

\newcommand{\Dtask}{D_{\task}}
\newcommand{\Edis}{\E_{\dis}}
\newcommand{\Etask}{\E_{\task}}
\newcommand{\norm}[1]{\left\lVert#1\right\rVert}

\definecolor{commentblue}{RGB}{60, 100, 180} 

\definecolor{Hdr}{RGB}{210,210,210}   
\definecolor{Alt}{RGB}{236,241,246}   

\newcommand{\reddot}{\tikz[baseline=-0.6ex]\fill[red] (0,0) circle (4pt);}
\newcommand{\bluedot}{\tikz[baseline=-0.6ex]\fill[blue] (0,0) circle (4pt);}
\title{Permanent and Transient Representations for Continual Reinforcement Learning}

\author{%
 Nishanth Anand \thanks{corresponding author.} \\
  School of Computer Science\\
  McGill University and Mila\\
  \texttt{nishanth.anand@mail.mcgill.ca} \\
\And
  Doina Precup \\
  School of Computer Science \\
  McGill University, Mila, and Google DeepMind \\
  \texttt{dprecup@cs.mcgill.ca} \\
}

\begin{document}

\maketitle

\begin{abstract}
Continual Reinforcement Learning (CRL) agents struggle to adapt to new situations while retaining prior knowledge, leading to a stability–plasticity trade-off. One approach to solving this problem is by using complementary learning systems: one for slow, long-term learning and another for quick, transient adaptation. Anand \& Precup (2023) instantiated this idea using permanent and transient value functions, but used the same features in both systems. Intuitively, long-term learning should rely more on parametric representations for broad generalization, while transient learning would be best served by non-parametric representations for situation-specific adaptation. In this paper, we explore this idea both theoretically and empirically. By introducing a novel non-parametric approximation for estimating the transient value function in complex tasks, we demonstrate that our method enables online learning and outperforms competitive baselines on image-based tasks and Craftax-Classic, underscoring the effectiveness of our system-level decomposition.
\end{abstract}

\section{Introduction}
\label{sec:introduction}

Continual reinforcement learning (CRL) is a key ingredient underlying general intelligence~\citep{sutton2022alberta, silver2025welcome}. An important challenge in CRL is the tension between retaining knowledge already acquired and adapting to new information — called the \emph{stability–plasticity} trade-off~\citep{carpenter1987massively}. In contrast, humans and other biological agents adapt to environmental changes throughout their lifetimes. \citet{kumaran2016learning} posited that this natural ability is due to the existence of two complementary learning systems (CLS): one that adapts rapidly and another that slowly consolidates information across experiences.
\citet{anand2023prediction} instantiated this architecture in the context of reinforcement learning (RL) agents by decomposing the value function into a permanent component that provides a stable baseline estimate for any situation the agent may face and a transient component that adapts these estimates to the current context. Both components update using temporal-difference-style corrections, working in tandem. However, in their work, both value functions share the same sets of features. Intuitively, using separate feature spaces for these approximators would better align with the idea of keeping them complementary: one system should compensate for the other's weaknesses, making the framework both \textit{more effective for CRL and more biologically plausible}. Permanent features should encode \emph{stationary or slowly changing} components of the environment, such as the map of a city, while transient features can capture \emph{situation-specific} aspects, such as a road being blocked on a given day. Thus, permanent representations should aid in generalization across various situations, while transient representations should intuitively facilitate local adjustments needed for rapid adaptation. These transient updates must also support temporary adaptation by correcting the permanent behaviour based on the context, and seamlessly reverting to it when the situation changes (e.g., a temporary closure and subsequent reopening of a regular office route). This intuition suggests that the permanent function approximator should be parametric, while the transient one may need to complement it.

In this paper, we develop and validate this intuition, both theoretically and empirically. An important goal of our work is to scale this type of CRL agent to large environments. For this purpose, we introduce a novel non-parametric approximator that operates directly on raw observations and enables rapid, precise corrections, facilitating online learning and controlled generalization.

Our main contributions are as follows:
\begin{itemize}
\item We extend the permanent–transient value function framework to allow separate feature representations in the two approximators, and provide convergence guarantees under linear function approximation, using a two-timescale convergence technique;
\item We introduce a novel non-parametric method for estimating the transient value function, which can be used to complement neural network-based permanent features, mirroring the Dentate Gyrus's functionality to encode specific information without catastrophic interference rapidly~\citep{berron2016strong, severa2017combinatorial};
\item We evaluate our approach on small illustrative examples, image-based tasks, and the $18$M and $250$M Craftax tasks~\citep{matthews2024craftax}, showing favourable comparisons against other baselines, including those that perform stability-plasticity at the parameter level.
\end{itemize}

\section{Background}
\label{sec:background}

\textbf{Continual RL.} CRL agents must exhibit \emph{endless adaptation} (also called plasticity~\citep{abel2025plasticity}), instead of converging to a fixed solution~\citep{abel2023definition, sutton2007role}. The need for CRL arises when aspects of the environment, such as rewards or transition dynamics, change over time~\citep{khetarpal2022towards, pan2025survey}, or when the agent's resources are limited relative to the complexity of its environment, thereby forcing the agent to keep updating its knowledge~\citep{kumar2025continual, javed2024big, lewandowski2025world}. In such scenarios, the agent must balance retaining useful information from the past to adapt more quickly when similar situations reappear—\emph{stability}—while allocating resources to learn from new experiences—\emph{plasticity}.

\textbf{Mechanistic approaches.} Because neural networks have become the main approach to approximating value functions and policies~\citep{mnih2013playing}, most CRL research has focused on understanding the stability–plasticity trade-off in neural networks~\citep{lyle2022understanding, nikishin2022primacy, abbas2023loss, lyle2023understanding, lyle2024disentangling} and on developing new regularization techniques~\citep{lu2019adaptive, lewandowski2024learning, chung2024parseval} and optimizers~\citep{kirkpatrick2017overcoming, ash2020warm, d2022sample, jones2022learning, elsayed2023utility, schwarzer2023bigger, dohare2024loss, han2026fire} to improve it. A detailed overview on this topic can be found in~\cite{klein2024plasticity}.

\textbf{System-level approaches.} Building on earlier ideas of decomposing value functions in model-based RL~\citep{silver2008sample}, \citet{anand2023prediction} proposed decomposing both the value function and the action-value function into two components to trade-off stability and plasticity: permanent components, $\PMpred$ and $\PMcontrol$, which learn general estimates from the entire agent experience (similar to~\cite{abel2018policy} in transfer learning), and transient components, $\TMpred$ and $\TMcontrol$, which adapt these estimates to the current situation by computing residuals. The overall value functions, $\vpt$ and $\QPT$, are then expressed as sums of these two components:
\begin{align}
    \label{Eq:prediction-pt-definition}
    \vpt(s) &= \TMpred_\TMparams(s) + \PMpred_\PMparams(s), \\
    \QPT(s, a) &= \TMcontrol_\TMparams(s, a) + \PMcontrol_\PMparams(s, a),
\end{align}
where $\PMparams$ and $\TMparams$ are the parameters of the permanent and transient function approximators, respectively.

The permanent value function is updated more slowly, either every $k$ timesteps or at task boundaries (when available), by using experience samples from that phase:
\begin{small}
\begin{align}
    \label{Eq:permanent-update}
    \PMparams_{k+1} \gets \PMparams_k + \quad \PMlr_k \left(\vpt(S_k) - \PMpred_\PMparams (S_k) \right) \nabla_{\PMparams} \PMpred_{\PMparams}(S_k),
\end{align}
\begin{align}
    \label{Eq:permanent-update-ctrl}
    \PMparams_{k+1} \gets \PMparams_k + \ \PMlr_k \left(\QPT(S_k, A_k) - \PMcontrol_\PMparams(S_k, A_k)\right) \nabla_{\PMparams}\PMcontrol_{\PMparams}(S_k, A_k),
\end{align}
\end{small}
where $\PMlr$ is the learning rate for permanent updates.
In contrast, the transient value function updates rapidly to capture aspects of the value that are not yet reflected in the permanent estimates:
\begin{align}
    \label{Eq:transient-update}
    \TMparams_{t+1} \gets \TMparams_t + \TMlr_t \delta_t^V \nabla_{\TMparams} \TMpred_{\TMparams}(S_t), \\
    \TMparams_{t+1} \gets \TMparams_t + \TMlr_t \delta_t^Q \nabla_{\TMparams} \TMcontrol_{\TMparams}(S_t, A_t),
\end{align}
where $\TMlr$ is the learning rate for transient updates, $\delta_t^V = R_{t+1} + \gamma \vpt(S_{t+1}) - \vpt(S_t)$ and $\delta_t^Q = R_{t+1} + \gamma \max_{a'} \QPT(S_{t+1}, a') - \QPT(S_t, A_t)$ are TD errors. To maintain plasticity, transient parameters are decayed or reset after each permanent update.

Along similar lines, neural episodic control (NEC) approaches decompose overall learning into slow representation learning via a neural network and rapid value learning via a differentiable dictionary~\citep{pritzel2017neural,gershman2017reinforcement}. However, their applicability to continual RL, including other fast and slow approaches~\citep{botvinick2019reinforcement}, is limited by the complexities of tuning and constantly changing representations.

\section{Permanent and Transient Representations}
\label{sec:pt-rep-definition}

Since the permanent and transient value functions should complement each other, it is reasonable to assume they use distinct representations with different requirements. In CLS~\citep{kumaran2016learning}, the equivalent of the transient representation is considered to be the hippocampus, which stores specific, individual episodes or snapshots of latent embeddings of recent experience. These are later replayed during sleep, resulting in consolidation and slow learning of a permanent representation, thought to be located mainly in the prefrontal cortex. This representation provides good generalization to new situations and supports long-term planning.

In the rest of the paper, we consider architectural choices which implement this intuition for CRL. Specifically, the permanent representation should be expressive enough to learn baseline predictions for any situation the agent might encounter. Learning these representations can be slow, but the information acquired should persist and be useful for a long time (i.e., high stability). Therefore, the permanent representation should support the broad generalization of predictions between similar situations. Many \emph{neural network architectures}, such as feedforward, convolutional, or recurrent, meet these desiderata, making them good candidates for the permanent value function.

The transient representation should support rapid online learning (i.e., high plasticity) to adapt quickly to new situations. In certain cases, e.g. high-speed corners in an F1 circuit, they should facilitate learning precise estimates with minimal or carefully controlled generalization around the current experience. Moreover, the transient representations should enable fast, accurate information retrieval. As in CLS, knowledge stored in the transient representation should support long-term learning of the permanent representation.

\section{Illustrations}
\label{sec:small-scale-experiments}

In this section, we illustrate this intuition on simple prediction and control problems, using linear function approximation. In these experiments, we assume that the agent’s experience can be divided into tasks and that the task boundaries are known to the agent (semi-CRL). We make this assumption to separately analyze the permanent component’s ability for long-term retention and the transient component’s ability to adapt to non-stationarity, but we relax it in the large-scale experiments. In these tasks, the transition dynamics remain fixed, while the reward function changes. These experiments show that using separate features to approximate permanent and transient value functions yields faster adaptation and better retention of past values in CRL. The pseudocode is provided in Appendices~\ref{algo:PT-linear-prediction} and~\ref{algo:PT-linear-control}, and the details of the hyperparameter sweeps are in Appendix~\ref{app-sec:small-scale-experiments}.

\subsection{Prediction}
\label{subsec:prediction}

We consider 1D chain and 2D grid navigation environments for the prediction experiment.

\paragraph{1D chain task.} In the 1D chain task, the agent starts in $S_0$ and deterministically transitions to the next successive state at each timestep ($S_{i} \to S_{i+1}$) until termination. Rewards are constructed such that the true value of a state in task $\tau$, $V_{\tau}(s)$, is exactly $\phi_1(s) + \w_{\tau} \phi_2(s)$, where $\phi_1(s)$ and $\phi_2(s)$ are the first- and third-order Fourier features~\citep{konidaris2011value} for state $s$. The variable $\w_{\tau}$ is an autoregressive weight that induces non-stationarity across tasks, which is updated according to $\w_{\tau_{k+1}} = 0.25 * \w_{\tau_{k}} + X$, where $X \sim \mathcal{N}(0, 2.0)$.  We run $750$ episodes in total, sampling a new task every $75$ episodes. We set $\gamma=0.9$, set the number of states to 20, and report root mean squared value error (RMSVE) on online and other tasks, computed across $30$ seeds with $90\%$ confidence intervals using $z^{\ast}=1.645$.

We compare three variants of TD learning and two variants of PT-TD learning based on the features they use to approximate the value function. In the TD learning variants, we use either first-order, third-order, or both orders of features. For PT-TD learning, we use both feature orders to estimate the permanent and transient components in the first variant, and only first-order features to estimate the permanent component in the second variant (the transient component uses both feature orders).

\begin{figure}[htb]
    \centering
    \begin{subfigure}[b]{0.48\columnwidth}
        \centering
        \includegraphics[width=\textwidth]{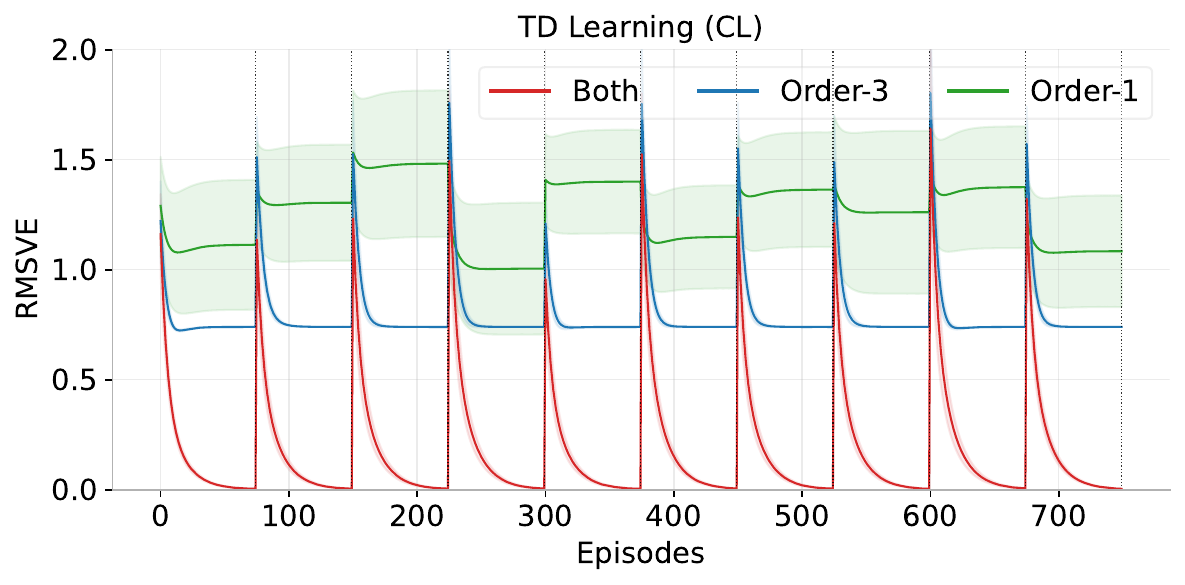}
        \caption{TD Learning.}
        \label{fig:td-1d-crl}
    \end{subfigure}
    \hfill
    \begin{subfigure}[b]{0.48\columnwidth}
        \centering
        \includegraphics[width=\textwidth]{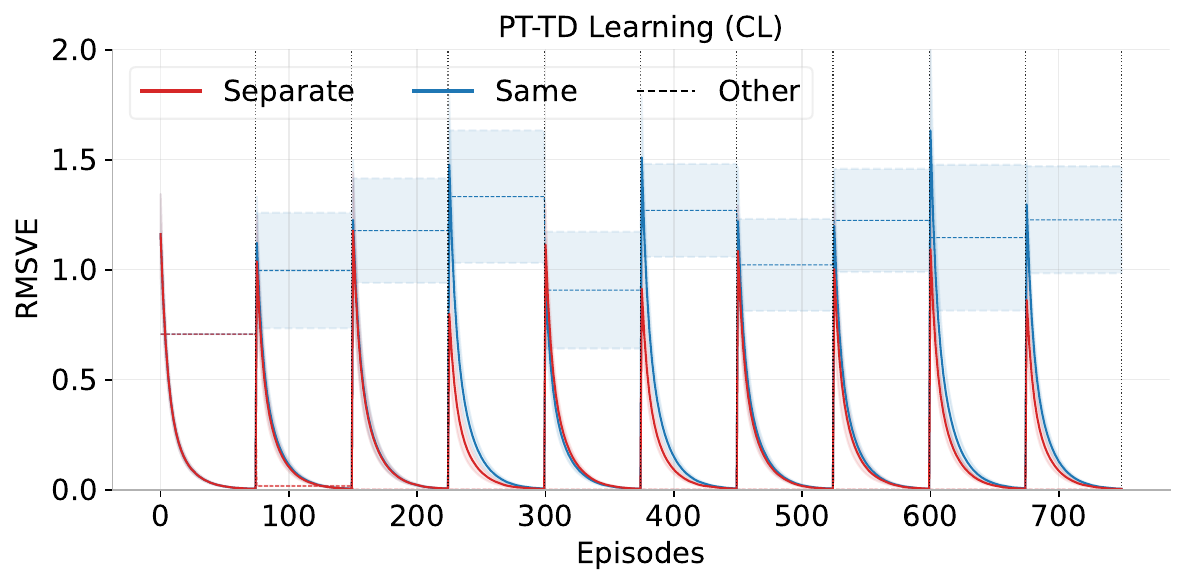}
        \caption{PT-TD Learning.}
        \label{fig:pt-td-1d-crl}
    \end{subfigure}
    \caption{\textbf{Results in 1D chain environment.} (a) TD learning using both features results in the lowest RMSVE. (b) The separate variant of PT-TD learning has a lower initial RMSVE following task changes, and its permanent component converges to the correct value after just one task.}
    \label{fig:1d-results}
\end{figure}

\textbf{Results:} The results are presented in Figure~\ref{fig:1d-results}. Because the true value function comprises both feature components, approximating it with only one component limits representational capacity and causes large prediction errors~\citep{tsitsiklis1996analysis}, as shown in Figure~\ref{fig:td-1d-crl}. \textit{Therefore, both features are necessary to estimate the value function accurately.}

In PT-TD, the permanent estimates of the separate variant converge quickly, yielding near-zero RMSVE on the other tasks (dotted line), as shown in Figure~\ref{fig:pt-td-1d-crl}. In contrast, the full variant continues to oscillate. Because the permanent weights take a step towards the latest estimate of the value function at the end of each task,\textit{ the full variant has the representational capacity to capture specifics of the recently ended task. Therefore, it suffers from high variance.} The separate variant prevents the permanent component from learning task-specific variations, so it remains highly stable and provides a good baseline estimate for future tasks, resulting in lower initial RMSVE on the online task, validating the need for separate features.

\begin{figure}[htb]
    \centering
    \begin{subfigure}[b]{0.23\columnwidth}
        \centering
        \includegraphics[height=2.2cm]{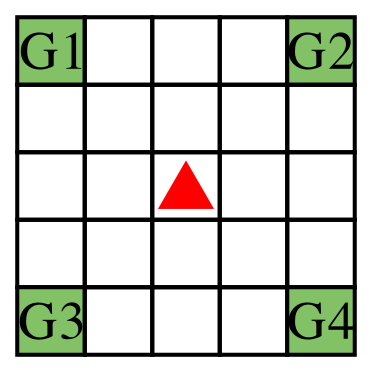}
    \caption{Grid Task.}
    \label{fig:prediction-grid}
    \end{subfigure}
    \hfill
    \begin{subfigure}[b]{0.23\columnwidth}
        \centering
        \includegraphics[height=2.2cm]{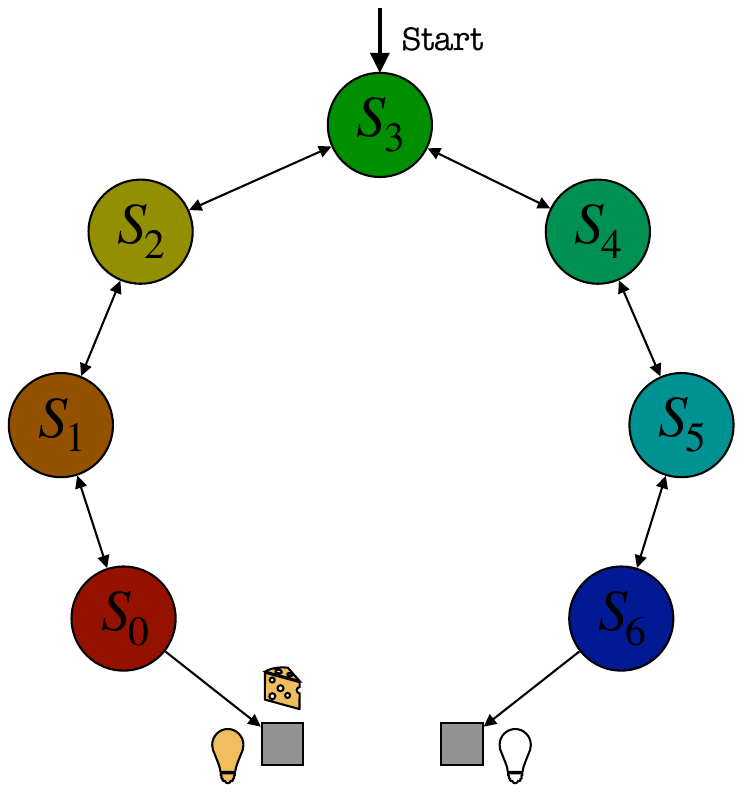}
        \caption{Chain Task.}
        \label{fig:chain_mdp}
    \end{subfigure}
    \hfill
    \begin{subfigure}[b]{0.23\columnwidth}
        \centering
        \includegraphics[height=2.2cm]{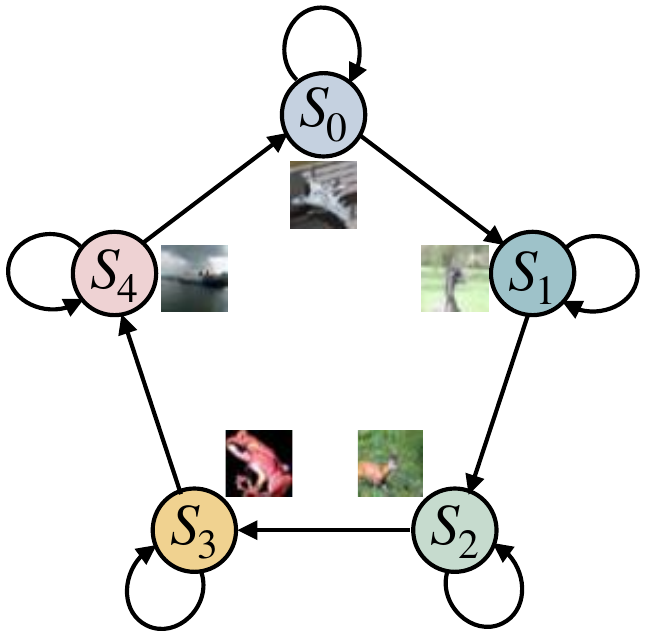}
        \caption{Image Task.}
        \label{fig:image-mdp}
    \end{subfigure}
    \hfill
    \begin{subfigure}[b]{0.23\columnwidth}
        \centering
        \includegraphics[height=2.2cm]{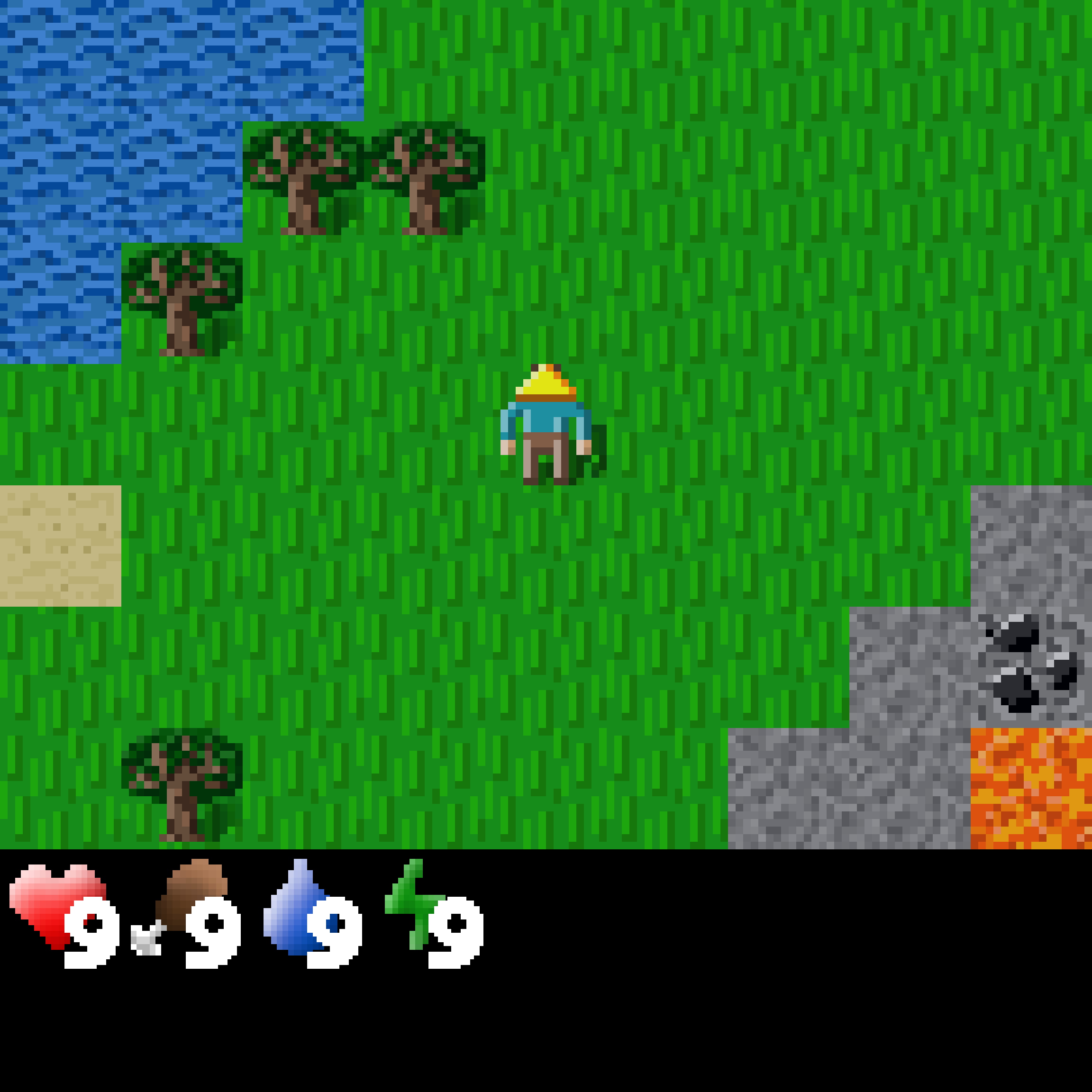}
        \caption{Craftax Task.}
        \label{fig:craftax-env}
    \end{subfigure}
    \caption{Environments used in the experiments.}
    \label{fig:envs-all}
\end{figure}

\paragraph{2D navigation task.} For the 2D navigation task, we use the $5\times5$ grid shown in Fig.~\ref{fig:prediction-grid}. The agent starts in the center state and can choose from four navigation actions, one for each cardinal direction. Each action typically moves the agent to an adjacent state, but with $10\%$ probability the intended action is replaced by one of the two perpendicular actions. The agent receives a reward when it transitions into one of the goal states at the green corners; otherwise, it receives no reward. Rewards are changed across tasks to introduce non-stationarity, as described in Table~\ref{tab:task-linear-pred} in the Appendix.

We use a Fourier basis up to second order to approximate the value function. In our variant of PT-TD learning, second-order features approximate the transient value function, while first-order features approximate the permanent value function. The PT-TD learning variant that uses all features to approximate both the permanent and transient value functions is denoted as ``full''. In both variants, the transient weights are reset to zero at the beginning of each task to induce plasticity, while previously learned values are retained through the permanent weights.

For comparison, we include two TD-learning baselines on which PT-TD learning is built. In the reset variant, weights are reset to zero at the start of each new task (full plasticity). In the continual variant, the agent continually updates its estimates on top of previously learned approximations. All algorithms are hyperparameter-tuned thoroughly as outlined in Appendix~\ref{app-sec:small-scale-experiments}.

To evaluate performance, we use a uniformly random policy with a discount factor of $0.9$. Each experiment runs for $1000$ episodes, with the task changing every $100$ episodes. The agent must therefore continually update its estimates to adapt to the current task. We use root-mean-square value error (RMSVE) as a performance metric (the lower, the better). We report the mean and $90\%$ confidence intervals over $30$ random seeds, computed with $z^{\ast}=1.645$.

\begin{figure}[ht]
    \centering
    \includegraphics[width=0.65\linewidth]{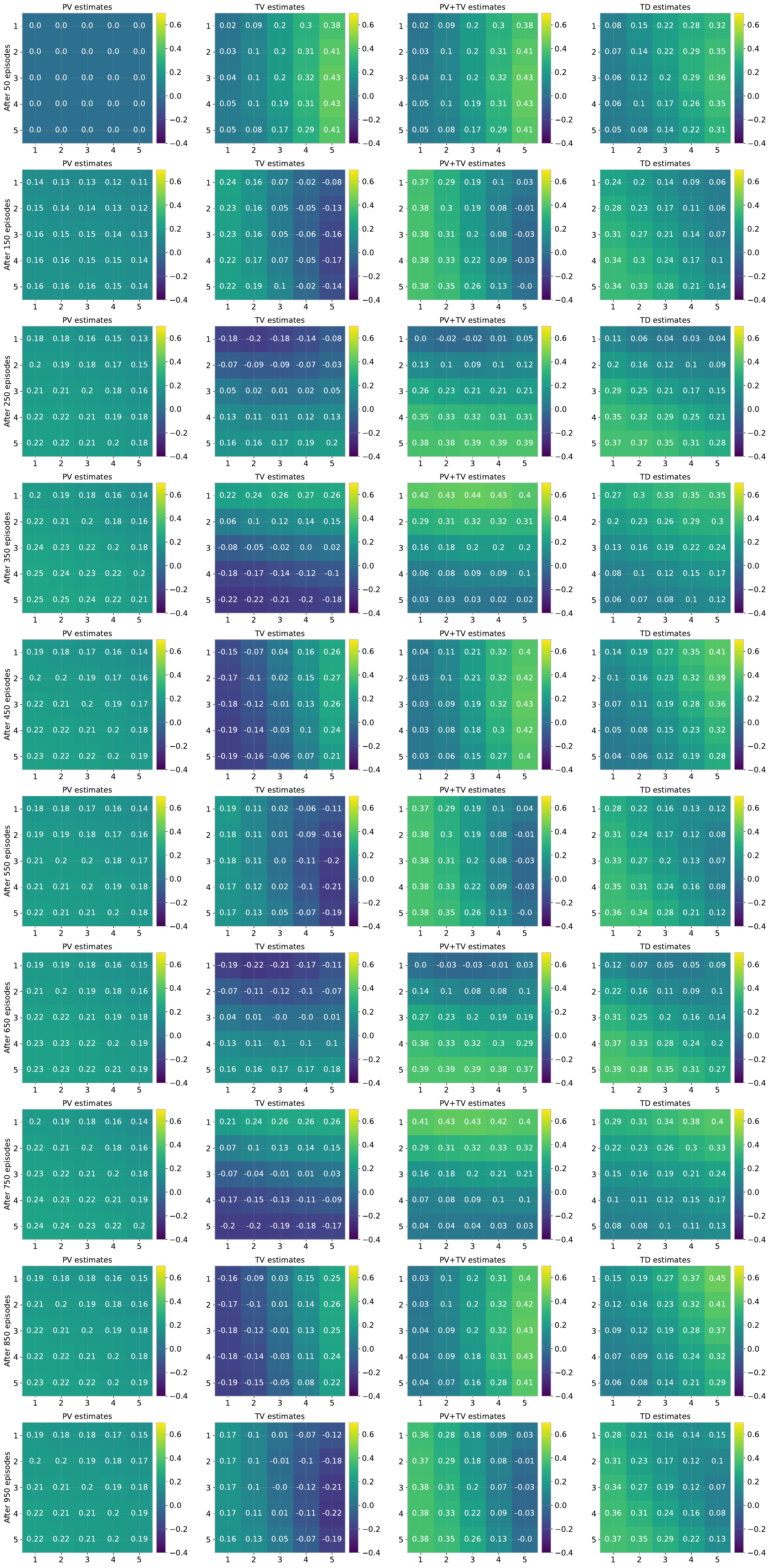}
    \caption{Value function heatmap for the prediction task}
    \label{app-fig:value-heatmap-prediction}
\end{figure}

\textbf{Results:} The results are shown in Fig.~\ref{fig:prediction-results}. In our variant, higher-order terms are used only to estimate the transient value function, enabling quick, low-variance, and precise adjustments to the permanent value function and resulting in the lowest overall RMSVE. First-order features provide sufficient expressivity while remaining low variance, making them well-suited for approximating the permanent value function (see Figure~\ref{app-fig:value-heatmap-prediction}). The PT variant that uses all features has a slightly higher RMSVE due to the increased variance introduced by using all features to approximate both value functions. Both PT-TD learning variants retain prior knowledge via the permanent value function, leading to lower error at the onset of recurring tasks. In contrast, the TD-learning baselines perform poorly, as they fail to preserve previous predictions while simultaneously adapting to new tasks, corroborating prior findings~\citep{anand2023prediction}.

\subsection{Control}
\label{subsec:control}

For the control problem, we use the simple chain environment with seven states shown in Fig.~\ref{fig:chain_mdp}. The agent starts in the central state and can choose either the \emph{left} or \emph{right} action to move to the neighbouring state in the corresponding direction, with a $10\%$ probability of a reversed effect. Goal states with positive rewards are located at both ends of the chain. Non-stationarity is introduced by varying the reward magnitude across tasks, as described in Table~\ref{tab:task-linear-control} in the Appendix.


Each state is represented by four features: the first three correspond to the red, green, and blue (RGB) components of the state, while the fourth encodes the light intensity, determined by the reward location in the first task for the entire experiment. In our variant of PTQ-learning, the first three features approximate the permanent value function, and the last feature approximates the transient value function. In contrast, all other algorithms we compare against, including the full variant of PTQ-learning and the reset and non-reset variants of Q-learning, use all features to approximate the value function. For the reset variant of Q-learning, the weights are reset at the beginning of each task.

We use a discount factor of $0.99$ and compute episodic returns to compare various algorithms. We run the experiment for $300$ episodes, with the task changing every $50$ episodes. We report the mean and $90\%$ confidence intervals over $30$ random seeds,  computed with $z^{\ast}=1.645$.

\begin{figure}[htb]
    \centering
    \begin{subfigure}[b]{0.53\columnwidth}
        \centering
        \includegraphics[width=\textwidth]{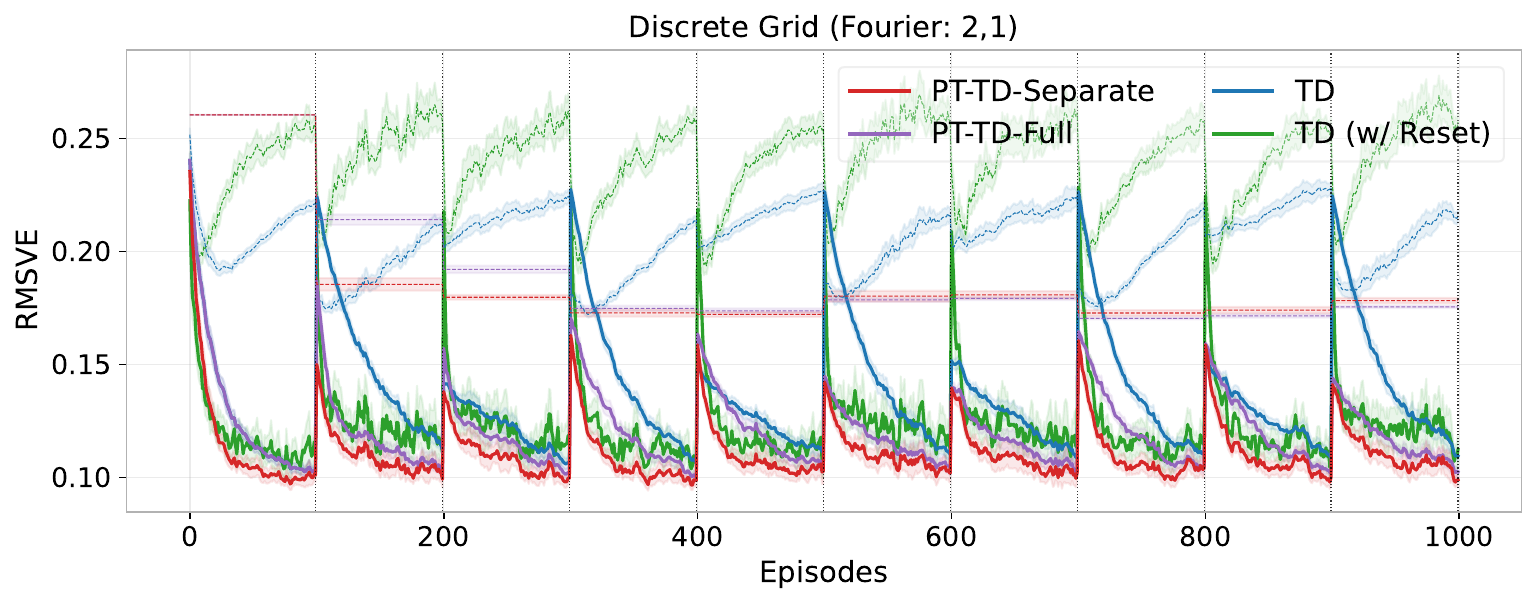}
        \caption{2D grid prediction results.}
        \label{fig:prediction-results}
    \end{subfigure}
    \hfill
    \begin{subfigure}[b]{0.44\columnwidth}
        \centering
        \includegraphics[width=\textwidth]{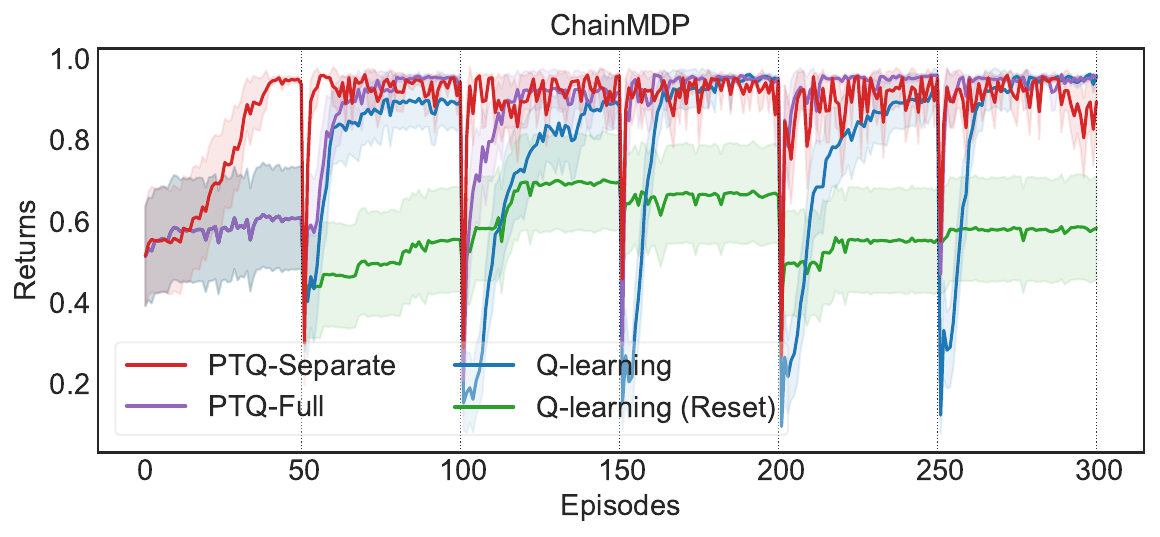}
        \caption{Control results.}
        \label{fig:control-results}
    \end{subfigure}
    \caption{(a) 2D grid prediction results. PT-TD-Separate variant has the lowest RMSVE in both online and other tasks (shown in solid and dotted lines, respectively). (b) Control results. The PTQ-Separate variant adapts faster when the tasks change.}
    \label{fig:small-scale-experiments-results}
\end{figure}

\textbf{Results:} The results are shown in Fig.~\ref{fig:control-results}. Because our approach uses a single reward-correlated feature to adjust transient value estimates, adaptation is rapid. It is the only method that learns a meaningful behaviour within the first $50$ episodes and re-adjusts it within five episodes whenever tasks change. PTQ-learning with all features fails to learn a meaningful policy during the initial $50$ episodes and adapts marginally more slowly when tasks subsequently change. The reset variant of Q-learning performs the worst, as no prior knowledge is retained at the start of each task. The continual variant of Q-learning takes longer to overcome bias in its action-value estimates and therefore adapts more slowly.

\section{Theoretical results}
\label{sec:theory}

In this section, we establish that both the permanent and the transient estimates converge to a unique fixed point when using linear function approximation, under standard assumptions. These results are significantly stronger than previously available tabular results~\citep{anand2023prediction} and, importantly, provide theoretical guarantees when using distinct feature spaces for the function approximators. \textit{That is, they demonstrate that using two distinct sets of features to estimate permanent and transient components does not lead to divergence, as is common in many linear RL scenarios.}

We leverage the two-timescale proof technique introduced by Borkar (see Appendix~\ref{thm:borkar-two})~\citep{borkar1997stochastic}, along with the conditions proposed in~\citet{bertsekas1996neuro} and~\citet{tsitsiklis1996analysis} in our proof. We consider updating the permanent value function at each timestep.

We make the following assumptions\footnote{The assumptions~\ref{assumption:step-size}, \ref{assumption:features}, and \ref{assumption:rapid-mixing} stem from stochastic approximation theory and were standardized by~\citet{bertsekas1996neuro} and~\citet{tsitsiklis1996analysis} for RL for the linear case. Assumption~\ref{assumption:task-distribution} was introduced by~\citet{anand2023prediction} to aid theoretical analysis in CRL.}:

\begin{assumption}
    \label{assumption:step-size}
    The step-sizes $\PMlr$ and $\TMlr$ satisfy $\sum_{t}{\PMlr_t} = \infty$, $\sum_{t}{\PMlr_t^2} < \infty$, $\sum_{t}{\TMlr_t} = \infty$, $\sum_{t}{\TMlr_t^2} < \infty$, $\lim_{t \to \infty} \tfrac{\PMlr_t}{\TMlr_t} \to 0$.
\end{assumption}

\begin{assumption}
    \label{assumption:features}
    The permanent and transient feature matrices, $\PMFeatMatrix \in \sR^{|\s| \times d}$ and $\TMFeatMatrix \in \sR^{|\s| \times l}$, are full column rank, i.e. the column vectors are linearly independent. Also, their norms are bounded, $\norm{\PMFeatMatrix} \leq M_1$, $\norm{\TMFeatMatrix} \leq M_2$ where $M_1$ and $M_2$ are constants.
\end{assumption}

\begin{assumption}
    \label{assumption:task-distribution}
    There are $N$ tasks and task $\tau$ is i.i.d. sampled according to $p_{\tau}$. Each task is an MDP, $\m_{\tau}=(\s, \act, \rew_{\tau}, \ptask, \gamma)$, let $\E_\tau$ denote the expectation with respect to the task distribution. The rewards for each task $\rew_{\tau}$ are bounded. The task boundaries are observable.
\end{assumption}

\begin{assumption}
    \label{assumption:rapid-mixing}
    Every task, $\task$, induces an irreducible, aperiodic Markov chain under the fixed evaluation policy $\pol$ and the chain is rapidly mixing:
    $$|\ppoltask(S_t = s| S_0) - \dis(s)| \leq C \sigma^t,\ \forall S_0 \in \s, \sigma < 1,$$ where $C$ is a constant. 
\end{assumption}

\begin{theorem}[Main result]
    \label{thm:main}
    Under Assumptions~\ref{assumption:step-size}--\ref{assumption:rapid-mixing}, the sequence of expected updates computed by permanent and transient updates converges to a unique fixed point:
    \begin{small}
    \begin{align*}
        \PMparams^* &= \Etask[\PMFeatMatrix^T \Dtask \Pi_{\TMFeatMatrix, \task} \PMFeatMatrix]^{-1} \Etask[\PMFeatMatrix^T \Dtask \TMFeatMatrix \w^{(TD)}_{\TMFeatMatrix, \tau}], \\
        \TMparams^{\ast}_\tau &= \w^{(TD)}_{\TMFeatMatrix, \tau} - \left(\TMFeatMatrix^T \Dtask (I- \gamma \ppoltask)\TMFeatMatrix \right)^{-1} \TMFeatMatrix^T \Dtask (I - \gamma \ppoltask) \PMFeatMatrix \PMparams^{\ast},\quad \text{where} \\
        \Pi_{\TMFeatMatrix, \tau} &= \TMFeatMatrix \left(\TMFeatMatrix^T \Dtask (I - \gamma \ppoltask)\TMFeatMatrix \right)^{-1} \TMFeatMatrix^T \Dtask (I - \gamma \ppoltask), \\
        \w^{(TD)}_{\TMFeatMatrix, \tau} &= \TMFeatMatrix \left(\TMFeatMatrix^T \Dtask (I- \gamma \ppoltask)\TMFeatMatrix \right)^{-1} \TMFeatMatrix^T \Dtask \rpoltask.
    \end{align*}
    \end{small}
\end{theorem}

\begin{proof}
    We outline the proof here, and provide full details in Appendix (Section~\ref{app-sec:proofs}).

    Because of the disparity in learning rates between permanent and transient updates, the permanent estimates appear stationary while the transient values are being updated. And the transient values appear to converge when updating permanent estimates.
    
    So, we first establish the convergence of the transient updates, treating the permanent values as fixed, by verifying the conditions outlined in Theorem~\ref{thm:tsitsiklis}~\citep{tsitsiklis1996analysis} (see Lemma~\ref{lemma:tm-convergence} in Appendix). By substituting the fixed point of the transient parameters into the permanent updates, we then show that the required conditions are satisfied for them as well, and therefore convergence follows from Theorem~\ref{thm:borkar-two}~\citep{borkar1997stochastic} (see Lemma~\ref{lemma:pm-convergence} in Appendix).
\end{proof}

\begin{corollary}
If $\TMFeatMatrix = \PMFeatMatrix$, then 
\begin{align*}
    \PMparams^* &= \Etask\!\left[\PMFeatMatrix^T \Dtask \PMFeatMatrix\right]^{-1} 
                   \Etask\!\left[\PMFeatMatrix^T \Dtask \PMFeatMatrix \w^{(TD)}_{\PMFeatMatrix, \tau}\right], \\
    \TMparams^{\ast}_\tau &= \w^{(TD)}_{\PMFeatMatrix, \tau} - \PMparams^{\ast}.
\end{align*}
Moreover, in the single-task setting, $\PMparams^* = \w^{(TD)}_{\PMFeatMatrix, \tau}$ and $\TMparams^{\ast}_\tau = 0$.
\end{corollary}

\begin{proof}
    The proof is included in Appendix~\ref{app:corollary}.
\end{proof}

The above corollary implies that in the single-task setting, if both the permanent and transient value functions are approximated using the same feature representation, then the permanent value function alone suffices to capture the predictions, while the transient component converges to zero.

\section{Non-parametric Transient Approximations}
\label{sec:scaling-up-PT-memories}

In this section, we introduce a new non-parametric architecture that meets the desiderata of transient representations (as described in Sec.~\ref{sec:pt-rep-definition}). Our architecture is inspired by tile coding~\citep{sutton1995generalization}, CMACs~\citep{miller1990cmac}, and related sparse distributed memory architectures that were used as standalone function approximators in the 1990s and early 2000s~\citep{kanerva1992sparse, ratitch2004asparse, ratitch2004bsparse}. While those early approximators suffered from the curse of dimensionality, we bypass this limitation by using hashing and pairing it with deep neural networks, rather than treating it as a stand-alone approximator.

We assume that each observation consists of several channels, each indicating the presence or absence of a specific item type at a particular location in the agent's view. Many commonly used RL environments, such as Minigrid~\citep{MinigridMiniworld23}, Craftax~\citep{matthews2024craftax}, and MinAtar~\citep{young2019minatar}, are structured in this way. This assumption will allow us to produce hash signatures of observations using MinHash~\citep{broder1997resemblance}.
\begin{figure}[htb]
    \centering
    \begin{subfigure}[b]{0.54\columnwidth}
        \centering
        \includegraphics[width=0.98\columnwidth]{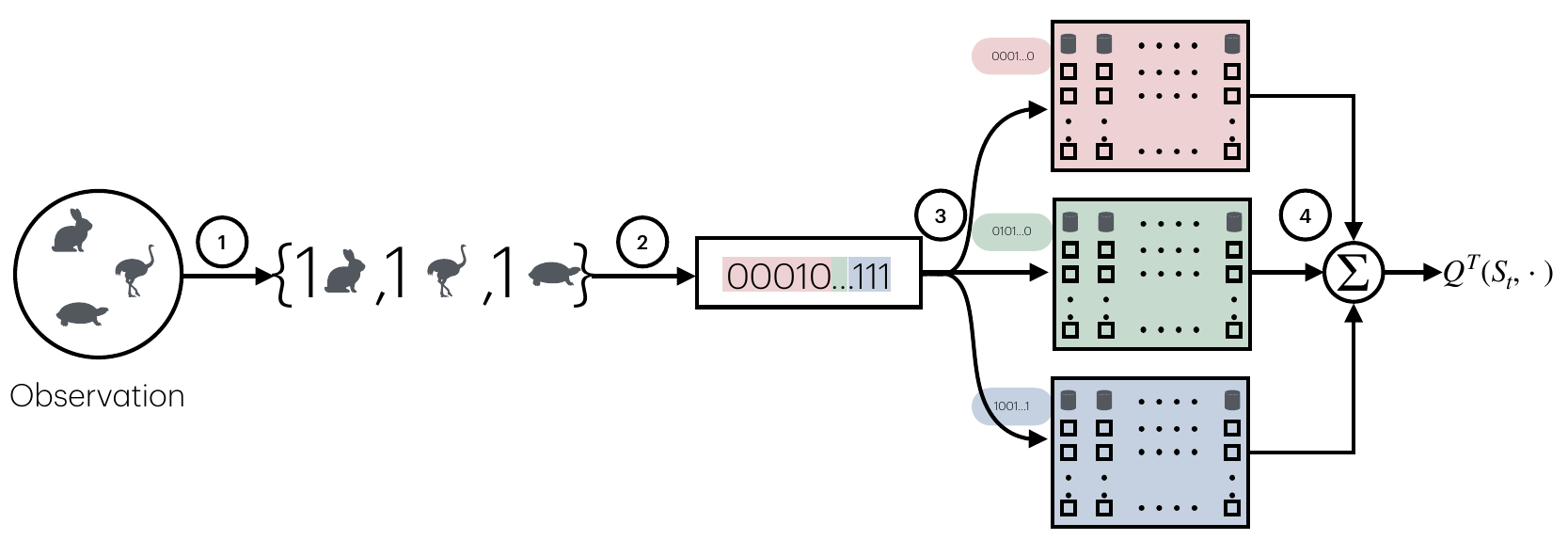}
        \caption{An end-to-end overview of the transient approximator.}
        \label{fig:transient-memory-ete}
    \end{subfigure}
    \hfill
    \begin{subfigure}[b]{0.44\columnwidth}
        \centering
        \includegraphics[width=0.98\columnwidth]{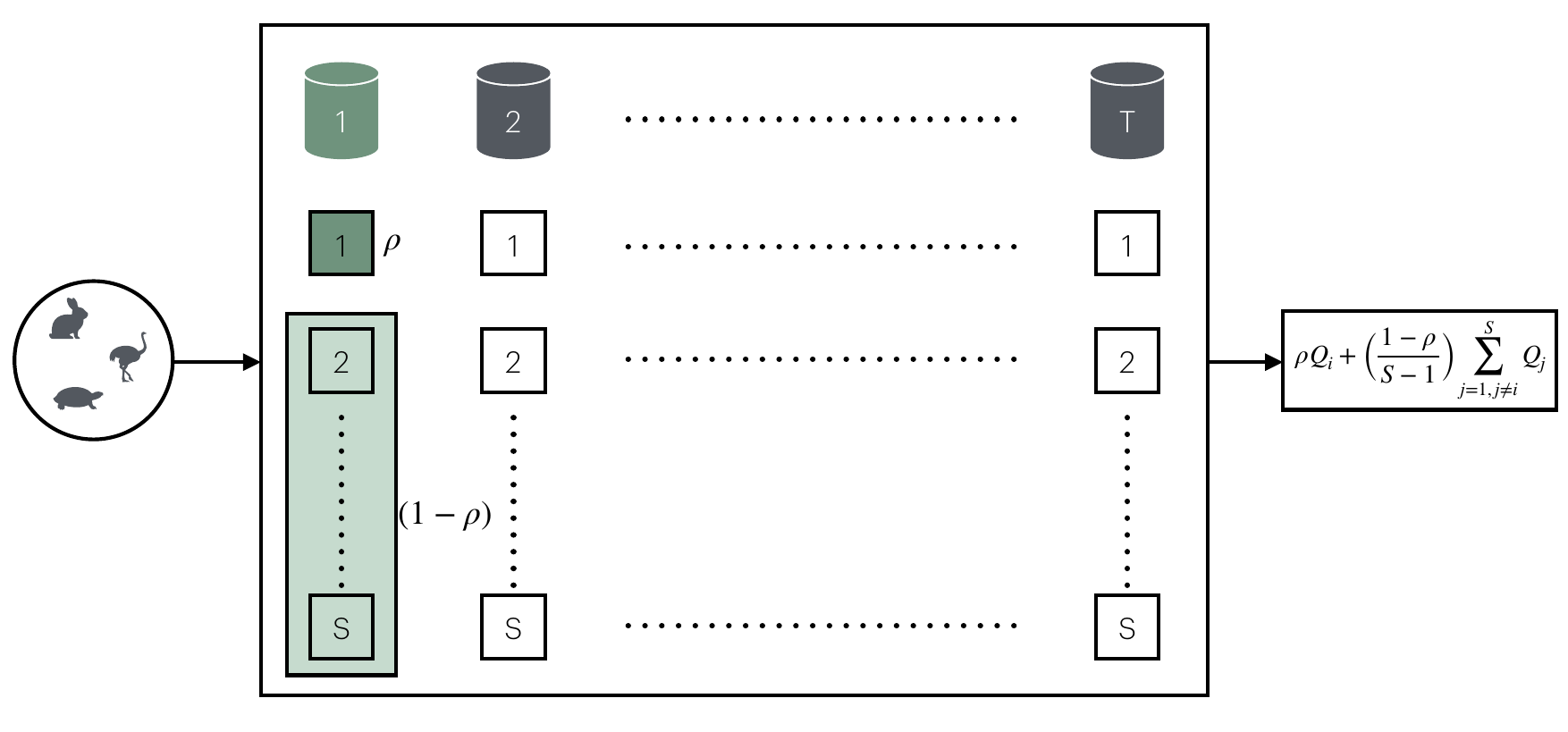}
        \caption{Transient table.}
        \label{fig:transient-table}
    \end{subfigure}
    \caption{(a)\textbf{Transient Approximator.} The observation is mapped to value function in four steps: (1) tokenization; (2) hashing; (3) binning; (4) value estimation. (b) \textbf{Individual Table.} Within each table, the value of the query observation is weighted by $\rho$, and the remaining mass ($1-\rho$) is distributed among its neighbours to form the estimate.}
    \label{fig:transient-memory}
\end{figure}

The overall process is illustrated in Fig.~\ref{fig:transient-memory}. It transforms an observation into a sparse, hashed representation that supports non-parametric value estimation with controlled generalization. The pipeline has four main stages:

\textbf{1. Tokenization.} Each observation is converted into a set of discrete tokens, where each token encodes the type of object present at a particular $(x,y)$ position in the agent’s view. This converts the raw grid-like observation into a symbolic set representation. In image-based tasks, grayscale observations (derived from the RGB counterpart) can be quantized into $N$ levels and treated as symbolic observations. Although image tokenization is still undergoing a lot of work, and any new method could be incorporated into our approach.

\textbf{2. Hashing.} The tokenized set is mapped to a compact hash signature and an auxiliary tag using several independent MinHash functions. MinHash is used because it preserves the Jaccard similarity: the probability that two sets produce the same MinHash value is exactly their Jaccard score.\footnote{The Jaccard score between two sets $A$ and $B$ is, $J(A, B) = \tfrac{|A \cap B|}{|A \cup B|}$, and it is widely used in retrieval~\citep{broder1997resemblance}, bio-informatics~\citep{ondov2016mash}, and clustering~\citep{tan2016introduction}.} As a result, observations that share many elements produce similar signatures (i.e. overlap in many bits), which form the basis for local generalization, preventing undesirable hash collisions.

\textbf{3. Binning.} The signature is partitioned into multiple subsequences, each of which indexes into a bin in a transient table. Conceptually, each table corresponds to one \emph{tiling} in tile coding~\citep{sutton1995generalization}. Within a bin, there are $S$ slots. When a new observation maps into a bin, it occupies an empty slot if one is available; otherwise, the least-recently used (LRU) slot is evicted. The slot's initial value is set to the average of values already stored in that bin, so new entries inherit some local context. The auxiliary tag computed in the previous step is also stored in the slot, making it unique to that observation (see Alg.~\ref{algo:put-method}).

\textbf{4. Value estimation.} To induce generalization akin to CMACs~\citep{miller1990cmac}, the value of a query observation is not taken from a single slot alone. Instead, its estimate is a convex combination of the focal slot and the other occupied slots in the \emph{matched bin} (see Fig.~\ref{fig:transient-table}): $\TMcontrol_{t+1}(x, a; k) = \rho \TMcontrol_{t+1}(x, a; k) + \Big(\frac{1-\rho}{S - 1}\Big)\sum_{j \neq i} \TMcontrol_{t+1}(o_j, a; k)$, where $\rho$ controls the relative weight of the focal slot versus its neighbours, $x$ is the query observation, $o_j$ is the observation stored in slot $j$ of the matched bin, $S$ is the total number of slots per bin, and $\TMcontrol_{t+1}(x, a; k)$ is the transient value of the $(x, a)$ pair estimated by the $k$-th table. This encourages smooth interpolation between similar observations while still preserving distinct slot identities. The final transient value is obtained by summing the estimates from all transient tables. For more details, see~\ref{algo:get-method} and~\ref{algo:put-method} in the Appendix.

Since several tables and their corresponding bins contribute to the overall estimate of the transient value function, the TD-error is distributed among them in proportion to their contributions (see~\ref{algo:update-td-error-method} in Appendix). Specifically, the TD-error is first split evenly across all $T$ tables, and within each table it is further divided among the slots in the \emph{matched bin}. The focal slot receives weight $\rho$, while its neighbouring slots share the remaining weight $\tfrac{1-\rho}{S-1}$. The updates are analogous to those in tile coding: $\TMcontrol_{t+1}(O_t, A_t; k, i) \gets \TMcontrol_{t}(O_t, A_t; k, i) + \frac{\TMlr}{T} \eta_{k,i} \delta_t$, where $\TMlr$ is the learning rate of the transient updates, $T$ is the total number of tables, and $\delta_t$ is the TD-error at time $t$. The weighting factor $\eta_{k,i}$ reflects the contribution of slot $i$ in table $k$: where $\eta_{k,i} = \rho$ if slot $i$ stores the query observation in the matched bin, $\eta_{k,i} = \tfrac{1-\rho}{S-1}$ if slot $i$ is a neighbour in the same bin, and $0$ otherwise.

Although the transient table has a fixed size and older observations within a bin are evicted using an LRU strategy, it can still retain observations from a long time ago if temporally adjacent states are similar and therefore continue to map to the same bin, while observations encountered much later differ from those earlier states and are hashed into different bins, leaving the older ones untouched.

\section{Large-Scale Experiments}
\label{sec:experiment}

We conducted experiments on CIFAR image tasks, an MDP variant of the commonly studied task in the plasticity literature (e.g.,~\citet{lyle2023understanding}) and on Craftax-Classic to demonstrate the effectiveness of our approach, which trades off stability and plasticity at the system level. Our implementation builds on the
\href{https://github.com/mttga/purejaxql/tree/main}{PQN library} and is publicly available at \url{https://github.com/NishanthVAnand/pt_representations_crl}.

\paragraph{Image Task.} The task is an image-based MDP consisting of five states and five actions. Observations for each state are sampled from specific categories of the CIFAR-10 dataset. We evaluate three ``full CRL'' variations of this task, where the task boundaries are hidden from the agent, by restricting the sub-sampled data size to $250$, $500$, and $1000$ images per category. In this task, transitions are stochastic: selecting the rewarding action (+1) leads to the next state with probability $0.8$ (otherwise, remaining in the current state with probability $0.2$), while suboptimal actions yield a small negative penalty and result in remaining in the current state with probability $0.8$. Rewards are perturbed with Gaussian noise ($\mu=0, \sigma=0.01$). To induce non-stationarity, the optimal action is switched (from a set of three) every $100k$ timesteps over a $600k$-timestep period. We use $\gamma=0.9$. We report mean rewards over $600$ timesteps ($90$\% C.I. over $30$ seeds). Hyperparameter details are provided in Appendix~\ref{app-sec:image-task}.

\textbf{Baselines:} We compare our approach against \textbf{PQN}, the state-of-the-art value-based algorithm on Craftax~\citep{Gallici25simplifying}; three variants of PT: \textbf{PT-IHT}, which utilizes Index Hash Tables (IHTs) rather than slot-based table (MinHash-IHT-based tile coding), \textbf{T-Only}, which relies solely on the non-parametric transient approximator (Sec.~\ref{sec:scaling-up-PT-memories}), and \textbf{PT-NN}~\citep{anand2023prediction}, a fully neural network-based baseline that is trained end-to-end; \textbf{EWC} that augments PQN with elastic weight consolidation (EWC) to preserve stability~\citep{kirkpatrick2017overcoming}; \textbf{CB} that pairs PQN with continual backprop (CB) to preserve plasticity~\citep{dohare2024loss}; and Neural Episodic Control (\textbf{NEC}) which has has a parametric, non-parametric split~\citep{pritzel2017neural}.

We also used another variant with $250$ images to test the adaptation speed. In this experiment, we train various agents on a single task for $250k$ timesteps and evaluate the speed of adaptation to a new task over the next $50k$ timesteps. We conduct this experiment to compare against popular methods that leverage task boundary information, \textit{while keeping it hidden in our approach}. In particular, we compared against \textbf{CompoNet} that uses attention to combine previously learned policies with the latest one~\citep{malagon2025self}, \textbf{Reset(Final)} that resets the final layer when the task changes~\citep{nikishin2022primacy}, \textbf{Reset(Full)} that fully reinitializes the network for full plasticity.

To enable MinHash hashing on image inputs, we first converted the RGB image to grayscale, then quantized it to $64$ levels, treating the resulting processed image as a set of pixel-level symbols.

\begin{figure}[htb]
    \centering
    \begin{subfigure}[b]{0.32\columnwidth}
        \centering
        \includegraphics[width=\columnwidth]{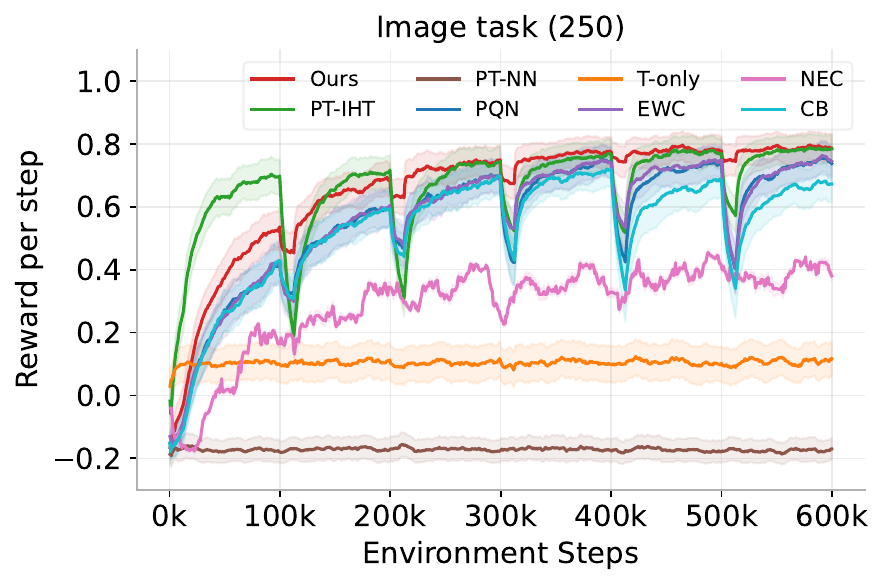}
        \caption{Image Task 250.}
        \label{fig:img-task-250}
    \end{subfigure}
    \hfill
    \begin{subfigure}[b]{0.32\columnwidth}
        \centering
        \includegraphics[width=\columnwidth]{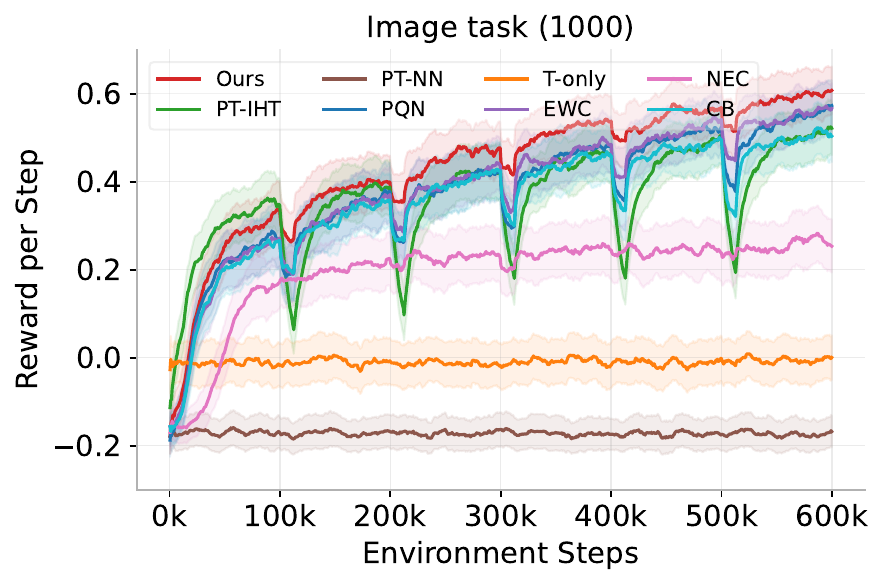}
        \caption{Image Task 1000.}
        \label{fig:img-task-1000}
    \end{subfigure}
    \hfill
    \begin{subfigure}[b]{0.32\columnwidth}
        \centering
        \includegraphics[width=\columnwidth]{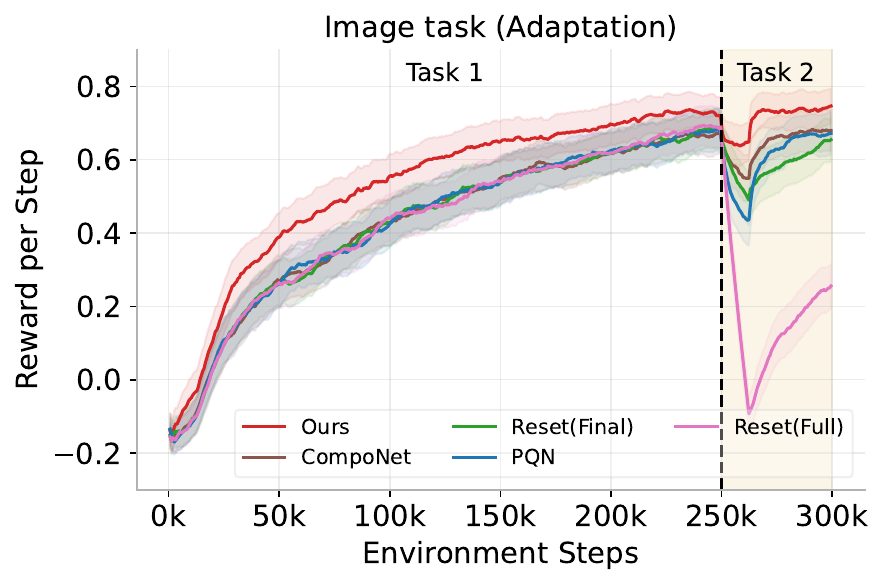}
        \caption{Image Task Adapt.}
        \label{fig:cifar-10-env}
    \end{subfigure}
    \caption{\textbf{Image Task Results.} Our algorithm adapts rapidly when the task changes, whether task boundaries are available or hidden, while other methods are slow.}
    \label{fig:image-mdp-results}
\end{figure}

\begin{figure}[h]
    \centering
    \includegraphics[width=0.5\columnwidth]{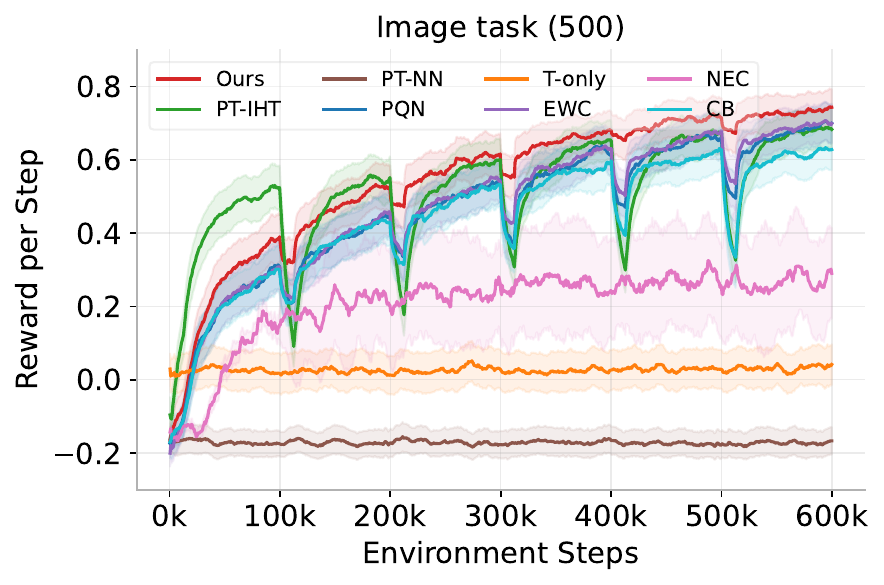}
    \caption{Image Task (500) Results.}
    \label{app-fig:img-task-500}
\end{figure}

\textbf{Results:} The results are presented in Figures~\ref{fig:image-mdp-results} and \ref{app-fig:img-task-500}. Our approach demonstrated superior performance compared to the baselines. Using a non-parametric approximation with controlled generalization ensured precise transient updates and rapid adaptation to task changes—a crucial advantage over the NN-based baselines. In contrast, PT-NN and T-only failed to learn meaningful policies due to instability in updates and insufficient capacity, respectively. PQN achieves stability through batch updates but suffers from slow initial learning and slow recovery after task changes. Augmenting it with EWC or CB didn't affect its performance, highlighting that mechanistic interventions don't necessarily result in rapid adaptation. Similarly, PT-IHT is hampered by the size of the observation space, where increased hash collisions lead to detrimental over-generalization and degraded value estimates, emphasizing the importance of our slot-based design. Because NEC's estimates depend on the features learnt by a NN, it is sensitive to training, and its overall performance is poor.

In the adaptation experiment, \textit{our approach adapted faster than methods that leverage task-boundary information (and more parameters), demonstrating the benefits of the system-level decomposition with complementary representations.}

\paragraph{Craftax-Classic.} The environment is open-ended, containing $22$ achievements of varying difficulty. At each step $t$, the agent receives a $7\times9$ grid observation containing the object types in its view (Fig.~\ref{fig:craftax-env}), along with its inventory (e.g., wood, stone, or crafted tools) and intrinsic variables (e.g., health, hunger, or thirst). The agent can take one of $17$ available actions. Within each episode, it receives a reward the first time it completes an achievement, such as collecting coal or crafting an iron pickaxe. Simpler achievements yield smaller rewards, while more advanced ones yield larger rewards. An episode ends when its length reaches $1000$ steps, when the agent completes all $22$ achievements, when it is killed by a zombie or a skeleton, or when its intrinsic variables are depleted. More details can be found in~\cite{matthews2024craftax}. We conducted two experiments: \emph{two-stream tasks} and \emph{benchmarking with $250$M using 1024 parallel environments} to demonstrate the competitiveness of our approach on complex tasks. We provide the full hyperparameter and architecture details in the Appendix (see~\ref{app-sec:craftax-details}).

\begin{wrapfigure}{r}{0.6\textwidth}
    \vspace{-5pt}
    \centering
    \begin{subfigure}[c]{0.48\linewidth}
        \centering
        \includegraphics[width=0.95\textwidth]{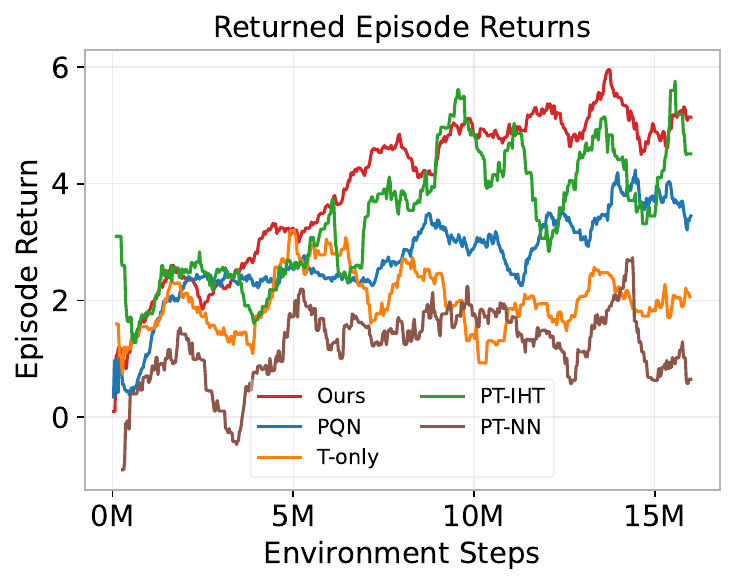}
        \caption{2-parallel envs.}
        \label{fig:online-craftax-results}
    \end{subfigure}
    \hfill
    \begin{subfigure}[c]{0.48\linewidth}
        \centering
        \includegraphics[width=0.95\textwidth]{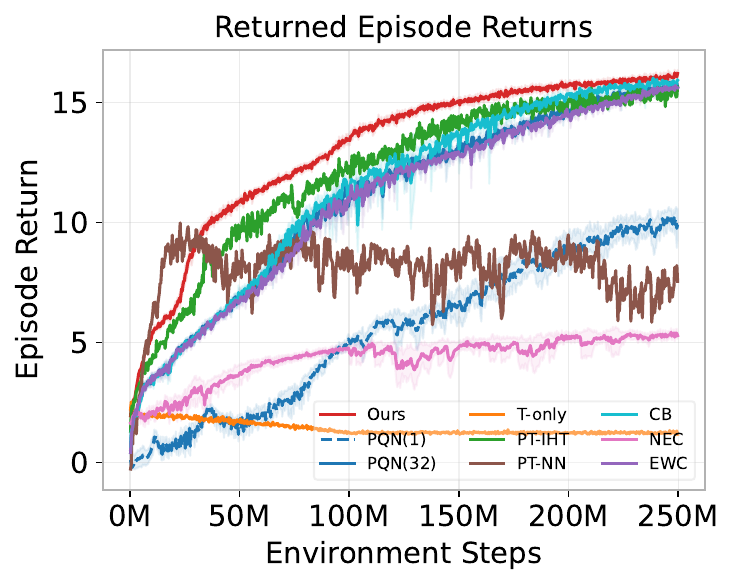}
        \caption{$1024$-parallel envs.}
        \label{fig:250M-craftax-results}
    \end{subfigure}
    \caption{\textbf{Craftax-Classic Results.} Our algorithm outperforms other continual learning algorithms, including ablations, in both two-parallel and 1024-parallel environments.}
    \label{fig:craftax-experiment-results}
    \vspace{-5pt}
\end{wrapfigure}

\begin{figure}[H]
  \centering
  \begin{subfigure}[c]{0.24\textwidth}
    \centering
    \includegraphics[width=0.99\columnwidth]{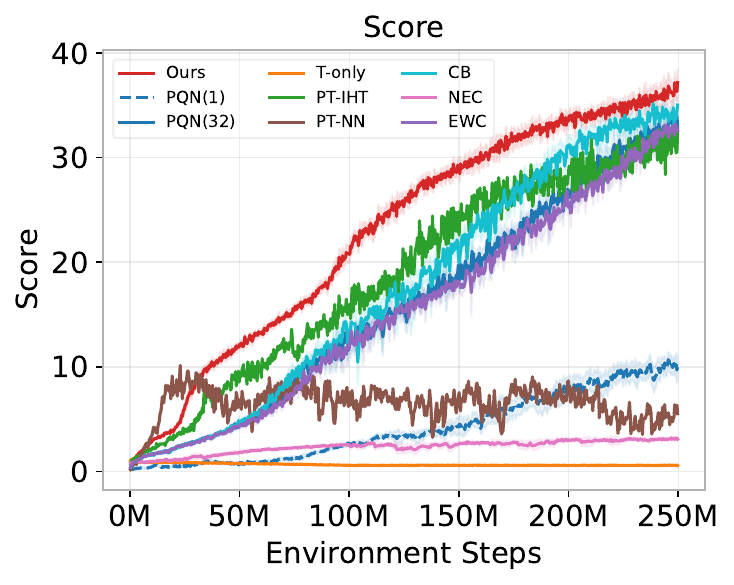}
    \caption{Score.}
    \label{fig:main-score}
  \end{subfigure}\hfill
  \begin{subfigure}[c]{0.24\textwidth}
    \centering
    \includegraphics[width=0.99\columnwidth]{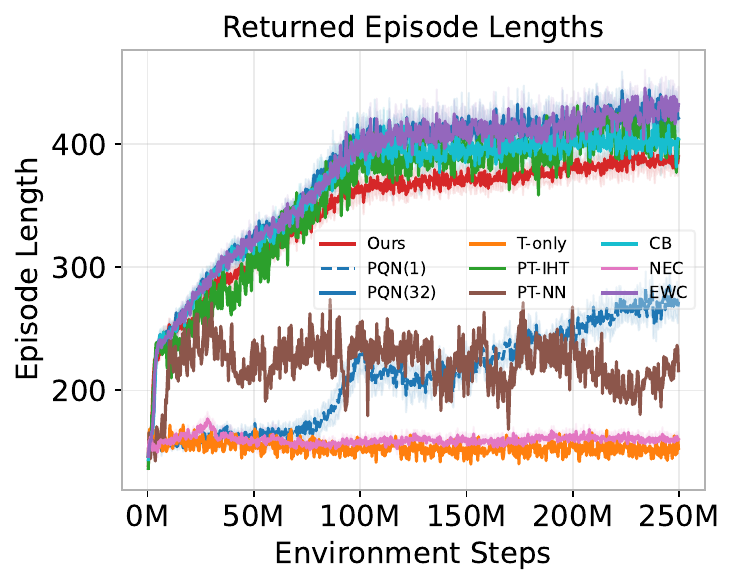}
    \caption{Episode Lengths.}
    \label{fig:main-episode-length}
  \end{subfigure}\hfill
  \begin{subfigure}[c]{0.24\textwidth}
    \centering
    \includegraphics[width=0.99\columnwidth]{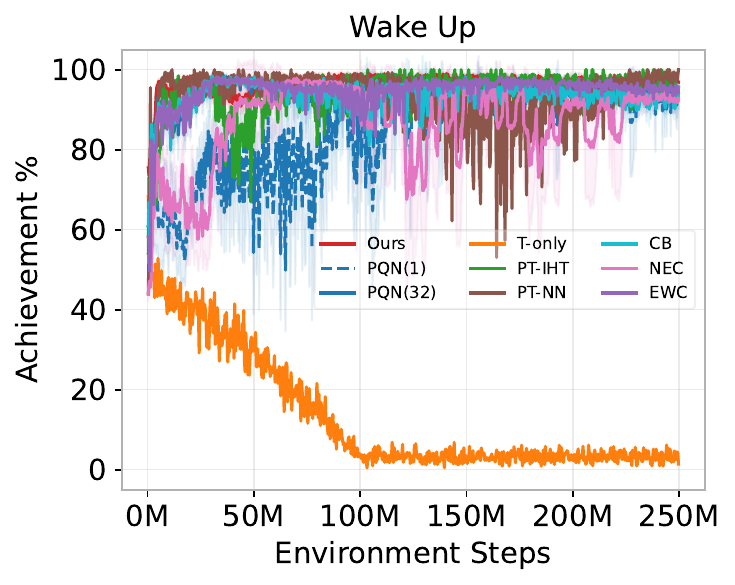}
    \caption{Wake Up.}
    \label{fig:main-wake-up}
  \end{subfigure}\hfill
  \begin{subfigure}[c]{0.24\textwidth}
    \centering
    \includegraphics[width=0.99\columnwidth]{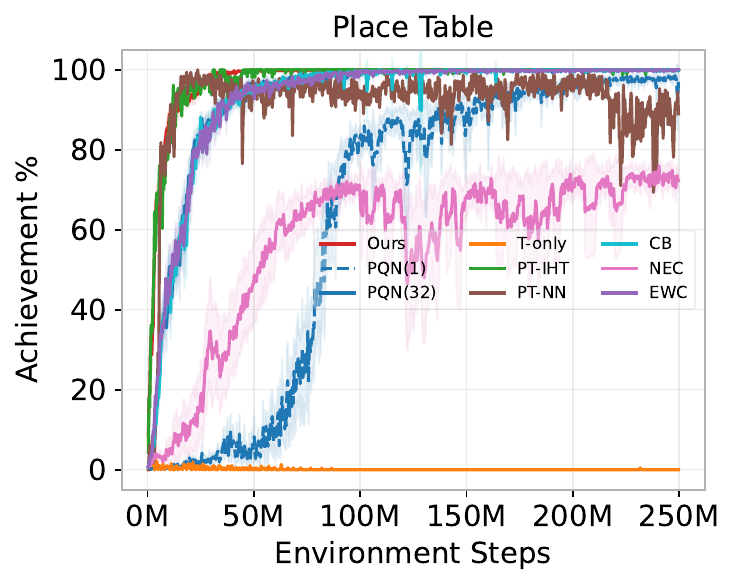}
    \caption{Place Table.}
    \label{fig:main-place-table}
  \end{subfigure}

  \begin{subfigure}[c]{0.24\textwidth}
    \centering
    \includegraphics[width=0.99\columnwidth]{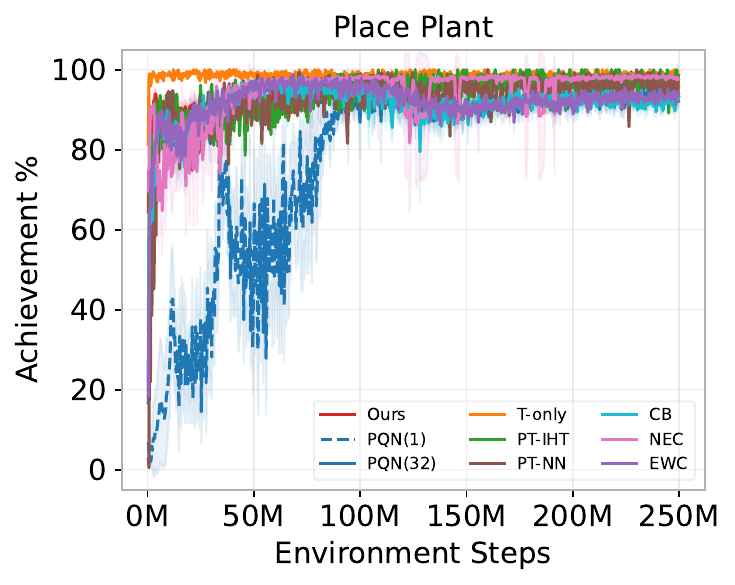}
    \caption{Place Plant.}
    \label{fig:main-place-plant}
  \end{subfigure}\hfill
  \begin{subfigure}[c]{0.24\textwidth}
    \centering
    \includegraphics[width=0.99\columnwidth]{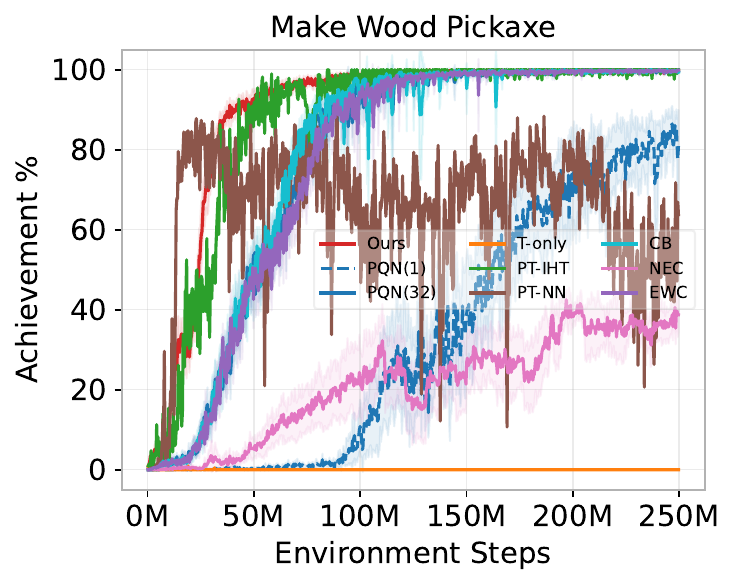}
    \caption{Make Wood Pickaxe.}
    \label{fig:main-make-wood-pickaxe}
  \end{subfigure}\hfill
  \begin{subfigure}[c]{0.24\textwidth}
    \centering
    \includegraphics[width=0.99\columnwidth]{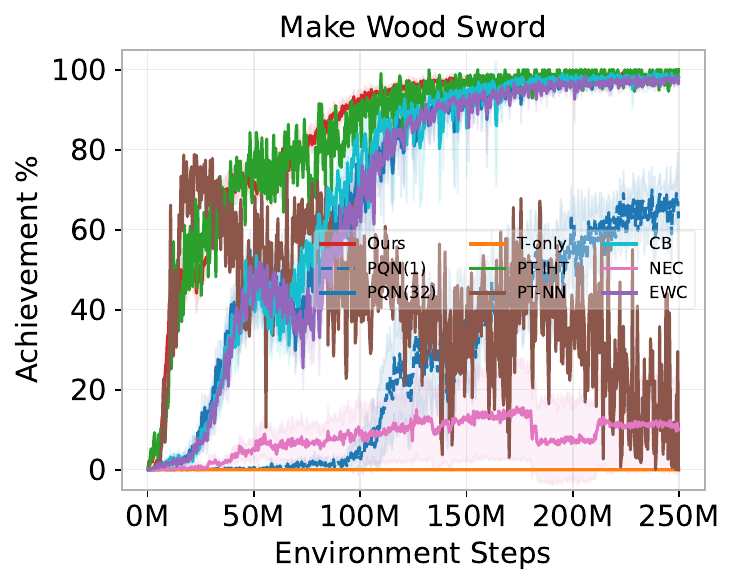}
    \caption{Make Wood Sword.}
    \label{fig:main-make-wood-sword}
  \end{subfigure}\hfill
  \begin{subfigure}[c]{0.24\textwidth}
    \centering
    \includegraphics[width=0.99\columnwidth]{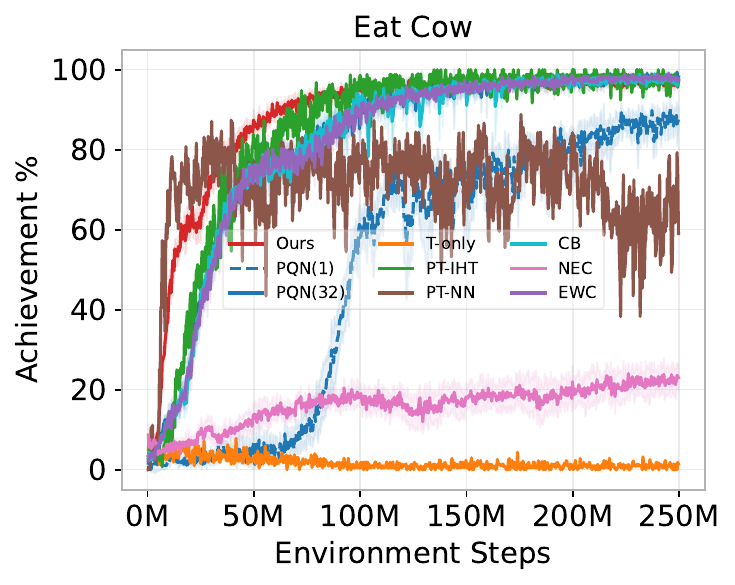}
    \caption{Eat Cow.}
    \label{fig:main-eat-cow}
  \end{subfigure}

    \begin{subfigure}[c]{0.24\textwidth}
    \centering
    \includegraphics[width=0.99\columnwidth]{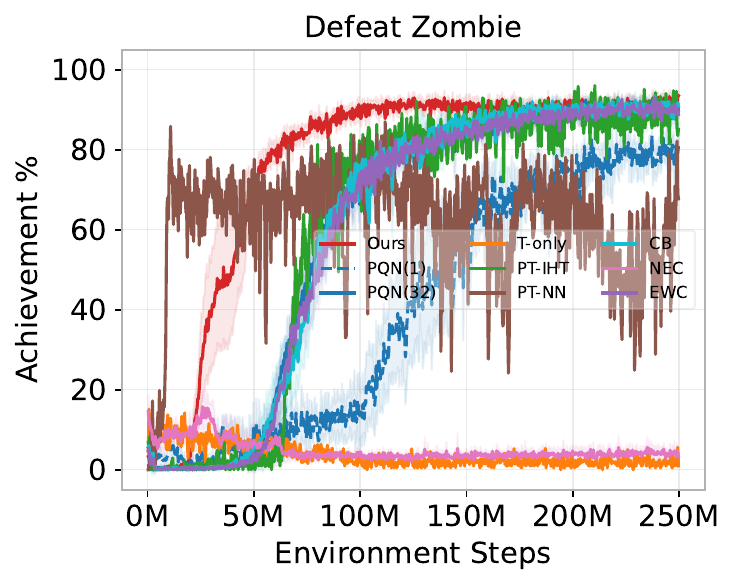}
    \caption{Defeat Zombie.}
    \label{fig:main-defeat-zombie}
  \end{subfigure}\hfill
  \begin{subfigure}[c]{0.24\textwidth}
    \centering
    \includegraphics[width=0.99\columnwidth]{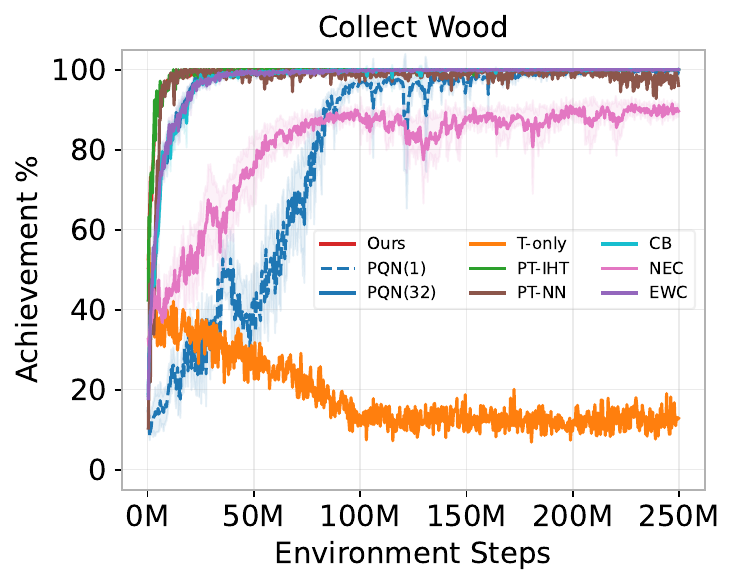}
    \caption{Collect Wood.}
    \label{fig:main-collect-wood}
  \end{subfigure}\hfill
  \begin{subfigure}[c]{0.24\textwidth}
    \centering
    \includegraphics[width=0.99\columnwidth]{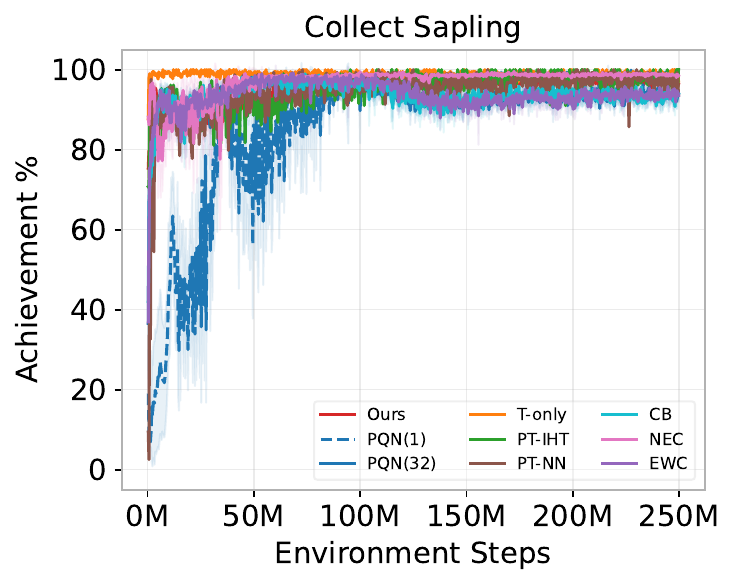}
    \caption{Collect Sapling.}
    \label{fig:main-collect-sapling}
  \end{subfigure}\hfill
  \begin{subfigure}[c]{0.24\textwidth}
    \centering
    \includegraphics[width=0.99\columnwidth]{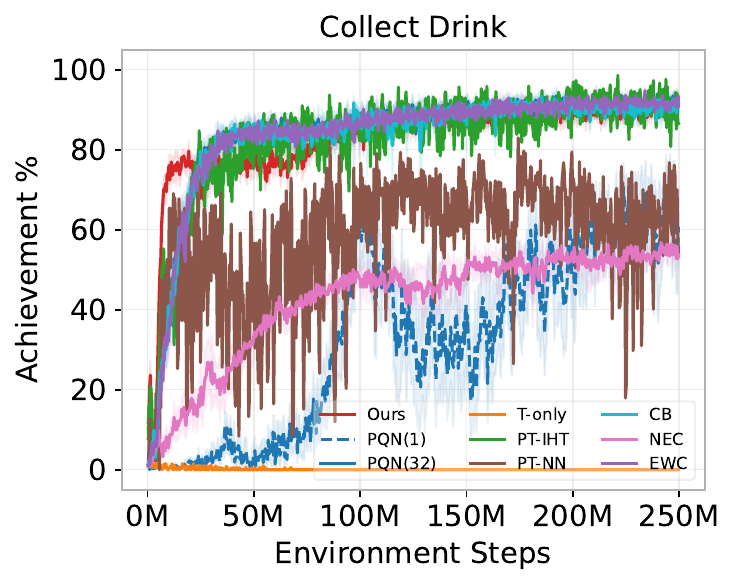}
    \caption{Collect Drink.}
    \label{fig:main-collect-drink}
  \end{subfigure}

    \begin{subfigure}[c]{0.24\textwidth}
    \centering
    \includegraphics[width=0.99\columnwidth]{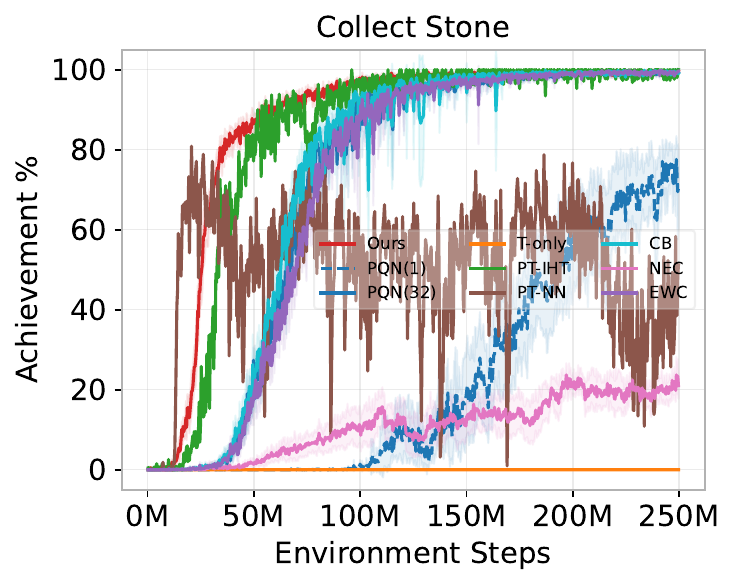}
    \caption{Collect Stone.}
    \label{fig:main-collect-stone}
  \end{subfigure}\hfill
  \begin{subfigure}[c]{0.24\textwidth}
    \centering
    \includegraphics[width=0.99\columnwidth]{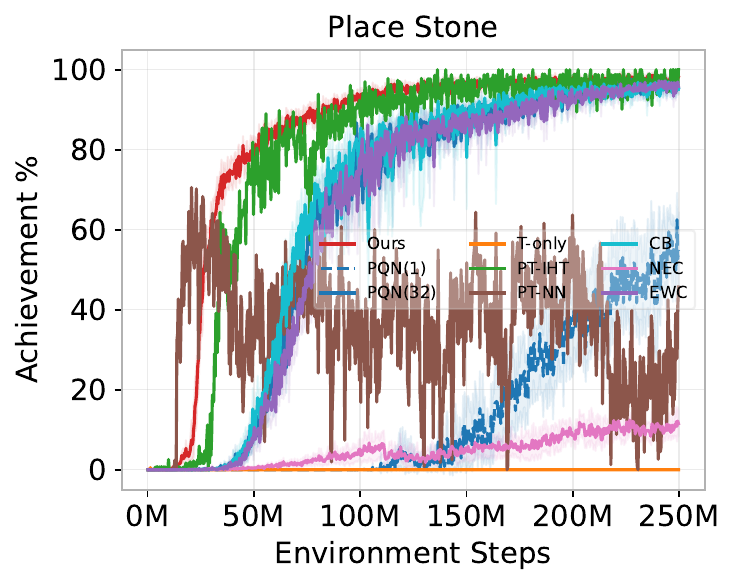}
    \caption{Place Stone.}
    \label{fig:main-place-stone}
  \end{subfigure}\hfill
  \begin{subfigure}[c]{0.24\textwidth}
    \centering
    \includegraphics[width=0.99\columnwidth]{figures/main_plots/Defeat_Zombie10_seeds.pdf}
    \caption{Defeat Zombie.}
    \label{fig:main-defeat-skeleton}
  \end{subfigure}\hfill
  \begin{subfigure}[c]{0.24\textwidth}
    \centering
    \includegraphics[width=0.99\columnwidth]{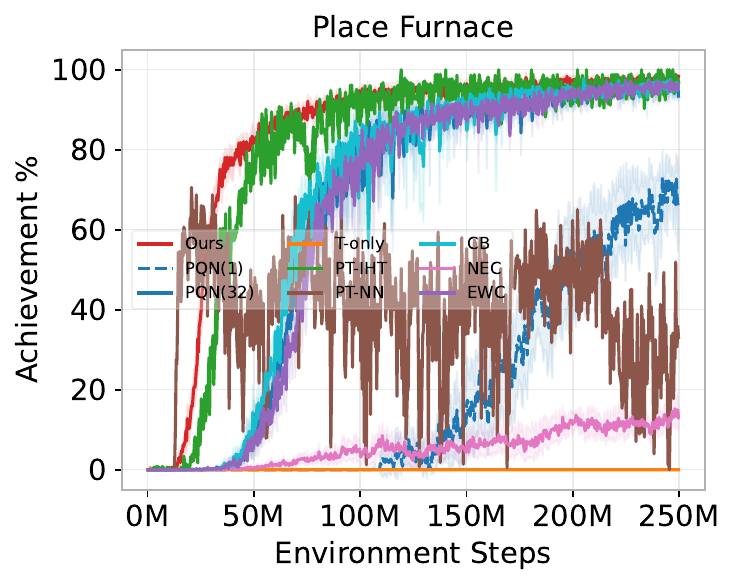}
    \caption{Place Furnace.}
    \label{fig:main-place-furnace}
  \end{subfigure}

    \begin{subfigure}[c]{0.24\textwidth}
    \centering
    \includegraphics[width=0.99\columnwidth]{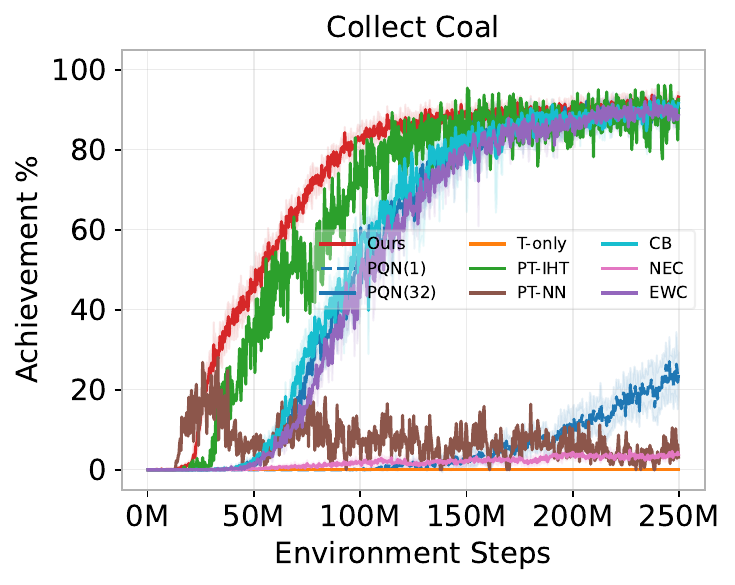}
    \caption{Collect Coal.}
    \label{fig:main-collect-coal}
  \end{subfigure}\hfill
  \begin{subfigure}[c]{0.24\textwidth}
    \centering
    \includegraphics[width=0.99\columnwidth]{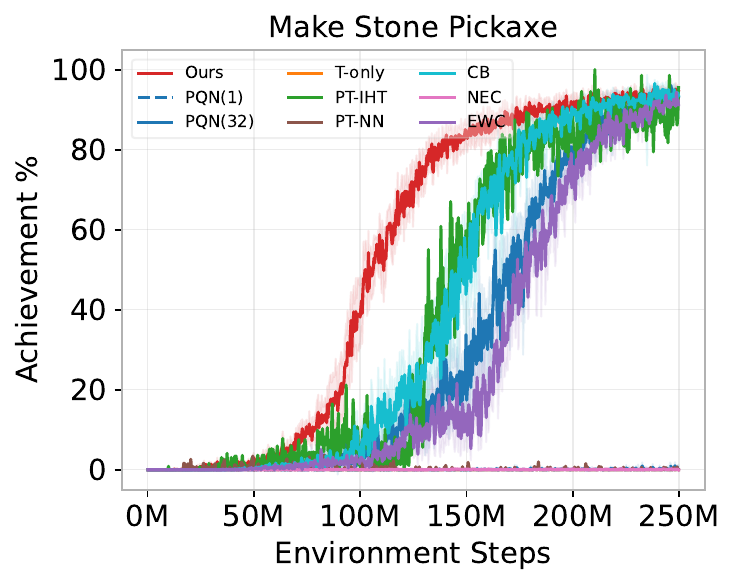}
    \caption{Make Stone Pickaxe.}
    \label{fig:main-make-stone-pickaxe}
  \end{subfigure}\hfill
  \begin{subfigure}[c]{0.24\textwidth}
    \centering
    \includegraphics[width=0.99\columnwidth]{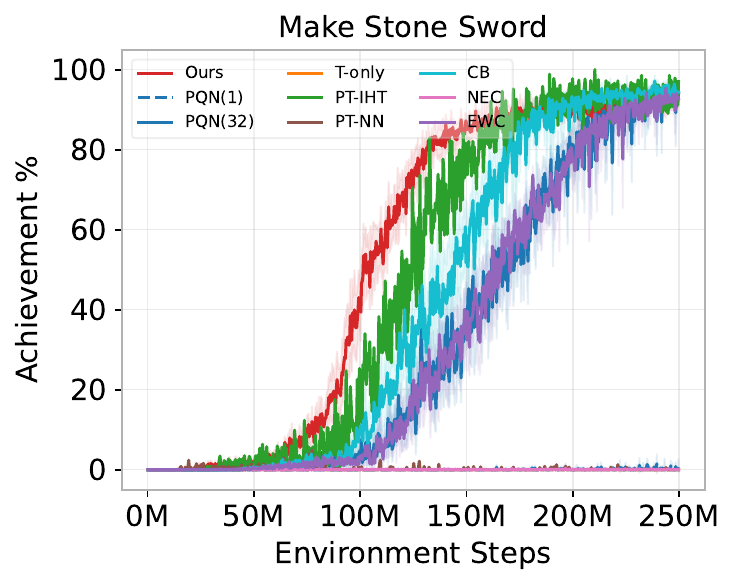}
    \caption{Make Stone Sword.}
    \label{fig:main-stone-sword}
  \end{subfigure}\hfill
  \begin{subfigure}[c]{0.24\textwidth}
    \centering
    \includegraphics[width=0.99\columnwidth]{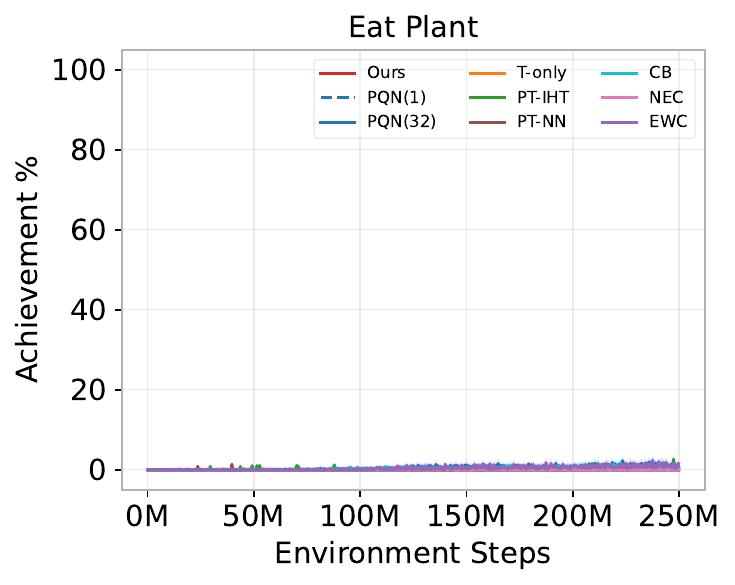}
    \caption{Eat Plant.}
    \label{fig:main-eat-plant}
  \end{subfigure}

    \begin{subfigure}[c]{0.24\textwidth}
    \centering
    \includegraphics[width=0.99\columnwidth]{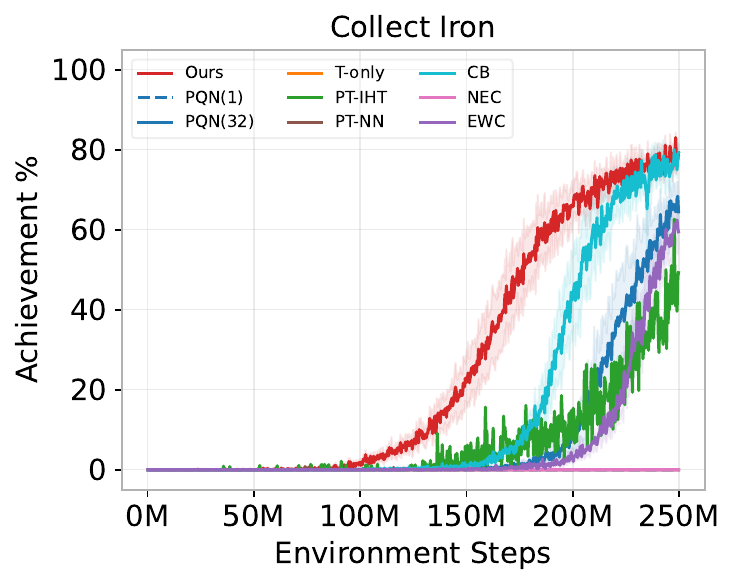}
    \caption{Collect Iron.}
    \label{fig:main-collect-iron}
  \end{subfigure}\hfill
  \begin{subfigure}[c]{0.24\textwidth}
    \centering
    \includegraphics[width=0.99\columnwidth]{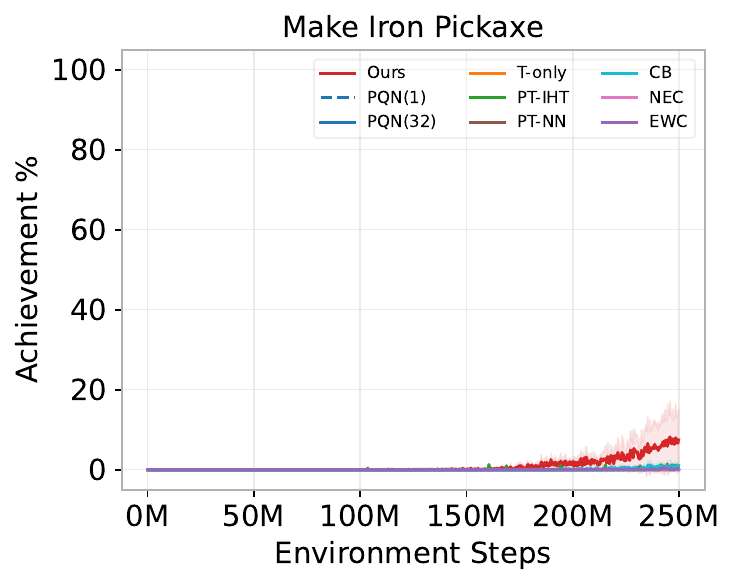}
    \caption{Make Iron Pickaxe.}
    \label{fig:main-make-iron-pickaxe}
  \end{subfigure}\hfill
  \begin{subfigure}[c]{0.24\textwidth}
    \centering
    \includegraphics[width=0.99\columnwidth]{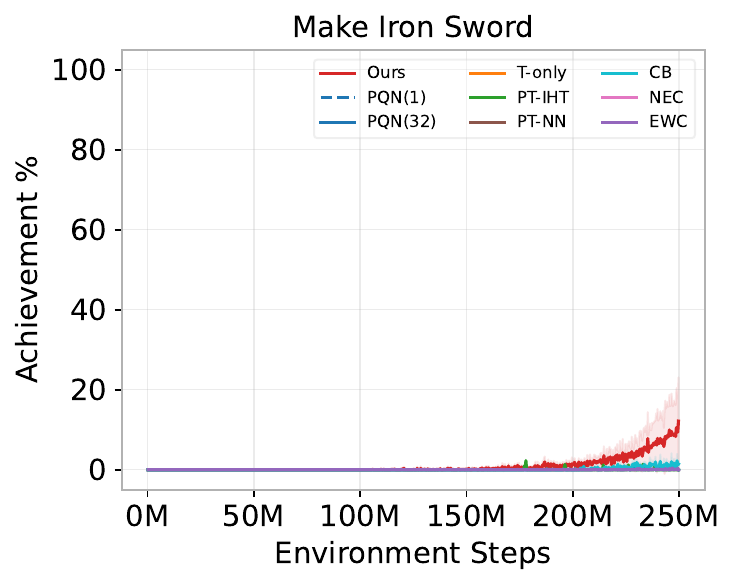}
    \caption{Make Iron Sword.}
    \label{fig:main-iron-sword}
  \end{subfigure}\hfill
  \begin{subfigure}[c]{0.24\textwidth}
    \centering
    \includegraphics[width=0.99\columnwidth]{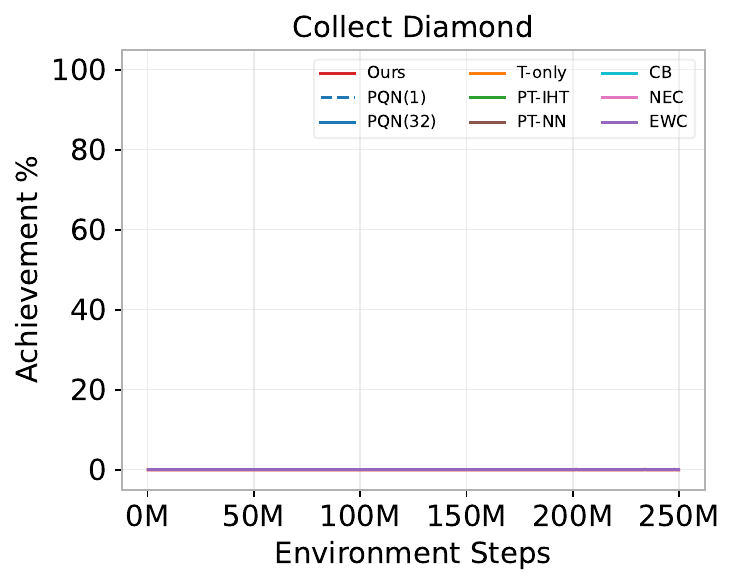}
    \caption{Collect Diamond.}
    \label{fig:main-collect-diamond}
  \end{subfigure}
  \caption{All achievements in the craftax 250M experiment.}
  \label{app-fig:main-achievements}
\end{figure}

\textbf{Results:} We used 3 seeds for the two-stream experiment and 10 seeds for the $250$M benchmarking. We include the main results here and present the parameter ablation and per-achievement results in the Appendix~\ref{app-fig:full-ablation-grid}. As shown in Fig.~\ref{fig:online-craftax-results}, Fig.~\ref{fig:250M-craftax-results}, and Fig.~\ref{app-fig:main-achievements}, our approach successfully learns in the two-stream setting and outperforms all baselines in the extensive 250M benchmark. It surpasses a return of 10 in under $50$M steps (twice as fast as PQN(32)) and reaches a return of $15$ within $150$M steps (PQN(32) requires over $200$M steps). This success stems from the transient table's ability to compute precise, local updates while explicitly controlling generalization between states. This, in turn, accelerates the permanent component's ability to estimate baseline predictions and generalize broadly. In contrast, the baselines lacking this slot-based or system-level decomposition design struggle. PT-IHT degrades over time due to unfavourable collisions, while T-only inherently lacks long-term capacity. The fully neural network approaches like PT-NN and PQN fail or learn without large batches or environment parallelization~\citep{mnih2013playing, elsayed2024streaming}, and their reliance on minibatching to stabilize updates risks introducing primacy bias~\citep{nikishin2022primacy, d2022sample}. The performance indifference between the EWC and CB baselines indicates that the mechanistic approaches maintain the neural network's stability and plasticity, but they don't enable the rapid adaptation needed for CRL.

\section{Discussion and Conclusion}
\label{sec:discussion}

In this work, we extended and enhanced the PT framework by incorporating slowly evolving or static features, either hand-crafted (e.g., Fourier bases) or learned using neural networks, for the permanent component, alongside reward-predictive, fast-evolving, or non-parametric features for the transient component to improve performance in CRL. The backbone of our approach to scale is a novel MinHash-based non-parametric approximator that enables rapid online learning like tabular RL, local generalization like CMACs and tile coding (but controlled via a hyperparameter, $\rho$), and efficient storage and retrieval of observations and values, all while remaining modest in size relative to the complexity of the environment.

Our theoretical results in the linear case are a significant improvement over the previous tabular results. These results demonstrate that using two distinct feature sets does not lead to divergence, as is common in many linear RL scenarios. We expect our proofs to play a foundational role in developing stronger theoretical results in the future, such as improved adaptation speed, reduced interference during permanent updates, or stability-plasticity analysis for CRL.

The new non-parametric approximator introduces several hyperparameters. The table size, number of tables, and number of hashes determine the approximator's capacity; in general, the more, the better, analogous to the number of layers and hidden units in NNs. Although the remaining HPs can be tuned either coordinate-wise (Craftax; a scalable, practical approach) or over all combinations (Image Task), in the future, we expect meta-learning to automate parameter tuning~\citep{xu2018meta}, or newer, simpler architectures for transient approximations. We also expect our non-parametric approximator to extend to other cases, such as associative memory and retrieval.

While tokenization is natural for symbolic observations, we quantized grayscale images derived from their RGB counterparts for the image tasks. To further improve this, advanced tokenization strategies from prior work on foundation models can be leveraged~\citep{dosovitskiy2020image, radford2021learning, luo2023survey}. Extending our approach to policy gradient algorithms, developing mechanisms for selective consolidation (determining when and what to transfer to the permanent), combining mechanistic approaches with ours to improve the permanent, and leveraging our framework in streaming RL are the immediate next steps.

Ultimately, an RL agent’s ability to continually learn from new experiences is crucial for advancing our scientific understanding of intelligence. Our framework takes a step towards this goal by demonstrating that maintaining separate representations is critical for balancing stability with plasticity and enabling rapid adaptation. By grounding our architecture in the CLS theory, we help bridge the gap between biological and artificial learning paradigms, providing yet another example of the fruitful exchange of ideas between neuroscience and RL. 

\newpage
\begin{ack}
We are grateful to NSERC and CIFAR grants for funding this research. We thank Aditya Mahajan, David Abel, Ivan Anokhin, and colleagues at Mila for helpful discussions, and the anonymous reviewers for constructive feedback. We also thank the Mila compute team for their technical support and assistance with computational infrastructure. This research was enabled in part by computing resources provided by Mila (mila.quebec), Calcul Quebec (calculquebec.ca), and the Digital Research Alliance of Canada (alliancecan.ca).
\end{ack}

\bibliographystyle{abbrvnat}
\bibliography{neurips_2026}

\newpage
\setcounter{section}{0}
\renewcommand{\thesection}{\Alph{section}}
\section{Appendix.}

\subsection{Theoretical Results}
\label{app-sec:proofs}

In this section, we study the convergence of permanent and transient value function updates when these are based on different representations, by leveraging the two-timescale proof technique introduced by Borkar (see Appendix~\ref{thm:borkar-two})~\citep{borkar1997stochastic}. While the approach is general and broadly applicable for various stochastic approximation algorithms, the conditions can be simplified for RL as shown by \cite{bertsekas1996neuro, tsitsiklis1996analysis} for a single iteration. In particular, the Lipschitz assumption is satisfied by showing that the expected update in matrix form (i.e. the key matrix) is well-defined and positive definite. Also, the final assumption is satisfied by first showing that the noise terms form \emph{martingale difference sequence} with zero mean and bounded variance.

\begin{theorem}[Borkar, 1997]
        \label{thm:borkar-two}
        Consider two $d$ and $l$ dimensional coupled iterates of the form:
    \begin{align}
        \label{Eq:twotimescale-iterates}
        \PMparams_{t+1} &\gets \PMparams_t + \PMlr \big(\gA(\PMparams_t, \TMparams_t) + \gM_{t+1}\big), \\
        \TMparams_{t+1} &\gets \TMparams_t + \TMlr \big(\gC(\PMparams_t, \TMparams_t) + \gN_{t+1}\big),
    \end{align}
    for $t\geq0$. If,
    \begin{enumerate}
        \item $\gA: \sR^{d+l} \to \sR^{d}$, $\gC: \sR^{d+l} \to \sR^{l}$ are Lipschitz,
        \item $\sum_{t}{\PMlr_t} = \sum_{t}{\TMlr_t} = \infty$, $\sum_{t}{\PMlr_t^2} = \sum_{t}{\TMlr_t^2} < \infty$, $\lim_{t \to \infty} \tfrac{\PMlr_t}{\TMlr_t} \to 0$,
        \item for $\gF_t \triangleq \sigma(\PMparams_k, \TMparams_k, \gM_k, \gN_k, k \leq t)$, $t \geq 0$, $(\gM_t, \gF_t)$, $(\gN_t, \gF_t)$ are sequences of random variables satisfying: $\sum_t \PMlr_t \gM_t$, $\sum_t \TMlr_t \gN_t \leq \infty$ almost surely,
    \end{enumerate}
    then the iterates converge converge almost surely to the asymptotically stable equilibria of the associated limiting ODEs.
\end{theorem}

\begin{theorem}[Tsitsiklis \& Van Roy (1996)]
\label{thm:tsitsiklis}
    Consider an iterative algorithm of the form $\TMparams_{t+1} = \TMparams_t + \TMlr_t (\gA(X_t) \TMparams_t + \gb(X_t))$ where,
\begin{enumerate}
    \item the step-size sequence $\TMlr_t$ satisfies $\sum_{t}{\TMlr_t} = \infty$, $\sum_{t}{\TMlr_t^2} < \infty$,
    \item $X_t$ is a Markov process with a unique invariant distribution,
    \item $\gA(\cdot)$ and $\gb(\cdot)$ are matrix and vector valued functions respectively, for which $\keyA = \Edis[\gA(X_t)]$ and $\keyb = \Edis[\gb(X_t)]$ are well defined and finite,
    \item the matrix $\keyA$ is positive definite,
    \item there exists constants $K$ and $q$ such that for all $X$
    \begin{itemize}
        \item $\sum_{t=0}^{\infty} \norm{\Ep[\gA(X_t) | X_0 = X] - \keyA} \leq K (1 + h^q(X))$, and
        \item $\sum_{t=0}^{\infty} \norm{\Ep[\gb(X_t) | X_0 = X] - \keyb} \leq K (1 + h^q(X))$,
    \end{itemize}
    \item for any $q > 1$ there exists a constant $\mu_q$ such that for all $X, t$
    \begin{itemize}
        \item $\Ep[h^q(X_t)|X_0=X] \leq \mu_q (1 + h^q(X)).$
    \end{itemize}
\end{enumerate}
Then, $\TMparams_t$ converges to $\TMparams_\pol$, with probability one, where $\TMparams_\pol$ is the unique vector that satisfies $\keyA \TMparams_\pol + \keyb = 0$.
\end{theorem}

\begin{lemma}
\label{lemma:tm-convergence}
For any fixed choice of $\PMparams$, for any task $\task$ in the task distribution, the sequence of expected transient updates converges to a unique fixed point.
\end{lemma}

\begin{proof}
We use the proof technique outlined in \cite{anand2021preferential, tsitsiklis1996analysis} to establish convergence.

\textbf{Transient update as linear stochastic approximation.} Dropping the $(\task, \pol)$ subscript for clarity, the transient update can be written as
\begin{align*}
    \TMparams_{t+1} 
    &= \TMparams_t + \TMlr_t \Big(R_{t+1} + \gamma (\PMparams^T \PMfeat_{t+1} + \TMparams_t^T \TMfeat_{t+1}) 
                                 - (\PMparams^T \PMfeat_{t} + \TMparams_t^T \TMfeat_{t})\Big)\TMfeat_{t} \\
    &= \TMparams_t + \TMlr_t\big(b(X_t) + A(X_t)\TMparams_t\big),
\end{align*}
where
\[
b(X_t) = \TMfeat_{t}(R_{t+1} + \gamma \PMparams^T \PMfeat_{t+1} - \PMparams^T \PMfeat_t),
\quad
A(X_t) = \TMfeat_t(\TMfeat_t - \gamma \TMfeat_{t+1})^T,
\]
and $X_t = (S_t, S_{t+1}, \PMfeat_t, \TMfeat_t)$.

\textbf{Limiting expectations.} Define
\[
\keyA = \lim_{t \to \infty} \E[A(X_t)], 
\qquad
\keyb = \lim_{t \to \infty} \E[b(X_t)].
\]
Explicitly,
\begin{align*}
    \keyA &= \sum_{s} d(s)\,\TMfeat(s)\Big(\TMfeat(s) - \gamma \sum_{s'}[\p]_{ss'}\TMfeat(s')\Big)^T 
          = \boxed{\TMFeatMatrix^T D (I - \gamma \p)\TMFeatMatrix}, \\
    \keyb &= \sum_{s} d(s)\,\TMfeat(s)\Big(\rew(s) + \gamma \PMparams^T \sum_{s'}[\p]_{ss'}\PMfeat(s') - \PMparams^T\PMfeat(s)\Big) \\
          &= \boxed{\TMFeatMatrix^T D \rew + \TMFeatMatrix^T D (\gamma \p - I)\PMFeatMatrix \PMparams}.
\end{align*}

$\keyA$ is positive definite because $D(I-\gamma \p)$ has positive row sums (since $\gamma \p$ is a sub-stochastic matrix) and column sums (since $\mathbf{1}^T D (I - \gamma \p) = d^T - \gamma d^T = (1 - \gamma) d^T > 0$).

\textbf{Bounded noise.}  
Using the mixing of the Markov chain,
\begin{align*} 
    \norm{\E[A(X_t) | X_0] - \keyA} &= \norm{\TMFeatMatrix^T D_t (I - \gamma \p) \TMFeatMatrix - \TMFeatMatrix^T D (I - \gamma \p) \TMFeatMatrix} \\
    &= \norm{\TMFeatMatrix^T (D_t - D) (I - \gamma \p) \TMFeatMatrix} \\
    &\leq \norm{\TMFeatMatrix^T} \, \norm{D_t - D} \, \norm{I - \gamma \p} \, \norm{\TMFeatMatrix} \\
    &\leq B_1 \cdot C \sigma^t \cdot B_2 \cdot B_1 \\
    &\boxed{\leq K_1 \sigma^t}, \\ 
    \norm{\E[b(X_t) | X_0] - \keyb} &= \norm{\TMFeatMatrix^T (D_t - D) \rew + \TMFeatMatrix^T (D_t - D) (\gamma \p - I) \PMFeatMatrix \PMparams} \\
    &\leq \norm{\TMFeatMatrix^T} \, \norm{D_t - D} \, \norm{\rew} + \norm{\TMFeatMatrix^T} \, \norm{D_t - D} \, \norm{(\gamma \p - I)} \, \norm{\PMFeatMatrix \PMparams} \\
    &\boxed{\leq K_2 \sigma^t}.
\end{align*}
Therefore,
\begin{align*}
    \sum_{t=0}^{\infty} \norm{\E[A(X_t) | X_0] - \keyA} &\leq \sum_{t=0}^{\infty} K_1 \sigma^t = \boxed{\frac{K_1}{1-\sigma} = \bar{K}_1}, \\
    \sum_{t=0}^{\infty} \norm{\E[b(X_t) | X_0] - \keyb} &\leq \sum_{t=0}^{\infty} K_2 \sigma^t = \boxed{\frac{K_2}{1-\sigma} = \bar{K}_2}.
\end{align*} 

\textbf{Fixed point.}  
Thus, the expected iterates converge to the unique fixed point
\begin{align*}
    \TMparams_{\tau}^{\ast} &= \keyA^{-1} \keyb \\
    &= {\TMFeatMatrix^T D (I - \gamma \p) \TMFeatMatrix)}^{-1} (\TMFeatMatrix^T D \rew + \TMFeatMatrix^T D (\gamma \p - I) \PMFeatMatrix \PMparams) \\
    &= {\TMFeatMatrix^T D (I - \gamma \p) \TMFeatMatrix)}^{-1} \TMFeatMatrix^T D \rew + {\TMFeatMatrix^T D (I - \gamma \p) \TMFeatMatrix)}^{-1} (\TMFeatMatrix^T D (\gamma \p - I) \PMpred_{\PMparams}) \\
    &= \boxed{\w^{(TD)}_{\TMFeatMatrix, \tau} - \left(\TMFeatMatrix^T \Dtask (I- \gamma \ppoltask)\TMFeatMatrix \right)^{-1} \TMFeatMatrix^T \Dtask (I - \gamma \ppoltask) \PMFeatMatrix \PMparams^{\ast}}
\end{align*}
\end{proof}

\begin{lemma}
\label{lemma:pm-convergence}
The sequence of expected permanent updates converges to a unique fixed point.
\end{lemma}

\begin{proof}
\textbf{Permanent update with transient fixed point.}  
Since permanent updates evolve on a slower timescale, we may treat the transient parameters as converged.  
The update is
\begin{align*}
    \PMparams_{t+1} &= \PMparams_t + \PMlr_t\,C(X_t), 
    \quad\text{where } C(X_t) = \TMparams_t^T \TMfeat_t \PMfeat_t.
\end{align*}
Define
\begin{align*}
    \keyC &= \sum_\task p(\task) \sum_{s} d_{\task}(s) \TMparams_{\task}^T \TMfeat(s) \PMfeat(s) \\
    &= \boxed{\sum_\task p(\task) \PMFeatMatrix^T \Dtask \TMFeatMatrix \TMparams^*_\task}
\end{align*}

\textbf{Boundedness and Lipschitz condition.}  
For each task $\task$,
\[
\keyC_\task = \PMFeatMatrix^T \Dtask \TMFeatMatrix \TMparams^*_\task,
\]
and using $\TMparams^*_\task = \keyA_\task^{-1}\big(\TMFeatMatrix^T \Dtask \rew_\task + \TMFeatMatrix^T \Dtask(\gamma\p_\task-I)\PMFeatMatrix\PMparams\big)$,  
we obtain
\begin{align*}
    \norm{\keyC_\task} &= \norm{\PMFeatMatrix^T \Dtask \TMFeatMatrix \TMparams^*_\task} \\
    &= \norm{\PMFeatMatrix^T \Dtask \TMFeatMatrix {\keyA_\task}^{-1} (\TMFeatMatrix^T \Dtask \rew_\task + \TMFeatMatrix^T \Dtask (\gamma \p_\task - I) \PMFeatMatrix \PMparams)} \\
    &\leq \norm{\PMFeatMatrix^T \Dtask \TMFeatMatrix {\keyA_\task}^{-1} \TMFeatMatrix^T \Dtask} +  \norm{(\rew_\task + (\gamma \p_\task - I) \PMFeatMatrix \PMparams)} \\
    &\leq K_3, \\
    \norm{\keyC} &= \norm{\sum_\task p(\task) \PMFeatMatrix^T \Dtask \TMFeatMatrix \TMparams_\task} \\
    &\leq \sum_\task \norm{\PMFeatMatrix^T \Dtask \TMFeatMatrix \TMparams_\task} \\
    &\leq K_3.
\end{align*}
Thus, the mapping is bounded and Lipschitz.

\textbf{Noise boundedness.}  
Because tasks are sampled i.i.d. and each task’s Markov chain is rapidly mixing, the noise terms have finite variance.

\textbf{Fixed point.}  
Therefore, the expected permanent updates converge to the unique fixed point:
\begin{align*}
    \keyC &= \sum_\task p(\task) \PMFeatMatrix^T \Dtask \TMFeatMatrix \TMparams^*_\task \\
    0 &= \sum_\task p(\task) \Big(\PMFeatMatrix^T \Dtask \TMFeatMatrix v^{(TD)}_{\TMFeatMatrix, \tau} -  \PMFeatMatrix^T \Dtask  \TMFeatMatrix \left(\TMFeatMatrix^T \Dtask (I- \gamma \ppoltask)\TMFeatMatrix \right)^{-1} \TMFeatMatrix^T \Dtask (I - \gamma \ppoltask) \PMFeatMatrix \PMparams^{\ast} \Big)\\
    0 &= \Etask[\PMFeatMatrix^T \Dtask \TMFeatMatrix v^{(TD)}_{\TMFeatMatrix, \tau}] - \Etask[\PMFeatMatrix^T \Dtask \Pi_{\TMFeatMatrix, \task} \PMFeatMatrix \PMparams^*] \\
    \Etask[\PMFeatMatrix^T \Dtask \Pi_{\TMFeatMatrix, \task} \PMFeatMatrix] \PMparams^* &= \Etask[\PMFeatMatrix^T \Dtask \TMFeatMatrix v^{(TD)}_{\TMFeatMatrix, \tau}] \\
    \Aboxed{\PMparams^* &= \Etask[\PMFeatMatrix^T \Dtask \Pi_{\TMFeatMatrix, \task} \PMFeatMatrix]^{-1} \Etask[\PMFeatMatrix^T \Dtask \TMFeatMatrix v^{(TD)}_{\TMFeatMatrix, \tau}]}
\end{align*}
\end{proof}

\begin{corollary}
\label{app:corollary}
If $\TMFeatMatrix = \PMFeatMatrix$, then 
\begin{align*}
    \PMparams^* &= \Etask\!\left[\PMFeatMatrix^T \Dtask \PMFeatMatrix\right]^{-1} 
                   \Etask\!\left[\PMFeatMatrix^T \Dtask \PMFeatMatrix \w^{(TD)}_{\PMFeatMatrix, \tau}\right], \\
    \TMparams^{\ast}_\tau &= \w^{(TD)}_{\PMFeatMatrix, \tau} - \PMparams^{\ast}.
\end{align*}
Moreover, in the single-task setting, $\PMparams^* = \w^{(TD)}_{\PMFeatMatrix, \tau}$ and $\TMparams^{\ast}_\tau = 0$.
\end{corollary}

\begin{proof}
    When $\PMFeatMatrix = \TMFeatMatrix$,
    \begin{align*}
        \Pi_{\TMFeatMatrix, \tau} &= \Pi_{\PMFeatMatrix, \tau} = \PMFeatMatrix \left(\PMFeatMatrix^T \Dtask (I - \gamma \ppoltask)\PMFeatMatrix \right)^{-1} \PMFeatMatrix^T \Dtask (I - \gamma \ppoltask), \\
        \Pi_{\PMFeatMatrix, \tau} \PMFeatMatrix &= \PMFeatMatrix \left(\PMFeatMatrix^T \Dtask (I - \gamma \ppoltask)\PMFeatMatrix \right)^{-1} \left(\PMFeatMatrix^T \Dtask (I - \gamma \ppoltask) \PMFeatMatrix\right) = \PMFeatMatrix.
    \end{align*}
    Therefore,
    \begin{align*}
        \PMparams^* &= \Etask[\PMFeatMatrix^T \Dtask \PMFeatMatrix]^{-1} \Etask[\PMFeatMatrix^T \Dtask \PMFeatMatrix \w^{(TD)}_{\PMFeatMatrix}] \\
        \TMparams^{\ast}_\tau &= \w^{(TD)}_{\PMFeatMatrix, \tau} - \left(\PMFeatMatrix^T \Dtask (I- \gamma \ppoltask)\PMFeatMatrix \right)^{-1} \PMFeatMatrix^T \Dtask (I - \gamma \ppoltask) \PMFeatMatrix \PMparams^{\ast}, \\
        &= \w^{(TD)}_{\PMFeatMatrix, \tau} - \PMparams^{\ast}.
    \end{align*}
    In the single task setting,
    \begin{align*}
        \PMparams^* &= \Etask[\PMFeatMatrix^T \Dtask \PMFeatMatrix]^{-1} \Etask[\PMFeatMatrix^T \Dtask \PMFeatMatrix \w^{(TD)}_{\PMFeatMatrix}], \\
        &= \left(\PMFeatMatrix^T \Dtask \PMFeatMatrix\right)^{-1} \left(\PMFeatMatrix^T \Dtask \PMFeatMatrix\right) \w^{(TD)}_{\PMFeatMatrix, \tau} = \w^{(TD)}_{\PMFeatMatrix, \tau}, \\
         \TMparams^{\ast}_\tau &= \w^{(TD)}_{\PMFeatMatrix, \tau} - \PMparams^{\ast} = \w^{(TD)}_{\PMFeatMatrix, \tau} - \w^{(TD)}_{\PMFeatMatrix, \tau} = 0.
    \end{align*}
\end{proof}

\subsection{Small-Scale Experiments}
\label{app-sec:small-scale-experiments}

\begin{figure}[h]
    \centering
    \includegraphics[width=0.75\linewidth]{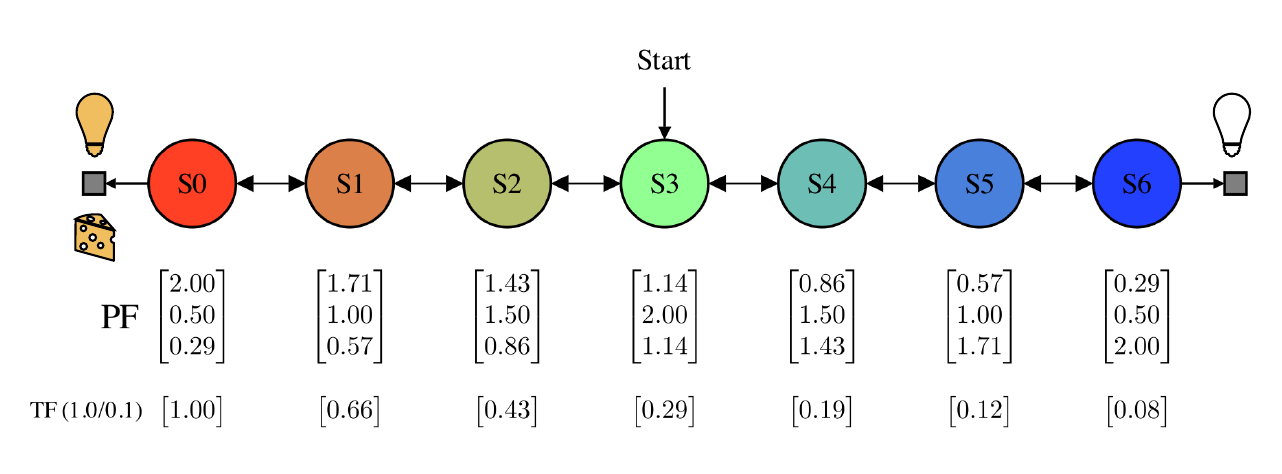}
    \caption{Chain MDP with the two feature sets}
    \label{app-fig:chain-mdp-with-features}
\end{figure}

\subsubsection{Pseudocode for Prediction and Control.}

\begin{algorithm}[H]
\caption{Prediction with Linear Approximations}
\label{algo:PT-linear-prediction}
\begin{algorithmic}[1]
    \State \textbf{Initialize:} buffer $\mathcal{B}$, parameters $\PMparams$, $\TMparams$
    \For{$t = 0 \to \infty$}
        \State Take action $A_t$
        \State Store state $S_t$ in $\mathcal{B}$
        \State Observe reward $R_{t+1}$ and next state $S_{t+1}$
        \State \Comment{Update transient parameters}
        \State $\TMparams_{t+1} \gets \TMparams_t 
            + \TMlr \big(R_{t+1} + \gamma \vpt(S_{t+1}) - \vpt(S_t)\big)\TMfeat(S_t)$
        \If{Task ends}
            \For{Every $S_k$ in $\mathcal{B}$}
                \State \Comment{Update permanent parameters}
                \State $\PMparams_{k+1} \gets \PMparams_k + \PMlr \left(\vpt(S_k) - \PMpred(S_k) \right) \PMfeat(S_k)$
            \EndFor
            \State \Comment{Reset transient parameters}
            \State $\TMparams_{t+1} \gets 0$
            \State \Comment{Clear buffer}
            \State Reset $\mathcal{B}$
        \EndIf
    \EndFor
\end{algorithmic}
\end{algorithm}

\begin{algorithm}[H]
\caption{Control with Linear Approximations}
\label{algo:PT-linear-control}
\begin{algorithmic}[1]
    \State \textbf{Initialize:} buffer $\mathcal{B}$, parameters $\PMparams$, $\TMparams$
    \For{$t = 0 \to \infty$}
        \State Take action $A_t$
        \State Store state $S_t$, $A_t$ in $\mathcal{B}$
        \State Observe reward $R_{t+1}$ and next state $S_{t+1}$
        \State \Comment{Update transient parameters}
        \State $\TMparams_{t+1} \gets \TMparams_t 
            + \TMlr \big(R_{t+1} + \gamma \max_{a'}\QPT(S_{t+1}, a') - \QPT(S_t, A_t)\big)\TMfeat(S_t, A_t)$ 
        \If{Task ends}
            \For{Every $(S_k, A_k)$ in $\mathcal{B}$}
                \State \Comment{Update permanent parameters}
                \State $\PMparams_{k+1} \gets \PMparams_k + \PMlr \left(\QPT(S_k, A_k) - \PMcontrol(S_k, A_k) \right) \PMfeat(S_k, A_k)$
            \EndFor
            \State \Comment{Reset transient parameters}
            \State $\TMparams_{t+1} \gets 0$
            \State \Comment{Clear buffer}
            \State Reset $\mathcal{B}$
        \EndIf
    \EndFor
\end{algorithmic}
\end{algorithm}

\subsubsection{Hyperparameters and Tasks}
\label{app-sec-hp-tuning-small-scale}

\noindent\textbf{Tasks Used in Experiments}

\begin{table}[ht]
\begin{minipage}[t]{0.54\textwidth}
    \centering
    \setlength{\tabcolsep}{12pt}
    \renewcommand{\arraystretch}{1.3}
    \rowcolors{2}{Alt}{white}
    \begin{tabular}{|>{\columncolor{Hdr}}c|c|c|c|c|}
    \hline
    \cellcolor{Hdr}\textbf{Task} & \cellcolor{Hdr}\textbf{G1} & \cellcolor{Hdr}\textbf{G2}
                                 & \cellcolor{Hdr}\textbf{G3} & \cellcolor{Hdr}\textbf{G4} \\
    \hline
    \cellcolor{Hdr}\textbf{1} & 0 & 1 & 0 & 1 \\
    \cellcolor{Hdr}\textbf{2} & 1 & 0 & 1 & 0 \\
    \cellcolor{Hdr}\textbf{3} & 0 & 0 & 1 & 1 \\
    \cellcolor{Hdr}\textbf{4} & 1 & 1 & 0 & 0 \\
    \hline
    \end{tabular}
    \vspace{10pt}
    \caption{Linear prediction tasks.}
    \label{tab:task-linear-pred}
\end{minipage}
\hfill
\begin{minipage}[t]{0.42\textwidth}
    \centering
    \setlength{\tabcolsep}{16pt}
    \renewcommand{\arraystretch}{1.35}
    \rowcolors{2}{Alt}{white}
    \begin{tabular}{|>{\columncolor{Hdr}}c|>{\centering\arraybackslash}m{2.6em}|>{\centering\arraybackslash}m{2.6em}|}
    \hline
    \cellcolor{Hdr}\textbf{Task} & \cellcolor{Hdr}\makecell{\reddot} & \cellcolor{Hdr}\makecell{\bluedot} \\
    \hline
    \cellcolor{Hdr}\textbf{1} & 1   & 0.1 \\
    \cellcolor{Hdr}\textbf{2} & 0.1 & 1   \\
    \hline
    \end{tabular}
    \vspace{10pt}
    \caption{Linear control tasks.}
    \label{tab:task-linear-control}
\end{minipage}
\end{table}

\noindent\textbf{Hyperparameter Sweeps for Linear Prediction}
\begin{Verbatim}[commandchars=\\\{\}]
TD-learning:
    LR = [3e-2, \textbf{1e-2}, 3e-3, 1e-3, 3e-4, 1e-4]

TD-learning (Reset):
    LR = [\textbf{3e-2}, 1e-2, 3e-3, 1e-3, 3e-4]

PT-TD (Full):
    LR-P = [1e-2, 3e-3, 1e-3, 3e-4, \textbf{1e-4}, 3e-5]
    LR-T = [1e-1, 3e-2, \textbf{1e-2}, 3e-3, 1e-3, 3e-4]

PT-TD (Separate):
    LR-P = [1e-2, 3e-3, 1e-3, \textbf{3e-4}, 1e-4, 3e-5]
    LR-T = [1e-1, \textbf{3e-2}, 1e-2, 3e-3, 1e-3, 3e-4]
\end{Verbatim}

\noindent\textbf{Hyperparameter Sweeps for Linear Control}
\begin{Verbatim}[commandchars=\\\{\}]
Q-learning:
    LR = [0.5, 0.3, \textbf{0.1}, 0.03, 0.01, 0.003, 0.001]

Q-learning (Reset):
    LR = [0.5, 0.3, \textbf{0.1}, 0.03, 0.01, 0.003, 0.001]

PT-Q (Full):
    LR-P = [\textbf{0.03}, 0.01, 0.003, 0.001, 0.0003]
    LR-T = [0.5, 0.3, \textbf{0.1}, 0.03, 0.01]

PT-Q (Separate):
    LR-P = [0.03, 0.01, 0.003, \textbf{0.001}, 0.0003]
    LR-T = [\textbf{0.5}, 0.3, 0.1, 0.03, 0.01]
\end{Verbatim}

\subsection{Non-parameteric Transient Approximations}
\label{app-sec:non-par-transient}
Here, we provide additional details on the non-parametric transient approximator, including pseudocode for several methods and a discussion of the roles of hyperparameters.

\subsubsection{Pseudocode}
The non-parametric transient approximator is implemented through three core methods: \emph{Put}, which stores incoming observations (see Algorithm~\ref{algo:put-method}); \emph{Get}, which retrieves stored value estimates (see Algorithm~\ref{algo:get-method}); and \emph{UpdateTDError}, which updates the estimates in proportion to their contribution (see Algorithm~\ref{algo:update-td-error-method}).

\begin{algorithm}[htb]
\caption{PUT Method}
\label{algo:put-method}
\begin{algorithmic}[1]
    \Function{put}{($\mathcal{M}$, $x$)}
        \Require $\mathcal{M}$: transient memory; $x$: observation
        \Ensure  Updated memory $\mathcal{M}$
        \State $(\textit{sig}, \textit{tag}) \gets \Call{Minhash}{x}$
        \State $\textit{bins} \gets \Call{GetBuckets}{\textit{sig}}$ 
        \State \Comment{Insert observation in all tables}
        \For{$t \gets 0 \ \mathbf{to}\ T-1$}
            \If{(\Call{ContainsTag}{$\mathcal{M}[t,\textit{bins}[t]]$, $\textit{tag}$} == \textbf{false})}
                \State \Comment{empty-first else LRU}
                \State $\textit{slot} \gets \Call{SelectSlot}{\mathcal{M}, t, \textit{bins}[t]}$
                \State $\mathcal{M}[t,\textit{bins}[t],\textit{slot}].\textit{tag} \gets \textit{tag}$
                \State \Comment{mean over other valid slots}
                \State $\mathcal{M}[t,\textit{bins}[t],\textit{slot}].\textit{values} \gets \Call{Initialize}{t, \textit{bins}[t]}$
            \Else
                \State $\textit{slot} \gets \Call{FindSlotByTag}{\mathcal{M}[t,\textit{bins}[t]], \textit{tag}}$
            \EndIf
            \State $\mathcal{M}[t,\textit{bins}[t],\textit{slot}].\textit{age} \gets \Call{Touch}{\mathcal{M}.\textit{clock}}$
        \EndFor
        \State $\mathcal{M}.\textit{clock} \gets \mathcal{M}.\textit{clock} + 1$
        \State \Return $\mathcal{M}$
    \EndFunction
\end{algorithmic}
\end{algorithm}

\begin{algorithm}[htb]
\caption{GET Method}
\label{algo:get-method}
\begin{algorithmic}[1]
    \Function{GET}{($\mathcal{M}$, $x$, $\rho$)}
        \Require $\mathcal{M}$: transient memory; $x$: observation; $\rho$: mixing weight
        \Ensure values
        \State $(\textit{sig}, \textit{tag}) \gets \Call{Minhash}{x}$
        \State bins $\gets$ \Call{GetBuckets}{sig}
        \State values $\gets 0$
        \State \Comment{Collect values from every table}
        \For{$t \gets 0\ \mathbf{to}\ T-1$}
            \State slot $\gets$ \Call{FindSlotByTag}{$\mathcal{M}[t,\textit{bins}[t]], \textit{tag}$}
            \State \Comment{Weigh the update for matching slot by $\rho$}
            \State values $\gets$ values $+ \rho \cdot \mathcal{M}[t, \textit{bins}[t], \textit{slot}].\textit{values}$
            \State OSlots $\gets$ \Call{FindOtherValidSlots}{$\mathcal{M}[t,\textit{bins}[t]], \textit{tag}$}
            \State $n \gets \Call{len}{\textit{OSlots}}$
            \State \Comment{Divide weight $(1-\rho)$ equally among other valid slots}
            \For{$s \gets 0\ \mathbf{to}\ n-1$}
                \State values $\gets$ values $+ \tfrac{(1-\rho)}{n} \cdot \mathcal{M}[t, \textit{bins}[t], \textit{OSlots}[s]].\textit{values}$
            \EndFor
        \EndFor
        \State \Return values
    \EndFunction
\end{algorithmic}
\end{algorithm}

\begin{algorithm}[htb]
\caption{UpdateTDError Method}
\label{algo:update-td-error-method}
\begin{algorithmic}[1]
    \Function{UpdateTDError}{($\mathcal{M}$, $x$, $a$, $\rho$, $\delta$, $\TMlr$)}
        \Require $\mathcal{M}$: transient memory; $x$: observation; $a$: action; $\rho$: mixing weight; $\delta$: TD-error; $\TMlr$: learning rate
        \Ensure Updated memory $\mathcal{M}$
        \State $(\textit{sig}, \textit{tag}) \gets \Call{Minhash}{x}$
        \State bins $\gets$ \Call{GetBuckets}{sig}
        \For{$t \gets 0\ \mathbf{to}\ T-1$}
            \State slot $\gets$ \Call{FindSlotByTag}{$\mathcal{M}[t,\textit{bins}[t]], \textit{tag}$}
            \State \Comment{Weight the matching slot by $\rho$}
            \State $\mathcal{M}[t,\textit{bins}[t],\textit{slot}].\textit{values}[a] \gets \mathcal{M}[t,\textit{bins}[t],\textit{slot}].\textit{values}[a] + \tfrac{\TMlr}{T} \rho \delta$
            \State OSlots $\gets$ \Call{FindOtherValidSlots}{$\mathcal{M}[t,\textit{bins}[t]], \textit{tag}$}
            \State $n \gets \Call{len}{\textit{OSlots}}$
            \State \Comment{Divide TD-error equally among other valid slots by weighting by $(1-\rho)$}
            \For{$s \gets 0\ \mathbf{to}\ n-1$}
                \State $\mathcal{M}[t,\textit{bins}[t],\textit{OSlots}[s]].\textit{values}[a] \gets \mathcal{M}[t,\textit{bins}[t],\textit{OSlots}[s]].\textit{values}[a] + \tfrac{\TMlr}{nT} (1 - \rho) \delta$
            \EndFor
        \EndFor
        \State \Return $\mathcal{M}$
    \EndFunction
\end{algorithmic}
\end{algorithm}

\subsubsection{Hyperparameters}
\label{app-sec:hp-strategy-transient}
\begin{enumerate}
    \item \textbf{Number of Tables ($T$) \& Generalization:} Similar to tiles in Tile Coding. More tables spread information, increasing generalization. Fewer tables concentrate information, reducing generalization.
    \item \textbf{Table Size ($B$) \& Capacity:} Determines the capacity of the approximator. The larger the table, the more observations it can store.
    \item \textbf{Slots ($S$) \& Eviction:} Determines capacity across bins. Too few slots lead to high eviction rates; too many slots approach tabular.
    \item \textbf{Generalization ($\rho$):} Explicitly controls the mixing of values between slots. $\rho=1$ forces isolation (no generalization); $\rho<1$ enables local smoothing.
    \item \textbf{Number of Hashes ($K$):} Determines the precision of the MinHash signature. A higher $K$ better preserves the Jaccard similarity property (improving retrieval accuracy), though with diminishing returns beyond a certain point.
\end{enumerate}

\textbf{Practical guide for choosing HPs:} Generally speaking, the larger the table, the better, as it can store many observations. When the number of slots is large, generalization is higher because many observations contribute to the values. So, the generalization parameter $\rho$ needs to be tuned carefully. For any environment, we recommend choosing the table size ($B$), number of slots ($S$), number of tables ($T$), and number of hashes ($K$). Then tune the remaining HPs (generalization parameter $\rho$, decay parameter $\lambda$, and transient learning rate $\TMlr$) using either coordinate or grid search.

Our use of MinHash introduces an inductive bias that yields desirable hash collisions. Its success is attributed to its property that preserves the Jaccard score in the hash signature, which is favourable for symbolic observations. In contrast, random hashing results in undesirable hash collisions, leading to incorrect transient corrections that, in turn, affect permanent learning.

\subsection{Large-Scale Experiments}
\label{app-sec:large-scale-experiments}

In this section, we provide baseline, HP tuning, and architecture details for the large-scale experiment, including the pseudocode.

\subsubsection{Image Task}
\label{app-sec:image-task}

\textbf{PQN.} PQN is a synchronous, JAX-based implementation of deep Q-learning designed for vectorized transitions across multiple parallel environments and periodically updating the network to minimize the 1-step Temporal Difference (TD) error using standard Q-learning error. We parameterize the Q-function by a neural network implemented in Flax. The input consists of flattened image observations (e.g., $32 \times 32 \times 3$). The network comprises two wide hidden layers of $1024$ units. Each hidden layer is followed sequentially by Layer Normalization and a ReLU activation function. The final dense linear layer outputs the Q-values for each discrete action in the environment. The HPs used in the final experiments are provided in Table~\ref{tab:pqn-hps-image}.

The network is optimized using the Adam optimizer. Gradients are clipped by global norm with a maximum threshold of $1.0$. Following hyperparameter tuning, the learning rate was set to $0.001$, and the data collected during environment steps was shuffled and divided into $4$ minibatches for the update step. The network undergoes $1$ epoch of training per update phase.

\begin{table}[htb]
\centering
\begin{tabular}{ll}
\toprule
\textbf{Parameter} & \textbf{Value} \\
\midrule
\multicolumn{2}{c}{\textbf{Hyperparameters}} \\
\midrule
ALG\_NAME & \texttt{PQN} \\
NUM\_MINIBATCHES & 4 \\
NUM\_EPOCHS & 1 \\
NORM\_INPUT & True \\
NORM\_TYPE & \texttt{layer\_norm} \\
LR & 0.001 \\
MAX\_GRAD\_NORM & 1.0 \\
\bottomrule
\end{tabular}
\vspace{5pt}
\caption{Hyperparameters for \texttt{PQN} for Image task.}
\label{tab:pqn-hps-image}
\end{table}

\textbf{PT-Ours.} PT-Ours represents our primary contribution, enhancing the Permanent-Transient value decomposition framework by replacing the neural network with novel, advanced, slot-based MinHash Locality Sensitive Hashing (LSH) non-parameteric approximations. By structuring the table into distinct slots, our approach successfully mitigates hash collisions and undesirable generalization that limit standard IHTs, enabling stable, high-fidelity transient value estimation.

The HPs used in the final experiments are provided in Table~\ref{tab:pt-minhash-hps}.

\begin{table}[htb]
\centering
\begin{tabular}{ll}
\toprule
\textbf{Parameter} & \textbf{Value} \\
\midrule
TRANSIENT\_TABLE\_SIZE & 128 \\
NUM\_TABLES & 2 \\
NUM\_HASHES & 128 \\
NUM\_SLOTS & 4 \\
TRANSIENT\_LR & 0.3 \\
DECAY & 0.95 \\
$\rho$ & 1.0 \\
\bottomrule
\end{tabular}
\vspace{5pt}
\caption{PT-Ours HPs.}
\label{tab:pt-minhash-hps}
\end{table}

\textbf{PT-NN.} PT-NN adapts the Permanent-Transient value decomposition framework but uses a neural network for estimating the transient value function. While the permanent network learns slowly over batches of data to ensure stability and generalization, the transient network is updated online at each transition to rapidly adapt to recent observations and minimize the immediate TD error. The permanent neural network retains HPs from PQN.

The HPs used in the final experiments are provided in Table~\ref{tab:PTQ(NeurIPS)-hps}.

\begin{table}[htb]
\centering
\begin{tabular}{ll}
\toprule
\textbf{Parameter} & \textbf{Value} \\
\midrule
TRANSIENT\_LR & 0.003 \\
DECAY & 1.0 \\
\bottomrule
\end{tabular}
\vspace{5pt}
\caption{PT-NN HPs.}
\label{tab:PTQ(NeurIPS)-hps}
\end{table}

\textbf{PT-IHT.} PT-IHT builds upon PQN architecture. For the slow-learning permanent, we use a neural network similar to PQN, and for the fast-learning transient non-parametric approximations, we use an Index Hash Table (IHT) with MinHash-based Locality Sensitive Hashing. While this allows rapid initial learning, the basic IHT implementation lacks advanced design choices, making the algorithm susceptible to hash collisions and to undesirable generalization across states.

The permanent neural network retains PQN's optimal HPs, and we tune only the table-based HPs. The final HPs used in the main results are presented in Table~\ref{tab:pt-iht-params}

\begin{table}[htb]
\centering
\begin{tabular}{ll}
\toprule
\textbf{Parameter} & \textbf{Value} \\
\midrule
TRANSIENT\_TABLE\_SIZE & 512 \\
NUM\_TABLES & 2 \\
NUM\_HASHES & 128 \\
TRANSIENT\_LR & 0.5 \\
DECAY & 1.0 \\
\bottomrule
\end{tabular}
\vspace{5pt}
\caption{PT-IHT HPs.}
\label{tab:pt-iht-params}
\end{table}

\textbf{T-Only.} T-Only serves as an ablation and a standalone baseline that entirely removes the permanent neural network. Instead, it uses our advanced slot-based approximator to estimate action-values. This algorithm updates purely online.

The HPs used in the final experiments are provided in Table~\ref{tab:tm-only-hps}.

\begin{table}[htb]
\centering
\begin{tabular}{ll}
\toprule
\textbf{Parameter} & \textbf{Value} \\
\midrule
TRANSIENT\_TABLE\_SIZE & 128 \\
NUM\_TABLES & 2 \\
NUM\_HASHES & 128 \\
NUM\_SLOTS & 4 \\
TRANSIENT\_LR & 0.3 \\
DECAY & 0.95 \\
$\rho$ & 1.0 \\
\bottomrule
\end{tabular}
\vspace{5pt}
\caption{T-Only HPs.}
\label{tab:tm-only-hps}
\end{table}

\textbf{EWC.} EWC is an extension of the base PQN algorithm that incorporates Elastic Weight Consolidation (EWC) to improve the stability of the neural network parameters during training. By explicitly regularizing the network weights against drastic shifts, EWC helps mitigate catastrophic interference between consecutive gradient updates, allowing the network to continuously adapt without destabilizing previously acquired value estimates.

The total loss is: $\mathcal{L}_{total} = \mathcal{L}_{TD} + \frac{\lambda}{2} \sum_i F_i (\theta_i - \theta_{i, old})^2$. The coefficient $\lambda$ controls the regularization strength and is set to $1.0$. The penalty restrains the current parameters ($\theta_i$) from deviating too far from a set of anchor parameters ($\theta_{i, old}$), with the constraint weighted by the diagonal elements of the empirical Fisher Information matrix ($F_i$).

This algorithm smoothly updates the regularization constraints online (every $32$ steps). After each update, both the Fisher Information matrix and the anchor parameters are updated using an Exponential Moving Average (EMA). The smoothing factors are configured to $\alpha_{fisher} = 0.99$ and $\alpha_{anchor} = 0.99$, respectively.

\textbf{CB.} CB integrates Continual Backprop into the standard PQN architecture to address the ``loss of plasticity'' commonly observed in neural networks. We stochastically re-initialize a small fraction of the network's hidden units, along with their associated optimizer statistics, every $32$ steps. This approach prevents the network from becoming overly rigid, enabling it to continuously adapt to non-stationary environments.

At the end of every $32$ steps, a subset of hidden units is randomly reset based on a Bernoulli distribution with a fixed, low probability ($p = 0.0001$).

\begin{figure}[ht]
    \centering
    \begin{subfigure}[b]{0.48\textwidth}
        \centering
        \includegraphics[width=\textwidth]{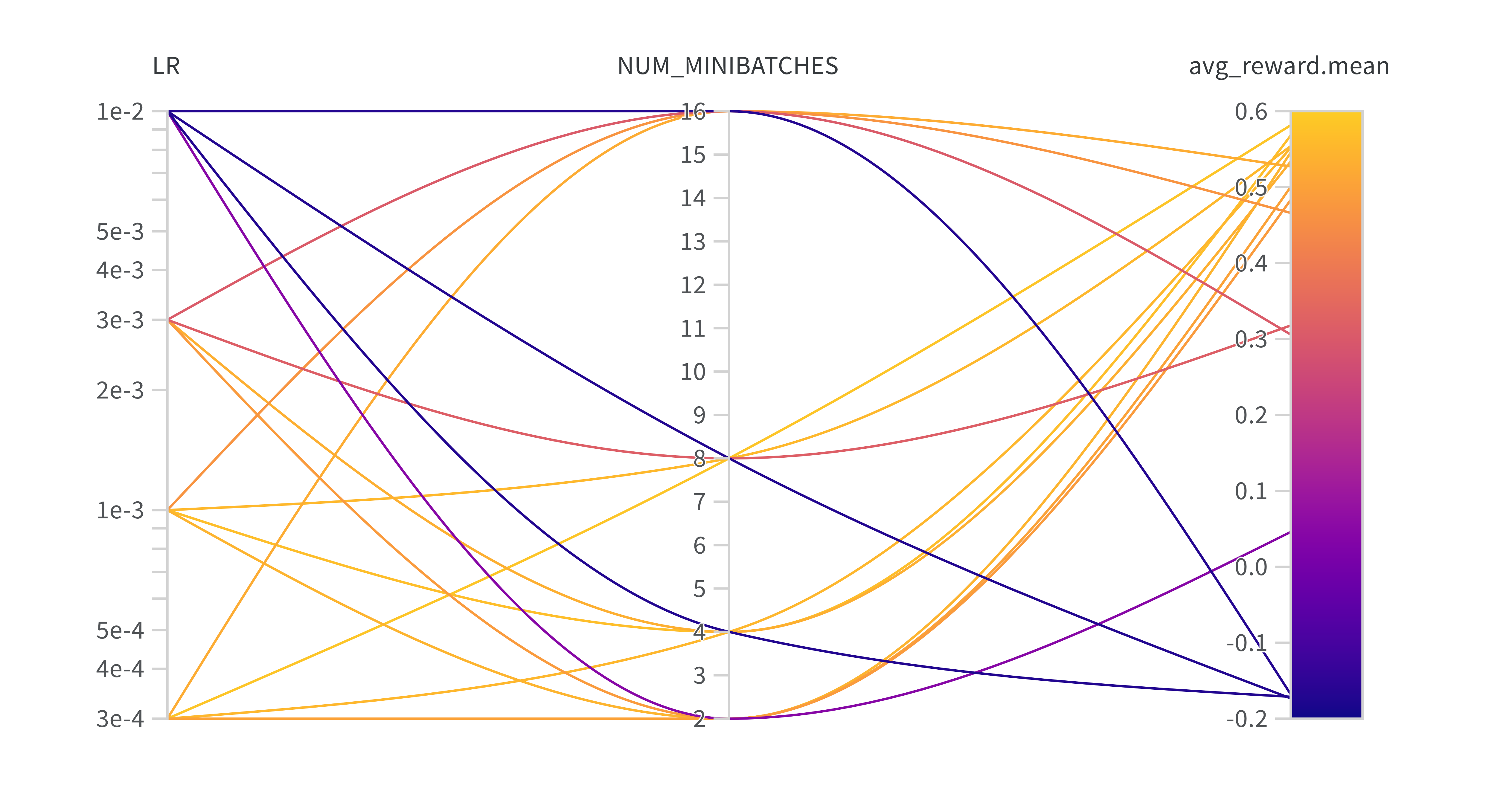}
        \caption{PQN}
        \label{fig:img-pqn-hp}
    \end{subfigure}
    \hfill
    \begin{subfigure}[b]{0.48\textwidth}
        \centering
        \includegraphics[width=\textwidth]{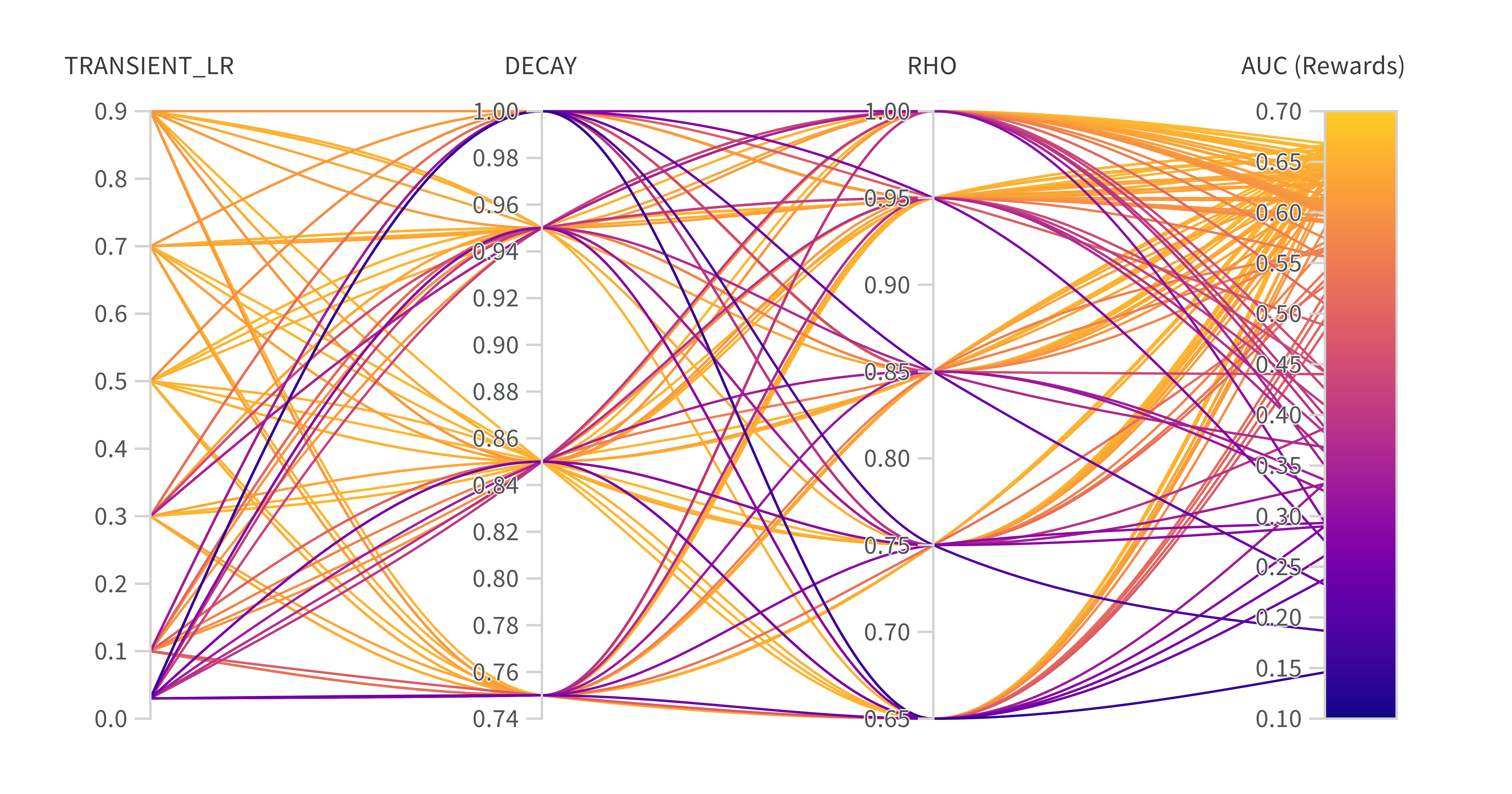}
        \caption{Ours}
        \label{fig:img-pt-ours-hp}
    \end{subfigure}
    \vspace{1em}
    \begin{subfigure}[b]{0.48\textwidth}
        \centering
        \includegraphics[width=\textwidth]{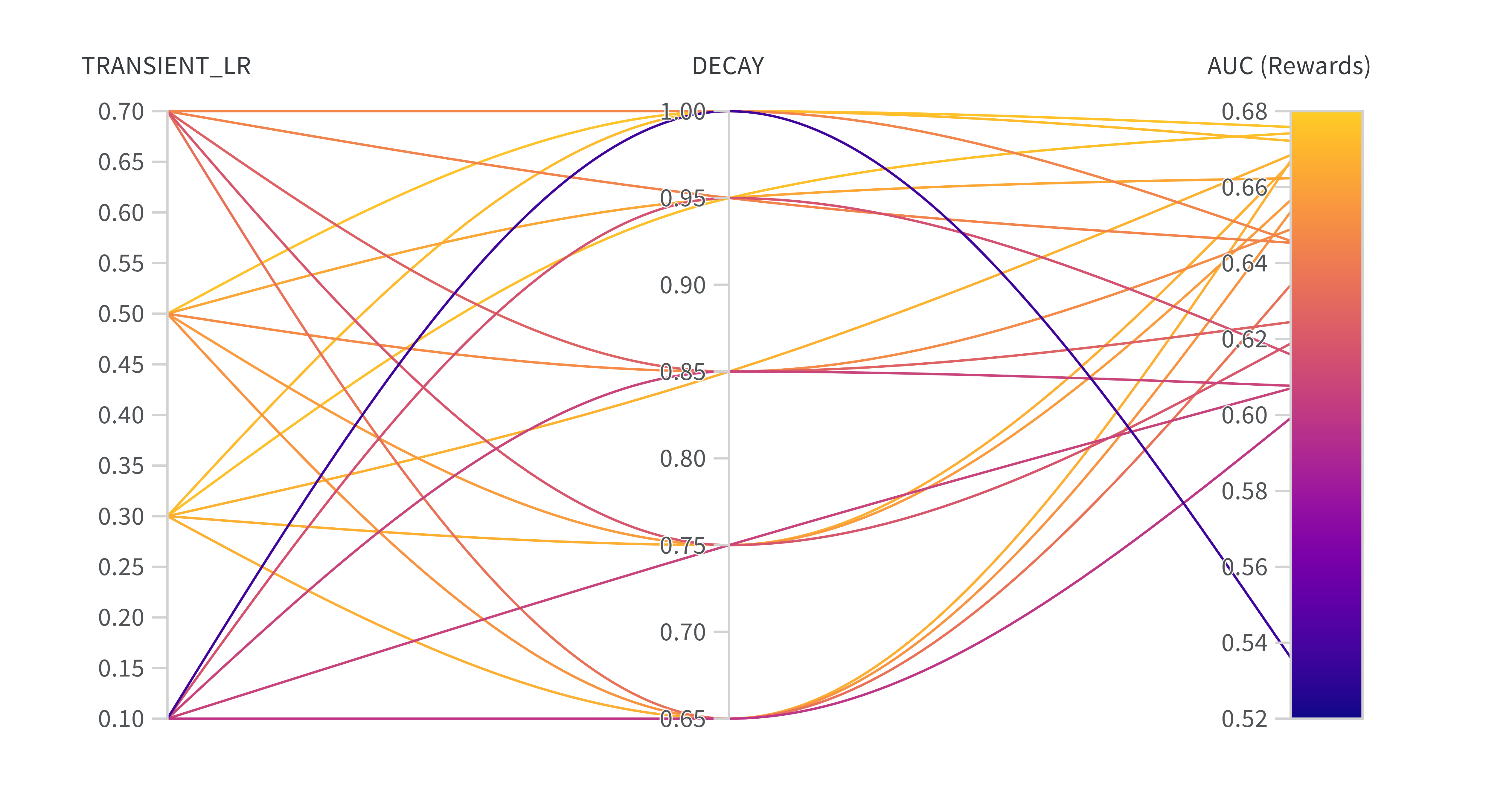}
        \caption{PT-IHT}
        \label{fig:img-pt-iht-hp}
    \end{subfigure}
    \hfill
    \begin{subfigure}[b]{0.48\textwidth}
        \centering
        \includegraphics[width=\textwidth]{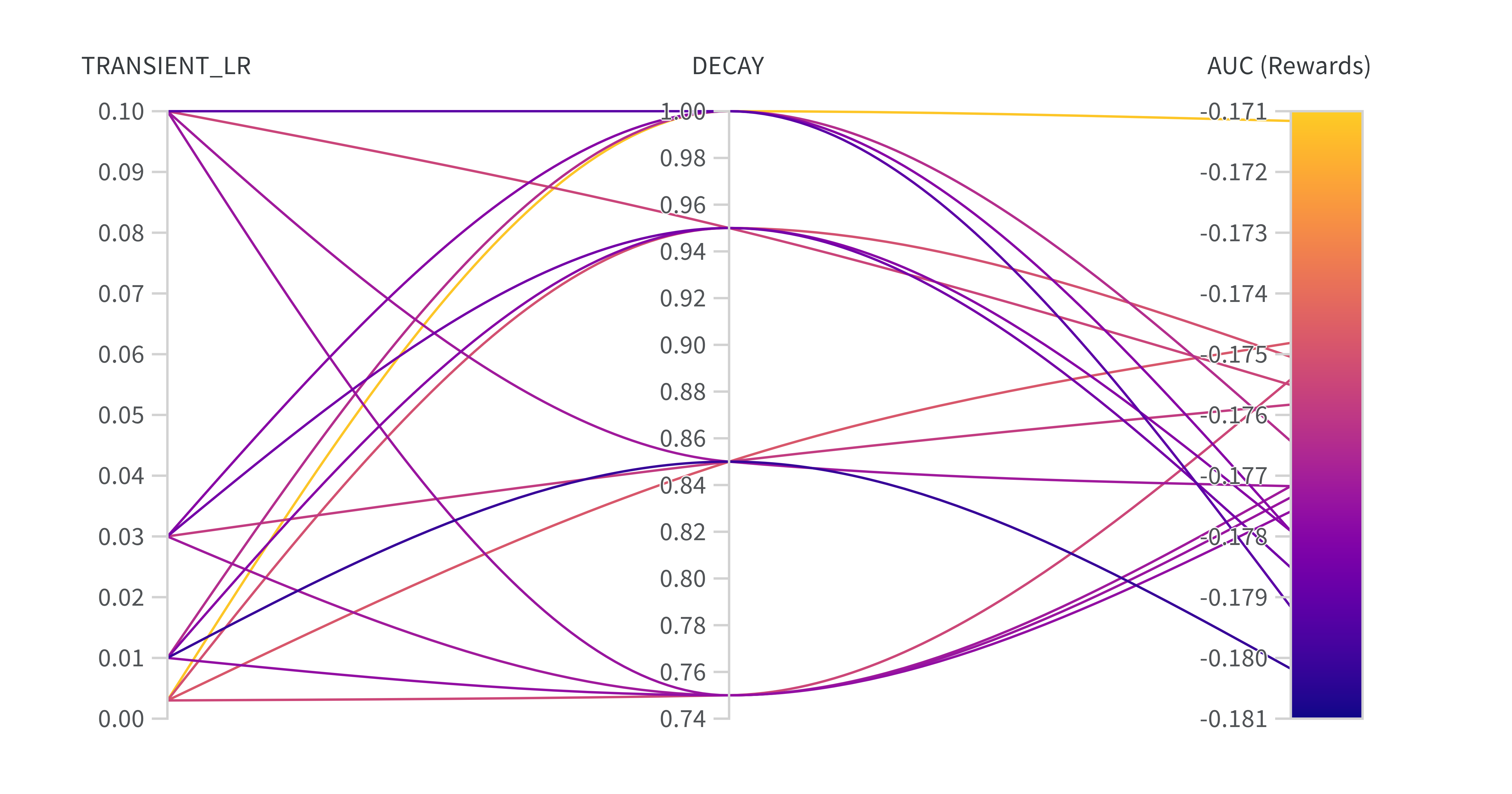}
        \caption{PT-NN}
        \label{fig:img-pt-nn-hp}
    \end{subfigure}
    \vspace{1em}
    \begin{subfigure}[b]{0.48\textwidth}
        \centering
        \includegraphics[width=\textwidth]{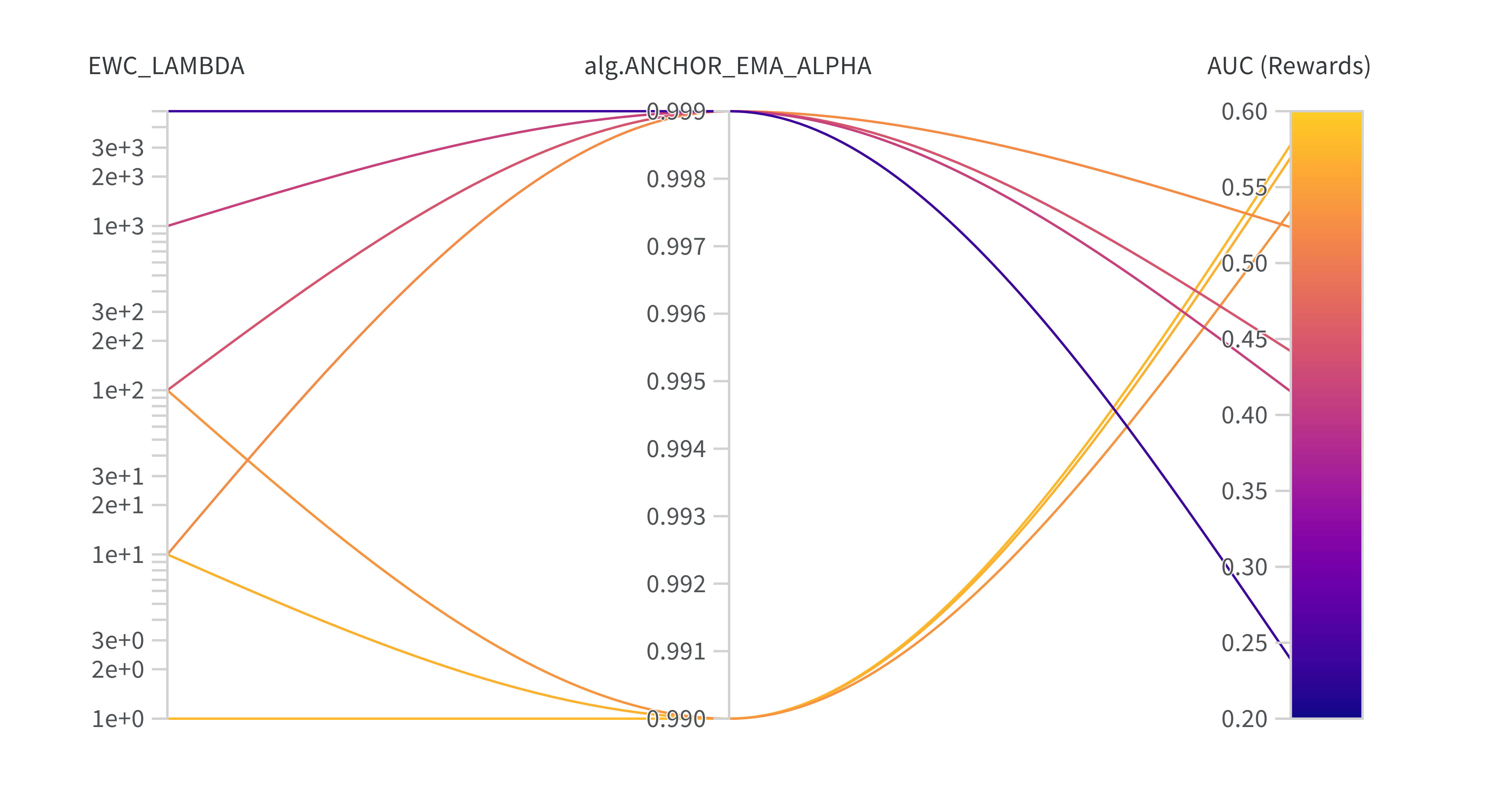}
        \caption{EWC}
        \label{fig:img-ewc-hp}
    \end{subfigure}
    \hfill
    \begin{subfigure}[b]{0.48\textwidth}
        \centering
        \includegraphics[width=\textwidth]{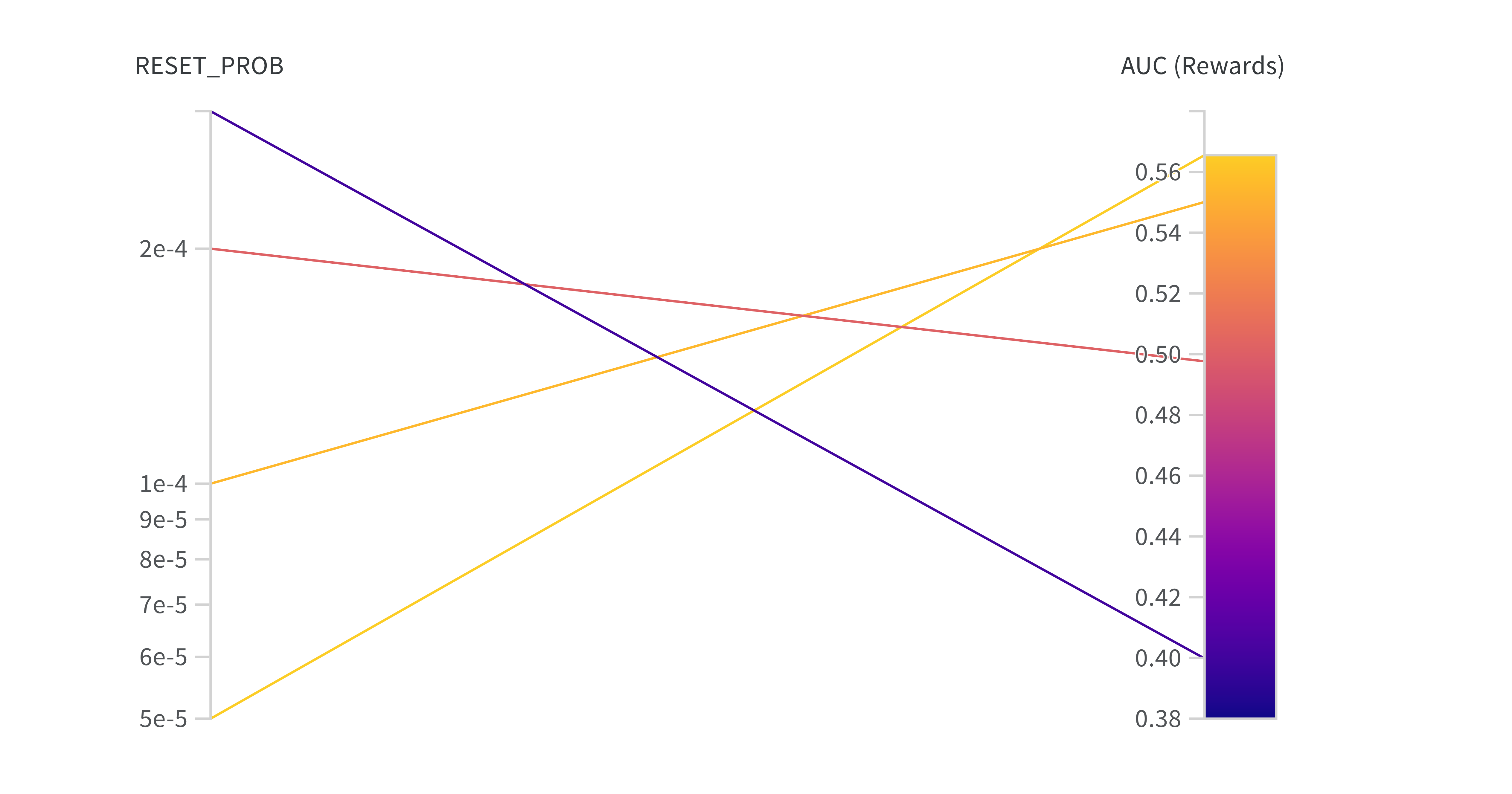}
        \caption{CB}
        \label{fig:img-cb-hp}
    \end{subfigure}
    \vspace{1em}
    \begin{subfigure}[b]{0.48\textwidth}
        \centering
        \includegraphics[width=\textwidth]{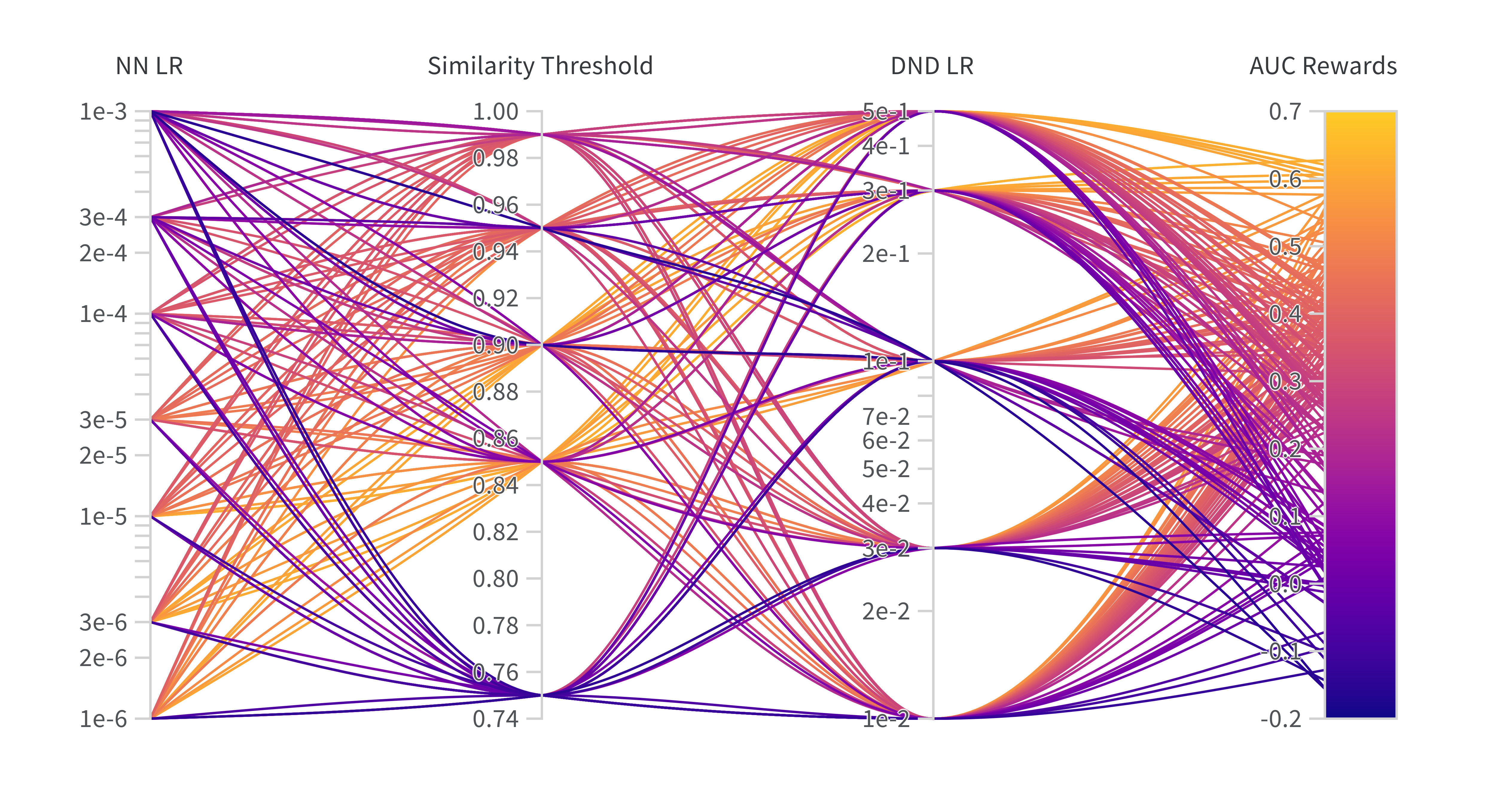}
        \caption{NEC}
        \label{fig:img-nec-hp}
    \end{subfigure}
    \hfill
    \begin{subfigure}[b]{0.48\textwidth}
        \centering
        \includegraphics[width=\textwidth]{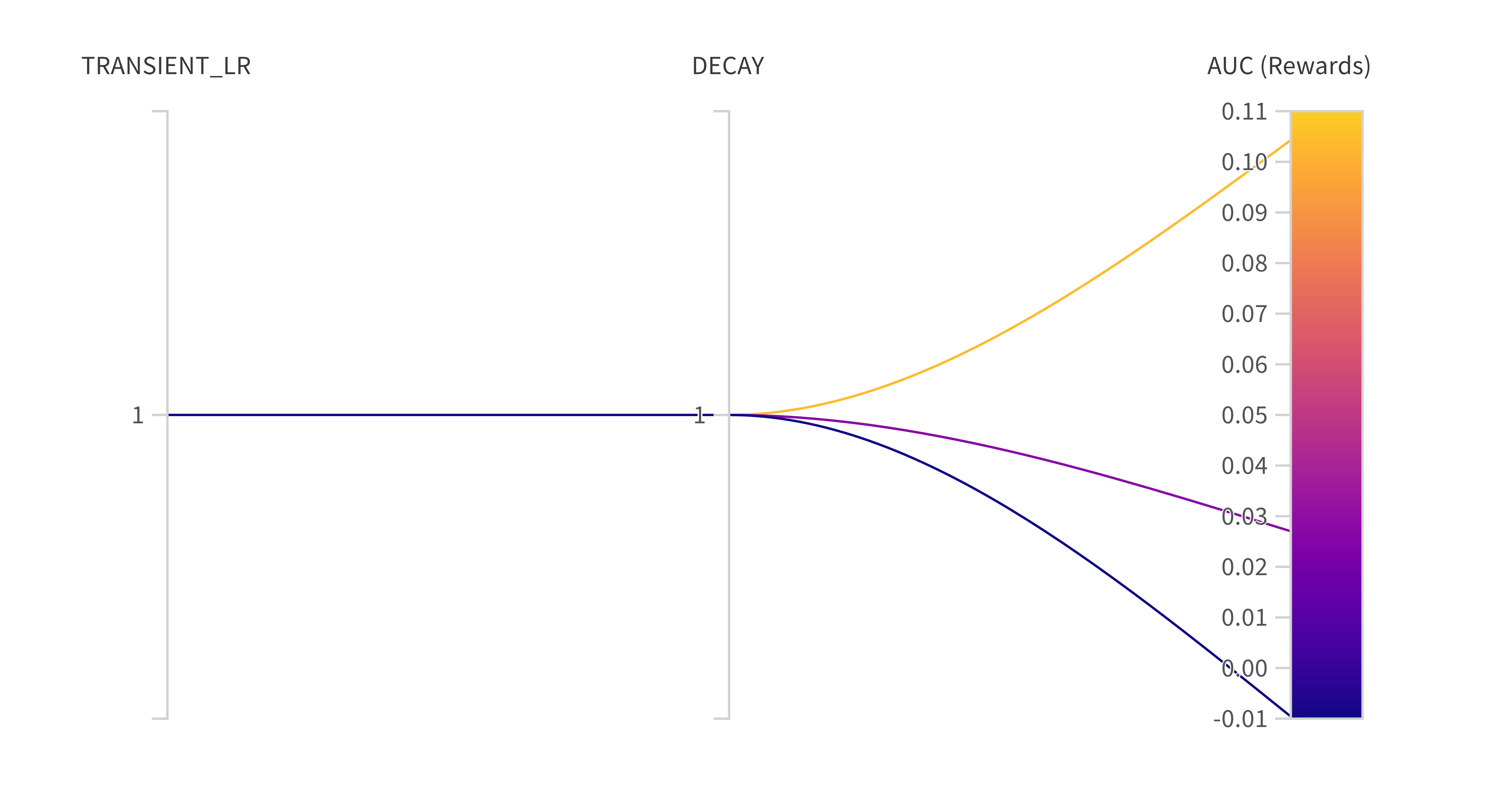}
        \caption{T-Only}
        \label{fig:img-t-only-HP}
    \end{subfigure}
    \caption{Hyperparameter tuning results.}
    \label{fig:img-hp-tuning-all}
\end{figure}

\begin{algorithm}[htb]
\caption{Deep PTQ for Image Task and Craftax-Classic}
\label{algo:ptq-craftax}
\begin{algorithmic}[1]
    \Require $\mathcal{M}$: transient memory; $\PMparams$: permanent parameters;
             \texttt{env}: environment; $T_s$: total timesteps; $\mathcal{B}$: buffer;
             $\PMlr$: permanent LR; $\TMlr$: transient LR; $\rho$: mixing weight;
             $\epsilon$: exploration rate; $\gamma$: discount; $k$: PM update period;
             $\lambda$: TM decay
    \Ensure  Updated $\PMparams$, $\mathcal{M}$
    \State $s \gets \texttt{env.reset()}$
    \State $\mathcal{M} \gets \Call{Put}{\mathcal{M}, s}$

    \For{$t \gets 1\ \mathbf{to}\ T_s$}
        \State \Comment{permanent values}
        \State $\PMcontrol(s) \gets \Call{GetPermanent}{s, \PMparams}$

        \State \Comment{transient values}
        \State $\TMcontrol(s) \gets \Call{Get}{\mathcal{M}, s, \rho}$ 

        \State \Comment{Compute $Q$ for current state}
        \State $\QPT(s) \gets \PMcontrol(s) + \TMcontrol(s)$

        \State \Comment{Select action (epsilon-greedy over $Q$)}
        \State $a \gets \Call{EpsilonGreedy}{\QPT(s), \epsilon}$

        \State \Comment{Step environment}
        \State $(s', r) \gets \texttt{env.step}(a)$
        \State $\mathcal{M} \gets \Call{Put}{\mathcal{M}, s'}$

        \State $\mathcal{B} \gets \mathcal{B} \cup \{(s, a, \PMcontrol(s))\}$

        \State \Comment{Evaluate $Q$ for next state}
        \State $\PMcontrol(s') \gets \Call{GetPermanent}{s', \PMparams}$
        \State $\TMcontrol(s') \gets \Call{Get}{\mathcal{M}, s', \rho}$
        \State $\QPT(s') \gets \PMcontrol(s') + \TMcontrol(s')$

        \State \Comment{TD error and transient update}
        \State $\delta \gets r + \gamma \max_{a'} \QPT(s', a') - \QPT(s, a)$
        \State $\mathcal{M} \gets \Call{UpdateTDError}{\mathcal{M}, s, a, \rho, \delta, \TMlr}$

        \State \Comment{Periodic permanent update and TM decay}
        \If{$\mathrm{mod}(t, k) = 0$}
            \State $\PMparams \gets \Call{UpdatePM}{\mathcal{M}, \mathcal{B}, \PMlr}$
            \State $\mathcal{B} \gets \{\}$
            \State $\mathcal{M} \gets \Call{DecayValues}{\mathcal{M}, \lambda}$
        \EndIf

        \State $s \gets s'$
    \EndFor
    \State \Return $\PMparams$, $\mathcal{M}$
\end{algorithmic}
\end{algorithm}

\textbf{NEC.} This is a JAX-compatible adaptation of Neural Episodic Control (NEC). It leverages a Differentiable Neural Dictionary (DND) to rapidly associate latent representations with action values. Because dynamic array structures and KD-trees are discouraged in compiled JAX code, the dynamically growing memory of the original NEC is replaced with a fixed-capacity table with a Least Recently Used (LRU) replacement policy.

The neural network architecture follows the PQN baseline (two 1024-unit hidden layers, Layer Normalization, ReLU) but is modified to output both the feature embeddings (the activations of the second hidden layer) and Q-values.

DND is a static table with a fixed capacity of 1024 slots (same capacity as ours). Each slot stores a 1024-dimensional latent embedding, a vector of corresponding Q-values, and an age/clock integer used for tracking the oldest entries. Q-values are estimated by querying the DND. The DND retrieves the top $k=16$ nearest neighbours based on cosine similarity between the current latent features and stored keys. The final Q-value is a weighted sum of these neighbours.

The DND is updated online after each environment step (similar to the transient estimates). If the cosine similarity between the current state's features and the closest existing key exceeds a threshold ($0.85$), the stored Q-value is updated toward the 1-step TD error using a fast learning rate of $0.3$. If the state is sufficiently novel (similarity $\le 0.85$), a new key-value pair is written into the dictionary, overwriting the oldest entry based on age.

For the main results presented in the paper, we used the same hyperparameters for the base neural network to maintain a consistent standard for comparison. We conducted an expanded hyperparameter search to tune the base neural network's learning rate and observed an improved performance. However, this was still substantially lower than our approach.

\textbf{CompoNet.} CompoNet is a recent compositional architecture designed for CRL. It addresses catastrophic forgetting by freezing the recently learnt policy and combining it with a newer module using an attention mechanism as new tasks are encountered. While effective, this approach requires explicit knowledge of task boundaries. Also, it is memory-intensive, as the network structure is continuously expanded with new heads whenever a new task is presented. This also limited its use case for continuously changing environments.

\textbf{Reset (Final).} Reset (Final) re-initializes only the final layer of the Q-network when the task changes. This variant, inspired by the primacy bias paper, aims to unlearn outdated action-value estimates while preserving representations learned in lower layers. This approach allows the network to rapidly adapt without discarding useful latent features.

\textbf{Reset (Full).} Reset (Full) completely discards the past learnings at each task boundary, forcing the network to learn entirely from scratch for every new task. This baseline emphasizes high neural network plasticity.

\subsubsection{Craftax Task}
\label{app-sec:craftax-details}

For the baselines and the permanent network in all variants of PTQ-learning, we used hyperparameters from the PQN repository, which are consistent with the results published in the PQN paper. We only tuned minibatches for the PQN baseline because it determined the number of steps the permanent network would take. All the baselines also reuse the PQN base parameters.

Transient parameters were determined by searching over a range of values. Given the large hyperparameter space, we used coordinate-wise search: we fixed all hyperparameters except one to a reasonable baseline and performed a grid search over that single hyperparameter. This process was then repeated iteratively for each hyperparameter, fixing the newly tuned value before moving to the next. For each configuration, we ran 150M steps across 1024 environments in parallel, using AUC as the selection metric. The final results were reported for 250M steps. We used the same HPs for the online craftax experiment (the minibatch was reduced to 8 to match the smaller batch size). A similar procedure was carried out for all the baselines we used.

\textbf{Wall-clock time.}  For the 2-stream experiment, the wall-clock time of our approach is very similar to PQN (14-15h for PQN, 16-18h for our approach). In the 1024-stream experiment, our approach is twice as slow as PQN, since the transient component loops over samples to store them (retrieval is parallelized, therefore cheap). We expect the newer, simplified transient architecture to yield faster runtime.

\begin{table}[htb]
\centering
\begin{tabular}{ll}
\toprule
\textbf{Parameter} & \textbf{Value} \\
\midrule
NUM\_TABLES & 2 \\
TRANSIENT\_TABLE\_SIZE & 2048 \\
NUM\_SLOTS & 32 \\
NUM\_HASHES & 128 \\
CROP\_SIZE & 7 \\
$\rho$ & 0.85 \\
TRANSIENT\_LR & 1.3 \\
DECAY & 0.95 \\
\bottomrule
\end{tabular}
\vspace{5pt}
\caption{PT-Ours HPs for Craftax.}
\label{tab:pqn-pt-hps-craftax}
\end{table}

\begin{table}[htb]
\centering
\begin{tabular}{ll}
\toprule
\textbf{Parameter} & \textbf{Value} \\
\midrule
NUM\_TABLES & 16 \\
TRANSIENT\_TABLE\_SIZE & 65536 \\
NUM\_SLOTS & 32 \\
NUM\_HASHES & 128 \\
CROP\_SIZE & 7 \\
TRANSIENT\_LR & 0.8 \\
DECAY & 0.85 \\
\bottomrule
\end{tabular}
\vspace{5pt}
\caption{PT-IHT HPs for Craftax.}
\label{tab:pqn-pt-iht-hps-craftax}
\end{table}

\begin{table}[htb]
\centering
\begin{tabular}{ll}
\toprule
\textbf{Parameter} & \textbf{Value} \\
\midrule
ALG\_NAME & \texttt{PQN} \\
TOTAL\_TIMESTEPS & $5\times 10^8$ \\
TOTAL\_TIMESTEPS\_DECAY & $5\times 10^8$ \\
NUM\_ENVS & 1024 \\
NUM\_STEPS & 32 \\
EPS\_START & 0.1 \\
EPS\_FINISH & 0.005 \\
EPS\_DECAY & 0.2 \\
NUM\_MINIBATCHES & 1 for PQN(1), 32 for PQN(32)\\
NUM\_EPOCHS & 1 \\
NORM\_INPUT & True \\
NORM\_TYPE & \texttt{layer\_norm} \\
HIDDEN\_SIZE & 1024 \\
NUM\_LAYERS & 4 \\
LR & 0.0001 \\
MAX\_GRAD\_NORM & 1.0 \\
LR\_LINEAR\_DECAY & True \\
REW\_SCALE & 1.0 \\
GAMMA & 0.99 \\
Q\_LAMBDA & False \\
LAMBDA & 0 \\
\midrule
\multicolumn{2}{c}{\textbf{Environment}} \\
\midrule
ENV\_NAME & \texttt{Craftax-Classic-Symbolic-v1} \\
USE\_OPTIMISTIC\_RESETS & True \\
OPTIMISTIC\_RESET\_RATIO & 16 \\
LOG\_ACHIEVEMENTS & True \\
\bottomrule
\end{tabular}
\vspace{5pt}
\caption{PQN HPs for Craftax}
\label{tab:pqn-hps}
\end{table}

For Neural Episodic Control (NEC), we evaluated seven configurations and determined the optimal settings to be a Differentiable Neural Dictionary (DND) learning rate of 0.1, a cosine similarity threshold of 0.9, and a base neural network learning rate of 1e-4. For Elastic Weight Consolidation (EWC), a search over four different values identified an optimal penalty coefficient ($\lambda$) of 3.0. Similarly, the Continual Backprop (CB) baseline was tuned across four distinct values, achieving peak performance with a reset probability of 2e-4. For the Transient-Only (T-Only) ablation, we maintained consistency by reusing the optimal table hyperparameters from our approach. Finally, the PT-NN baseline was configured with a transient network learning rate of 0.0001 and a decay factor of 0.99.

We now present per-achievement performance curves for all algorithms, including parameter-ablation curves for our approach.

\begin{figure}[p]
    \centering
    \small
    \setlength{\tabcolsep}{4pt}
    \begin{tabular}{@{} c c c c @{}} 
    & \textbf{Returns} & \textbf{Score} & \textbf{Episode Length} \\[8pt]

    \rotatebox[origin=c]{90}{\textbf{MB}} &
    \raisebox{-0.5\height}{\includegraphics[width=0.31\textwidth, height=0.13\textheight, keepaspectratio]{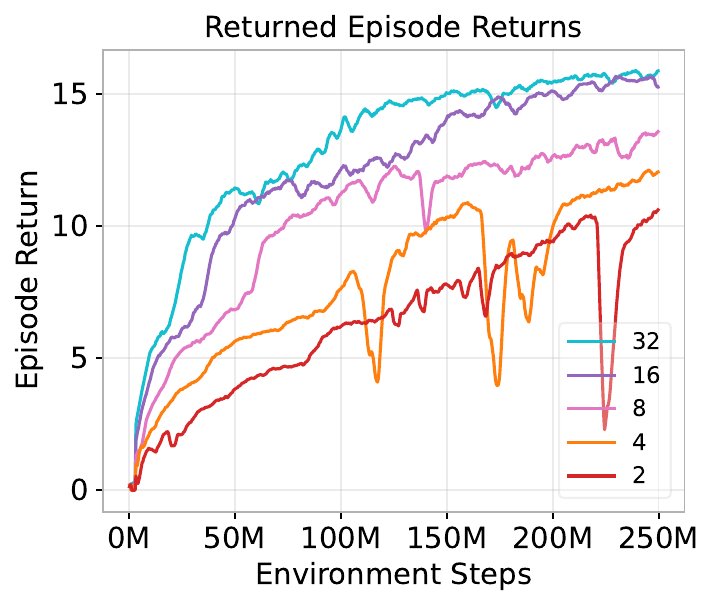}} &
    \raisebox{-0.5\height}{\includegraphics[width=0.31\textwidth, height=0.13\textheight, keepaspectratio]{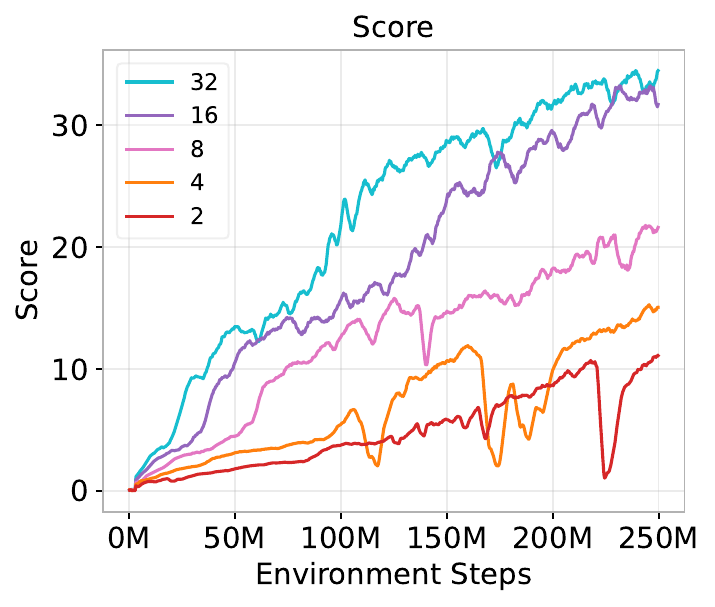}} &
    \raisebox{-0.5\height}{\includegraphics[width=0.31\textwidth, height=0.13\textheight, keepaspectratio]{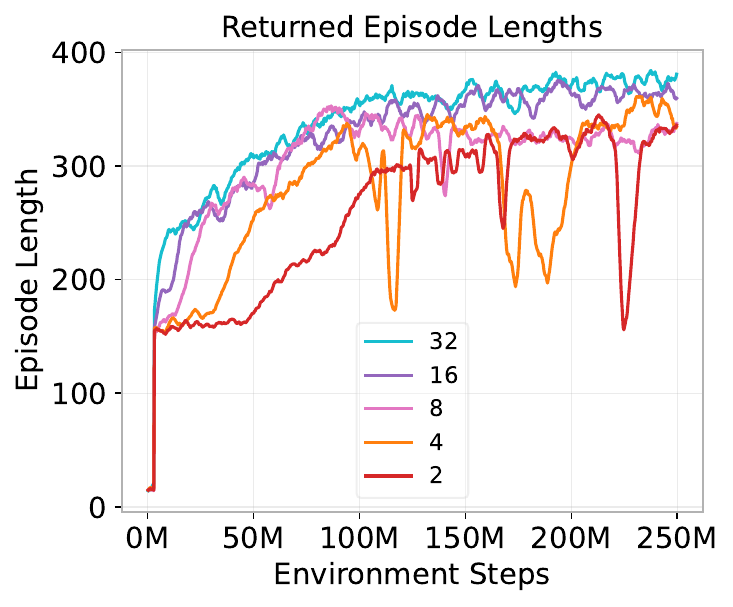}} \\[12pt]

    \rotatebox[origin=c]{90}{\textbf{Rho}} &
    \raisebox{-0.5\height}{\includegraphics[width=0.31\textwidth, height=0.13\textheight, keepaspectratio]{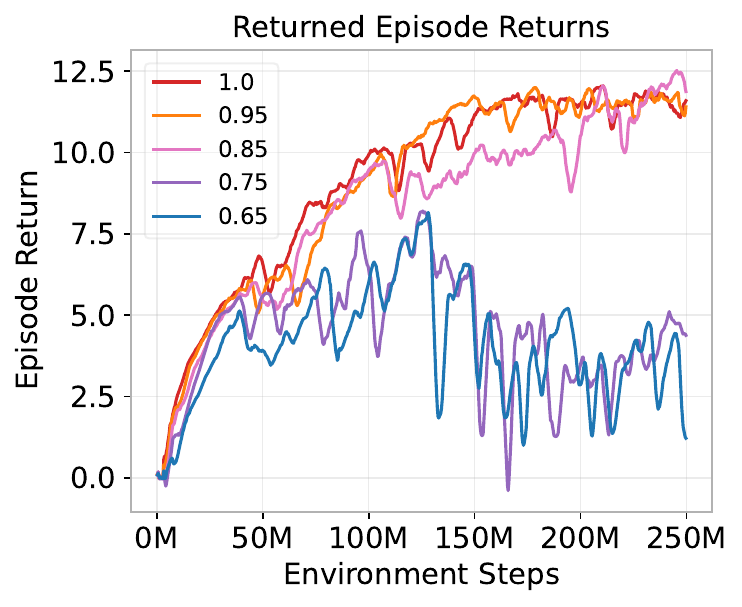}} &
    \raisebox{-0.5\height}{\includegraphics[width=0.31\textwidth, height=0.13\textheight, keepaspectratio]{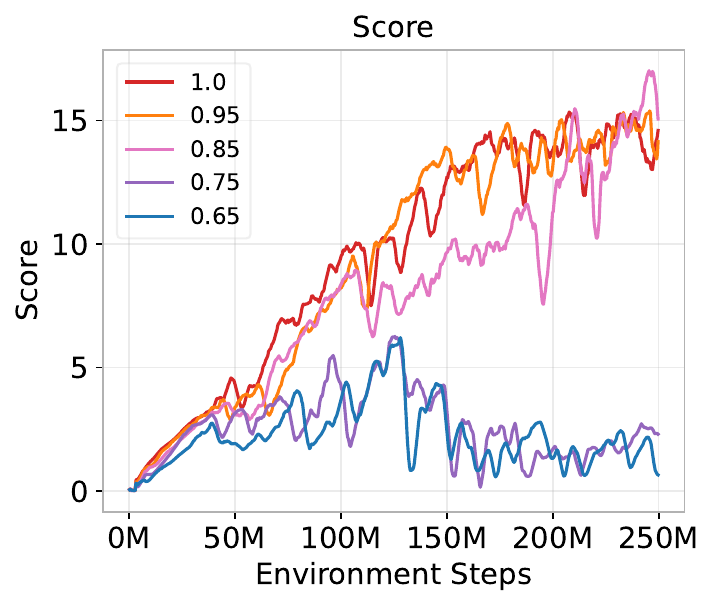}} &
    \raisebox{-0.5\height}{\includegraphics[width=0.31\textwidth, height=0.13\textheight, keepaspectratio]{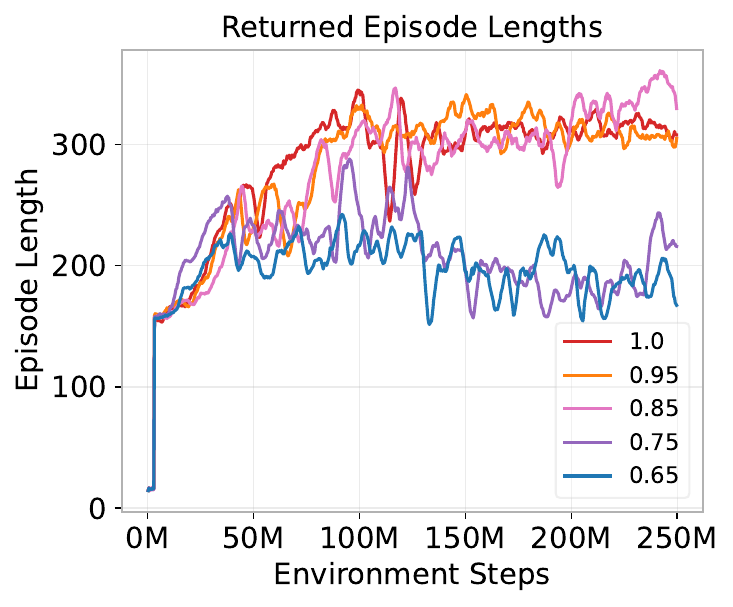}} \\[12pt]

    \rotatebox[origin=c]{90}{\textbf{Tables}} &
    \raisebox{-0.5\height}{\includegraphics[width=0.31\textwidth, height=0.13\textheight, keepaspectratio]{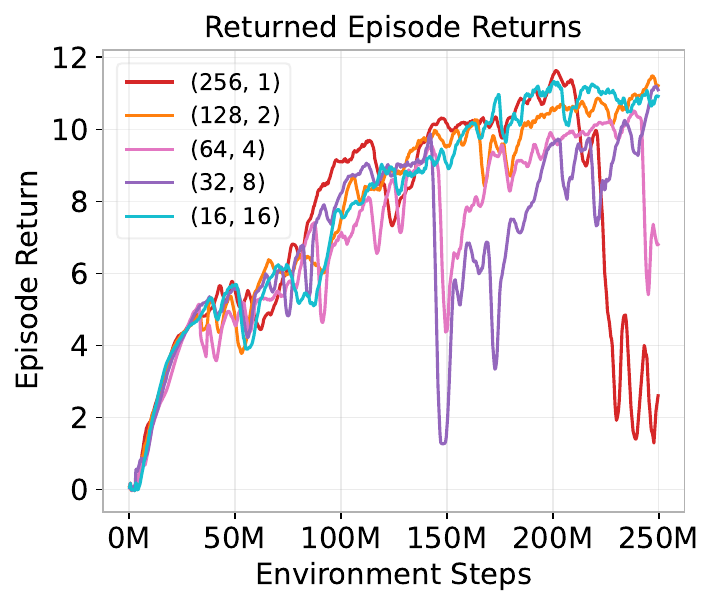}} &
    \raisebox{-0.5\height}{\includegraphics[width=0.31\textwidth, height=0.13\textheight, keepaspectratio]{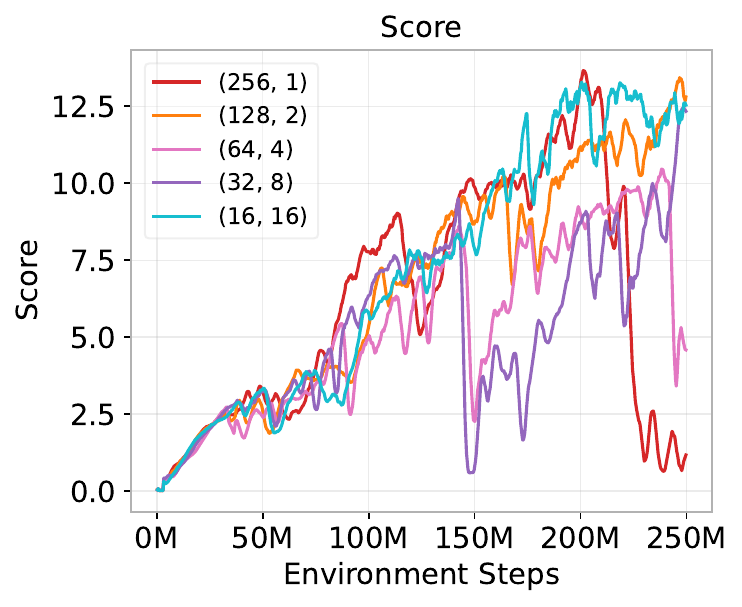}} &
    \raisebox{-0.5\height}{\includegraphics[width=0.31\textwidth, height=0.13\textheight, keepaspectratio]{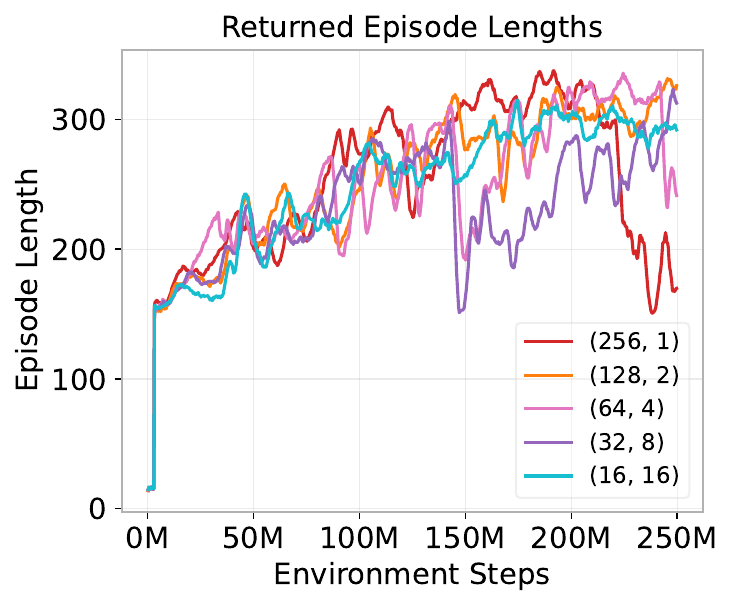}} \\[12pt]

    \rotatebox[origin=c]{90}{\textbf{Decay}} &
    \raisebox{-0.5\height}{\includegraphics[width=0.31\textwidth, height=0.13\textheight, keepaspectratio]{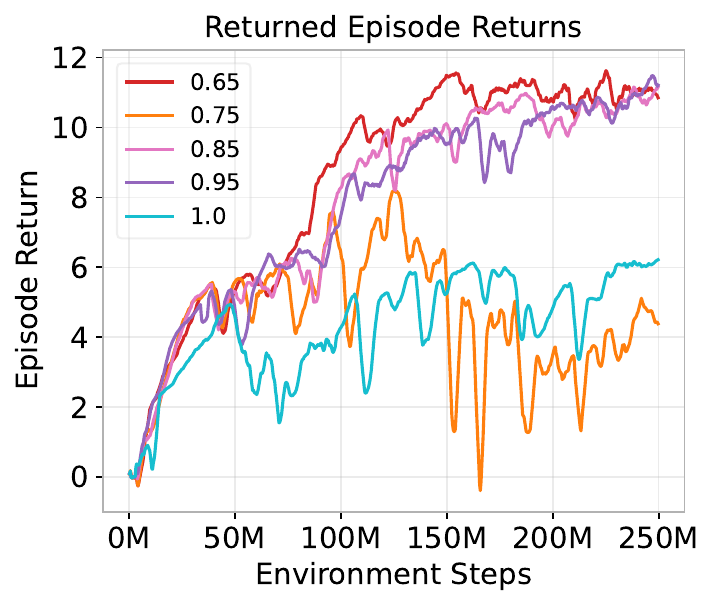}} &
    \raisebox{-0.5\height}{\includegraphics[width=0.31\textwidth, height=0.13\textheight, keepaspectratio]{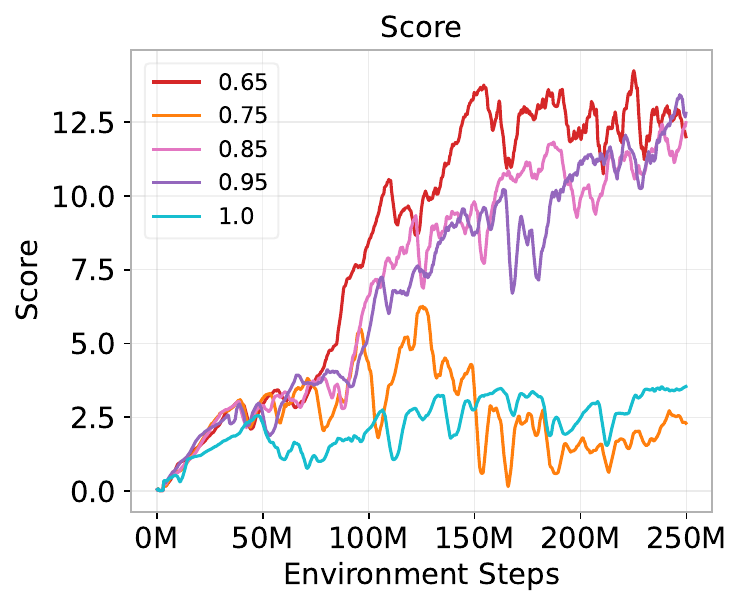}} &
    \raisebox{-0.5\height}{\includegraphics[width=0.31\textwidth, height=0.13\textheight, keepaspectratio]{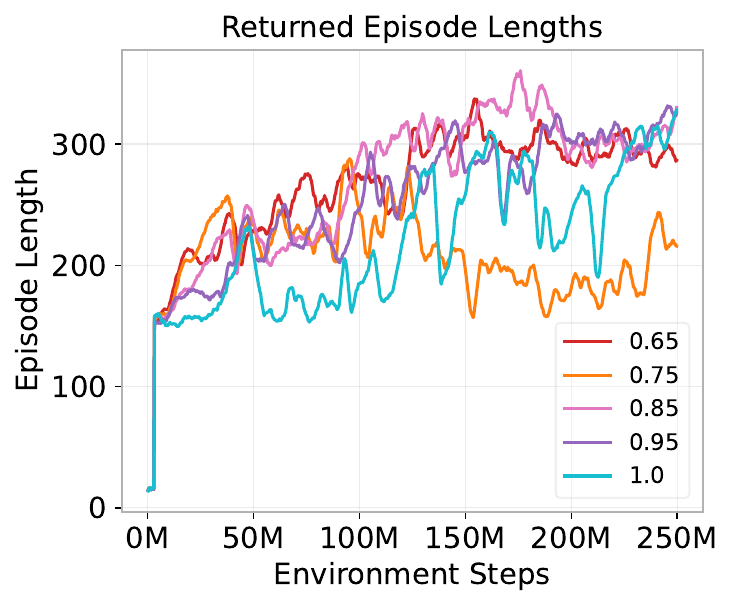}} \\[12pt]

    \rotatebox[origin=c]{90}{\textbf{LR (v2)}} &
    \raisebox{-0.5\height}{\includegraphics[width=0.31\textwidth, height=0.13\textheight, keepaspectratio]{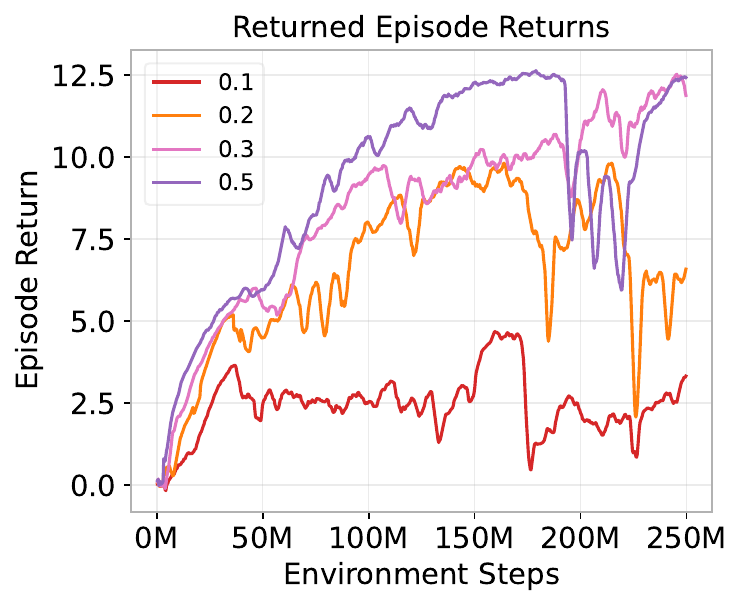}} &
    \raisebox{-0.5\height}{\includegraphics[width=0.31\textwidth, height=0.13\textheight, keepaspectratio]{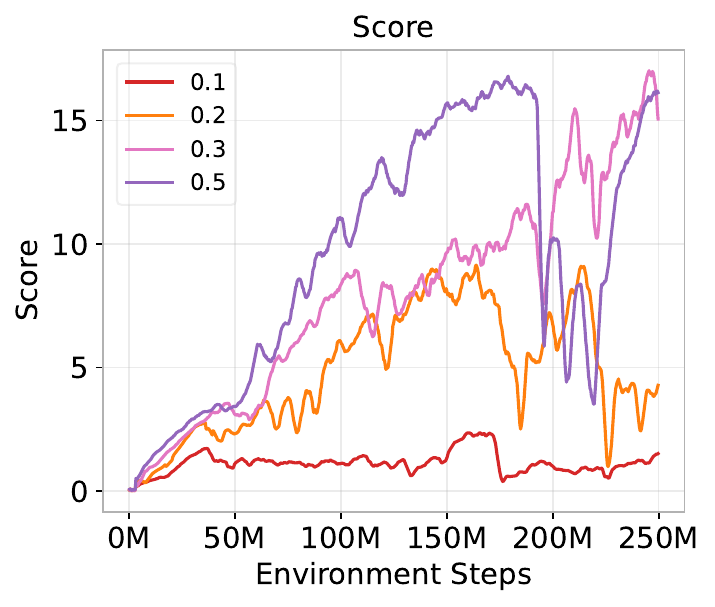}} &
    \raisebox{-0.5\height}{\includegraphics[width=0.31\textwidth, height=0.13\textheight, keepaspectratio]{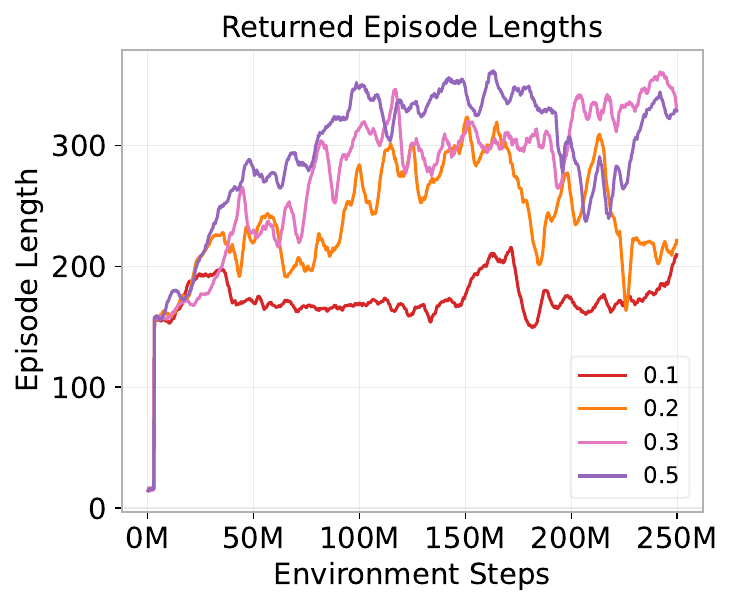}} \\[12pt]

    \rotatebox[origin=c]{90}{\textbf{LR (v1)}} &
    \raisebox{-0.5\height}{\includegraphics[width=0.31\textwidth, height=0.13\textheight, keepaspectratio]{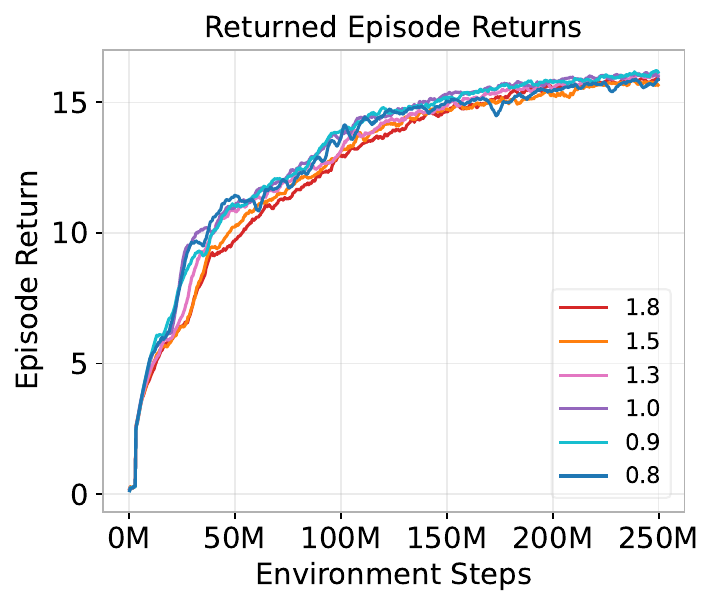}} &
    \raisebox{-0.5\height}{\includegraphics[width=0.31\textwidth, height=0.13\textheight, keepaspectratio]{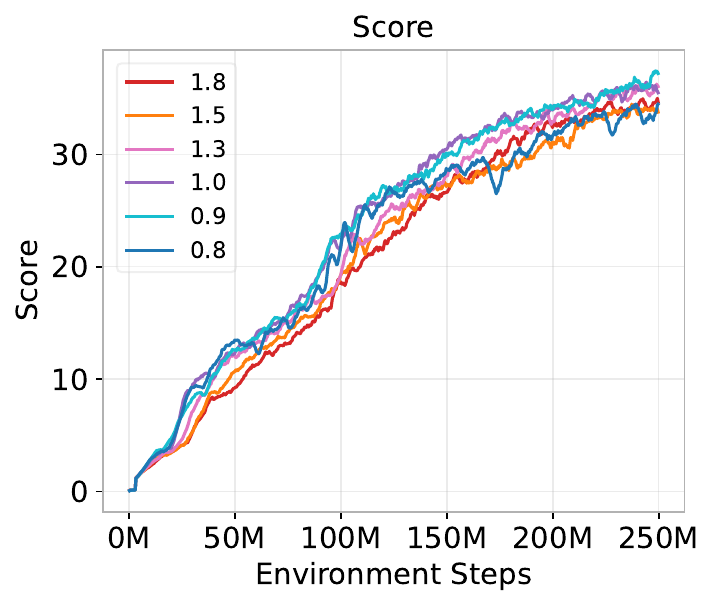}} &
    \raisebox{-0.5\height}{\includegraphics[width=0.31\textwidth, height=0.13\textheight, keepaspectratio]{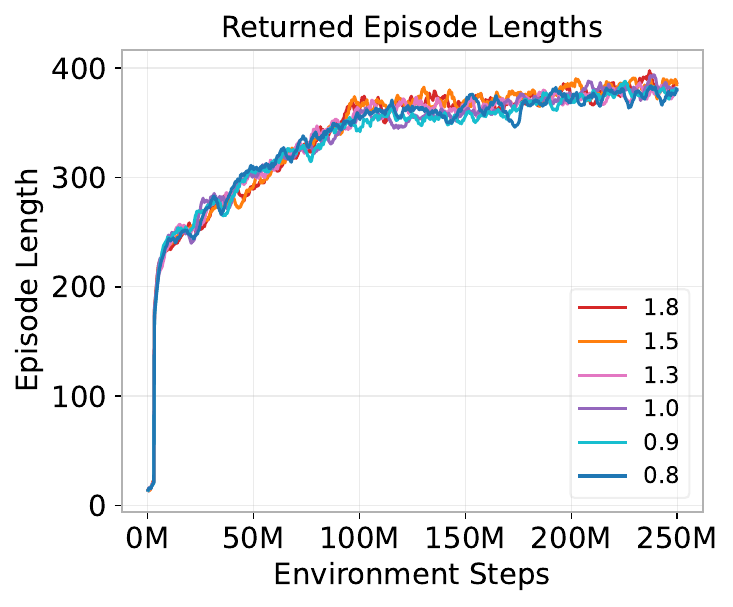}} \\

    \end{tabular}

    \vspace{8pt}
    \caption{\textbf{Parameter Ablation Studies.} Rows represent different parameters tested: Minibatch updates (MB), $\rho$, table configuration, decay parameter, and transient learning rates (v1 and v2). Columns represent standard metrics: Episode Returns, Score, and Episode Lengths.}
    \label{app-fig:full-ablation-grid}
\end{figure}

\begin{figure}[H]
  \centering
  \begin{subfigure}[c]{0.24\textwidth}
    \centering
    \includegraphics[width=0.99\columnwidth]{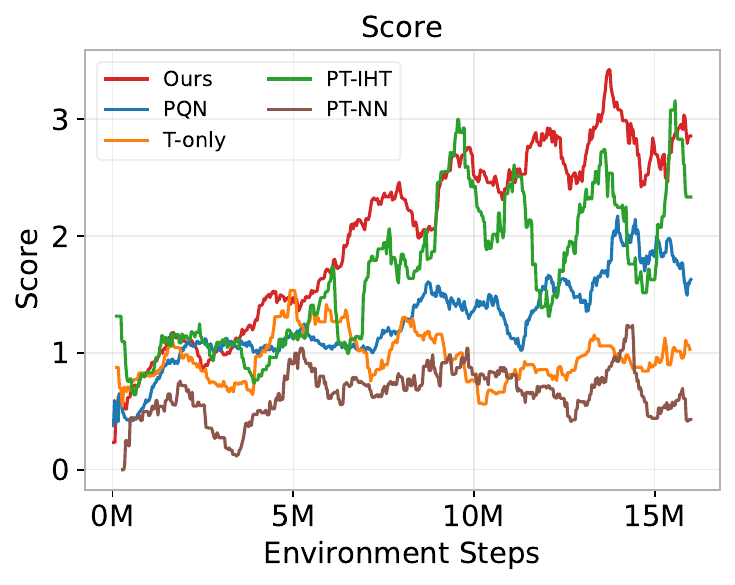}
    \caption{Score.}
    \label{fig:online-score}
  \end{subfigure}\hfill
  \begin{subfigure}[c]{0.24\textwidth}
    \centering
    \includegraphics[width=0.99\columnwidth]{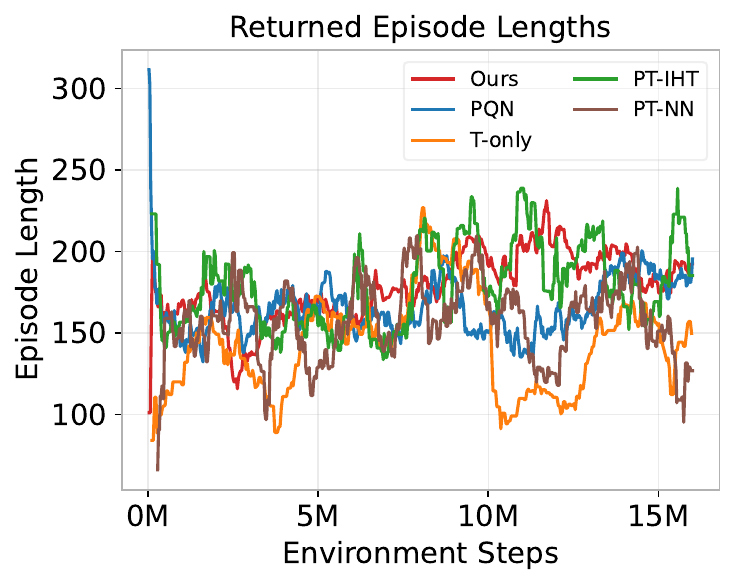}
    \caption{Episode Lengths.}
    \label{fig:online-episode-length}
  \end{subfigure}\hfill
  \begin{subfigure}[c]{0.24\textwidth}
    \centering
    \includegraphics[width=0.99\columnwidth]{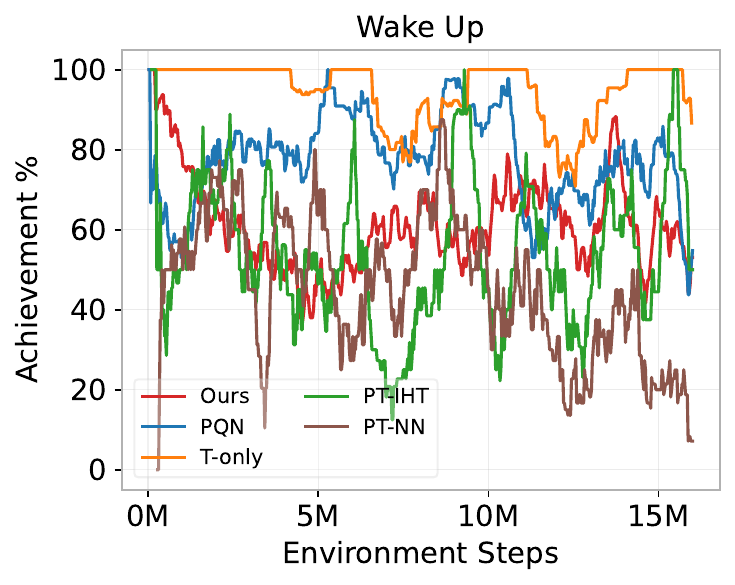}
    \caption{Wake Up.}
    \label{fig:online-wake-up}
  \end{subfigure}\hfill
  \begin{subfigure}[c]{0.24\textwidth}
    \centering
    \includegraphics[width=0.99\columnwidth]{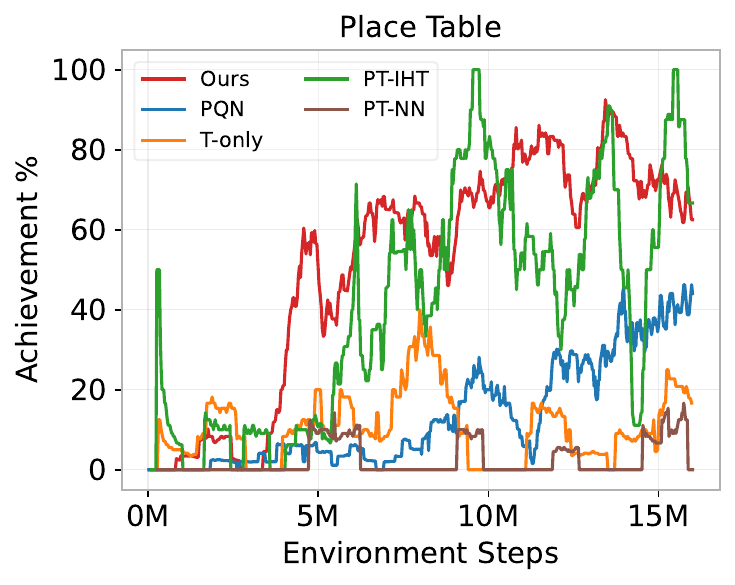}
    \caption{Place Table.}
    \label{fig:online-place-table}
  \end{subfigure}

  \begin{subfigure}[c]{0.24\textwidth}
    \centering
    \includegraphics[width=0.99\columnwidth]{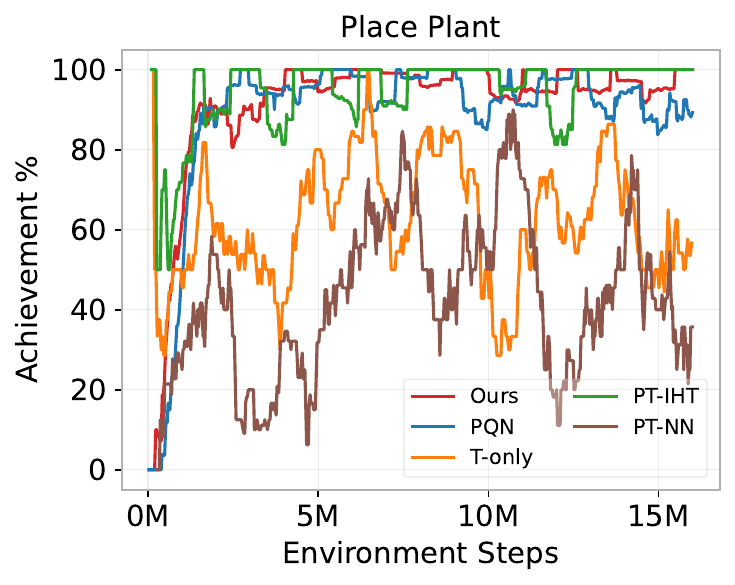}
    \caption{Place Plant.}
    \label{fig:online-place-plant}
  \end{subfigure}\hfill
  \begin{subfigure}[c]{0.24\textwidth}
    \centering
    \includegraphics[width=0.99\columnwidth]{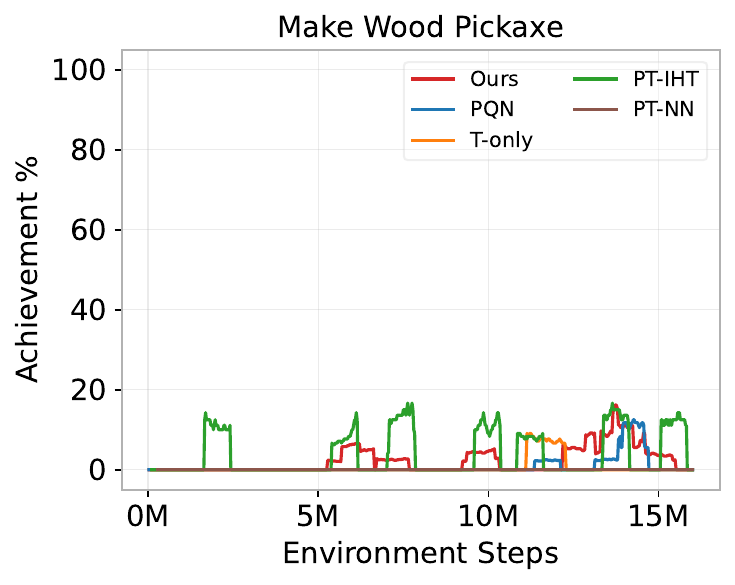}
    \caption{Make Wood Pickaxe.}
    \label{fig:online-make-wood-pickaxe}
  \end{subfigure}\hfill
  \begin{subfigure}[c]{0.24\textwidth}
    \centering
    \includegraphics[width=0.99\columnwidth]{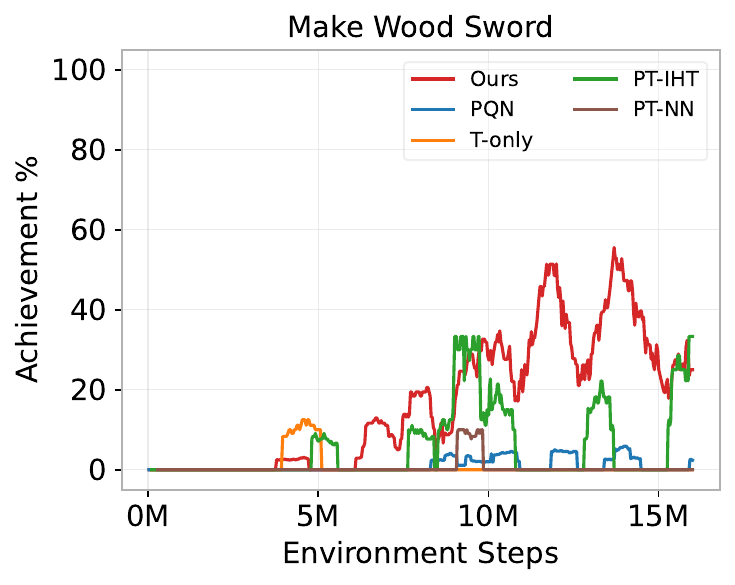}
    \caption{Make Wood Sword.}
    \label{fig:online-make-wood-sword}
  \end{subfigure}\hfill
  \begin{subfigure}[c]{0.24\textwidth}
    \centering
    \includegraphics[width=0.99\columnwidth]{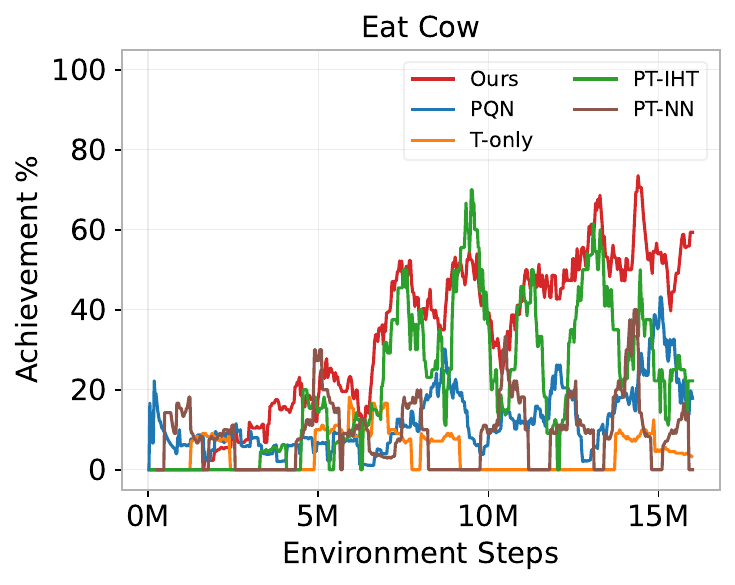}
    \caption{Eat Cow.}
    \label{fig:online-eat-cow}
  \end{subfigure}

    \begin{subfigure}[c]{0.24\textwidth}
    \centering
    \includegraphics[width=0.99\columnwidth]{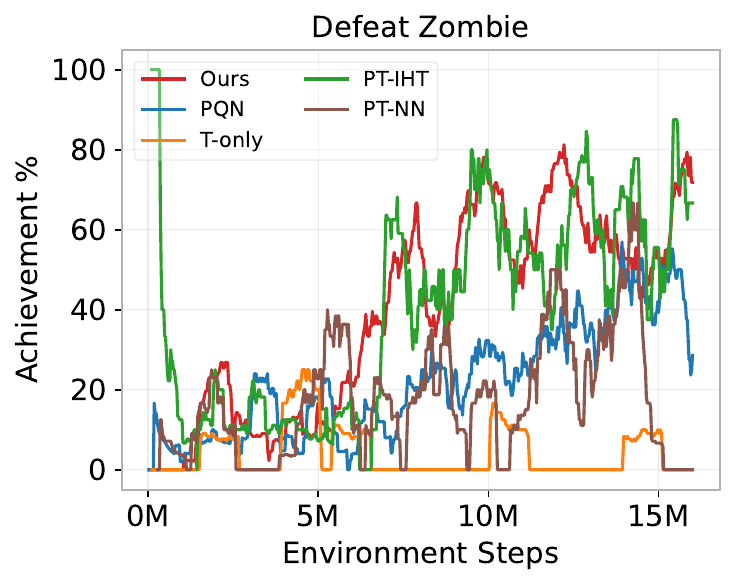}
    \caption{Defeat Zombie.}
    \label{fig:online-defeat-zombie}
  \end{subfigure}\hfill
  \begin{subfigure}[c]{0.24\textwidth}
    \centering
    \includegraphics[width=0.99\columnwidth]{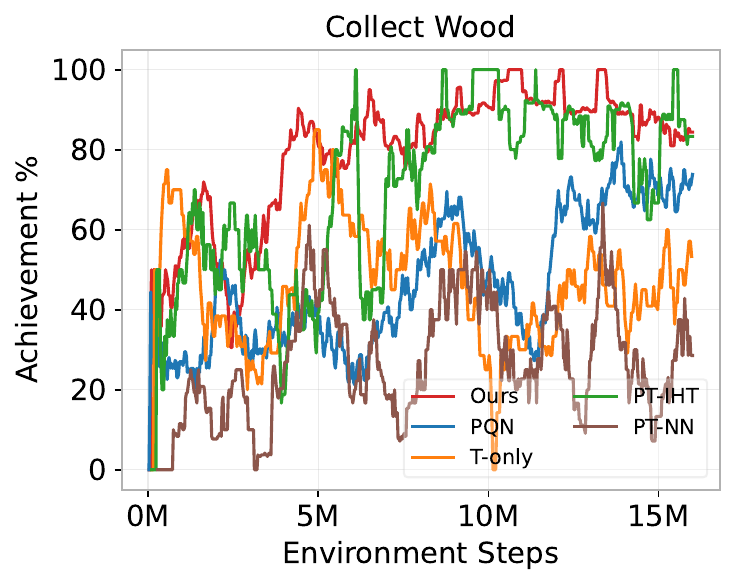}
    \caption{Collect Wood.}
    \label{fig:online-collect-wood}
  \end{subfigure}\hfill
  \begin{subfigure}[c]{0.24\textwidth}
    \centering
    \includegraphics[width=0.99\columnwidth]{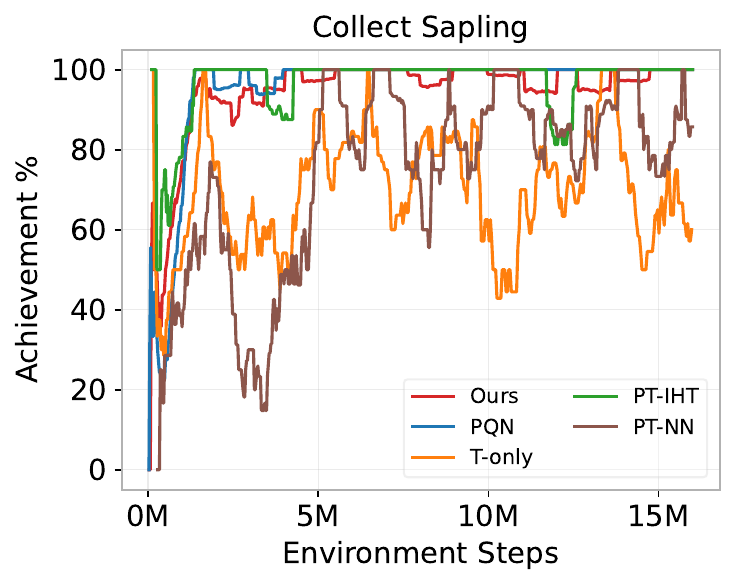}
    \caption{Collect Sapling.}
    \label{fig:online-collect-sapling}
  \end{subfigure}\hfill
  \begin{subfigure}[c]{0.24\textwidth}
    \centering
    \includegraphics[width=0.99\columnwidth]{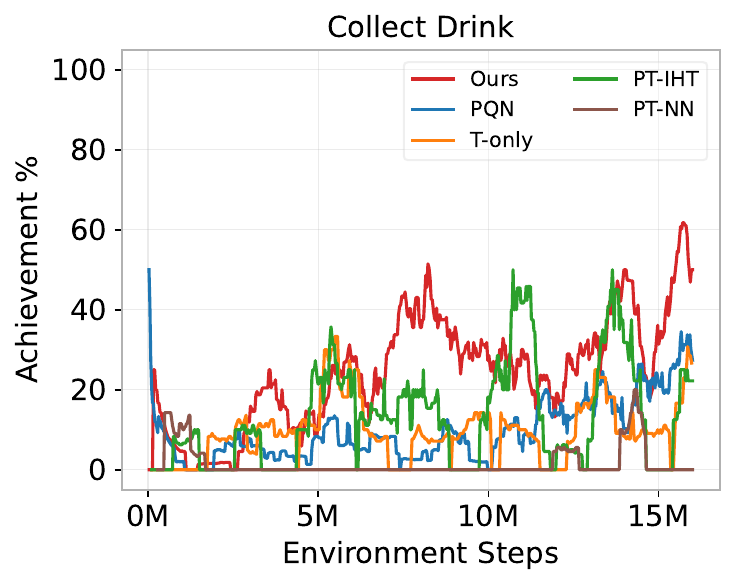}
    \caption{Collect Drink.}
    \label{fig:online-collect-drink}
  \end{subfigure}

    \begin{subfigure}[c]{0.24\textwidth}
    \centering
    \includegraphics[width=0.99\columnwidth]{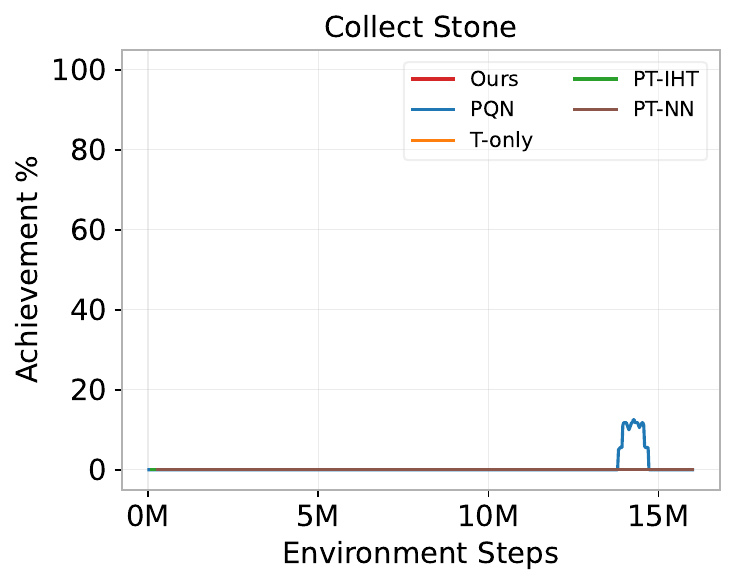}
    \caption{Collect Stone.}
    \label{fig:online-collect-stone}
  \end{subfigure}\hfill
  \begin{subfigure}[c]{0.24\textwidth}
    \centering
    \includegraphics[width=0.99\columnwidth]{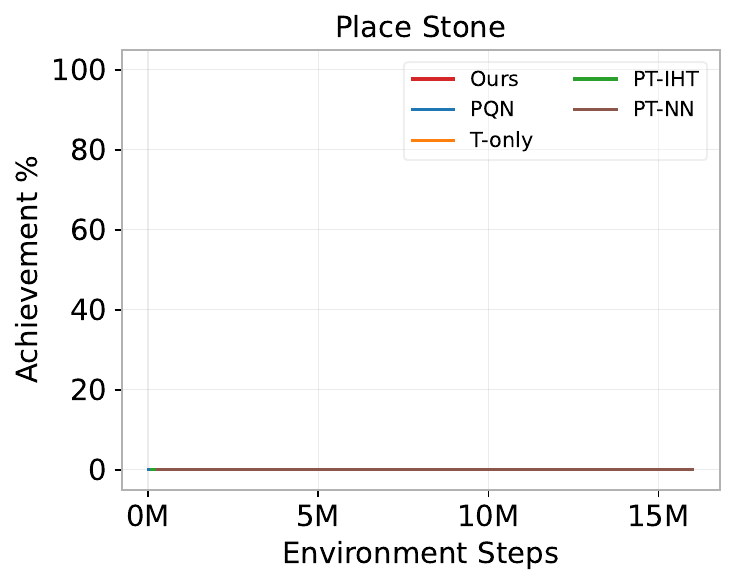}
    \caption{Place Stone.}
    \label{fig:online-place-stone}
  \end{subfigure}\hfill
  \begin{subfigure}[c]{0.24\textwidth}
    \centering
    \includegraphics[width=0.99\columnwidth]{figures/online_plots/Defeat_Zombie.pdf}
    \caption{Defeat Zombie.}
    \label{fig:online-defeat-skeleton}
  \end{subfigure}\hfill
  \begin{subfigure}[c]{0.24\textwidth}
    \centering
    \includegraphics[width=0.99\columnwidth]{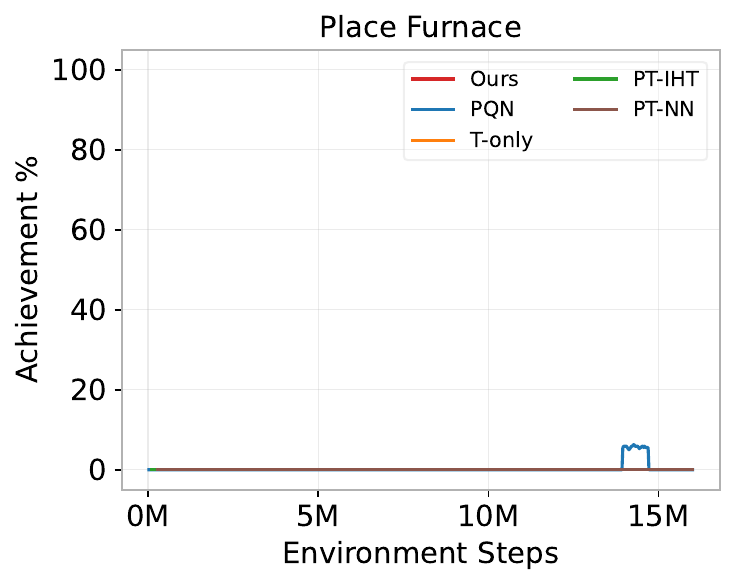}
    \caption{Place Furnace.}
    \label{fig:online-place-furnace}
  \end{subfigure}

    \begin{subfigure}[c]{0.24\textwidth}
    \centering
    \includegraphics[width=0.99\columnwidth]{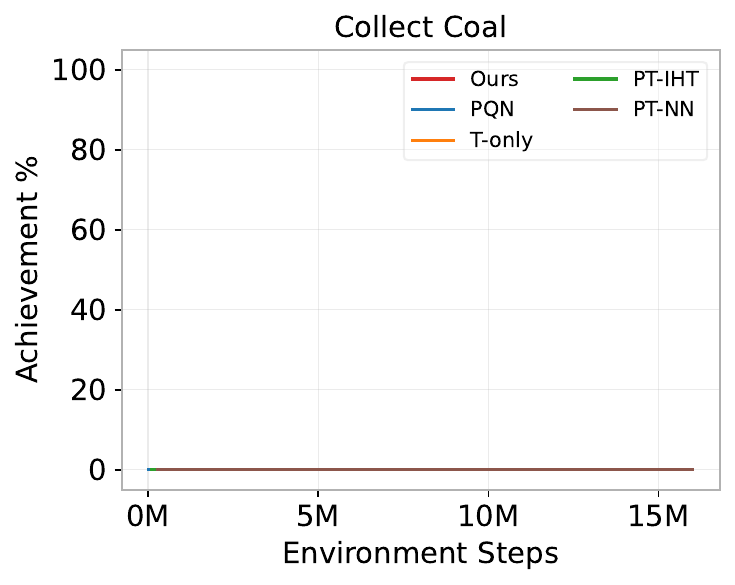}
    \caption{Collect Coal.}
    \label{fig:online-collect-coal}
  \end{subfigure}\hfill
  \begin{subfigure}[c]{0.24\textwidth}
    \centering
    \includegraphics[width=0.99\columnwidth]{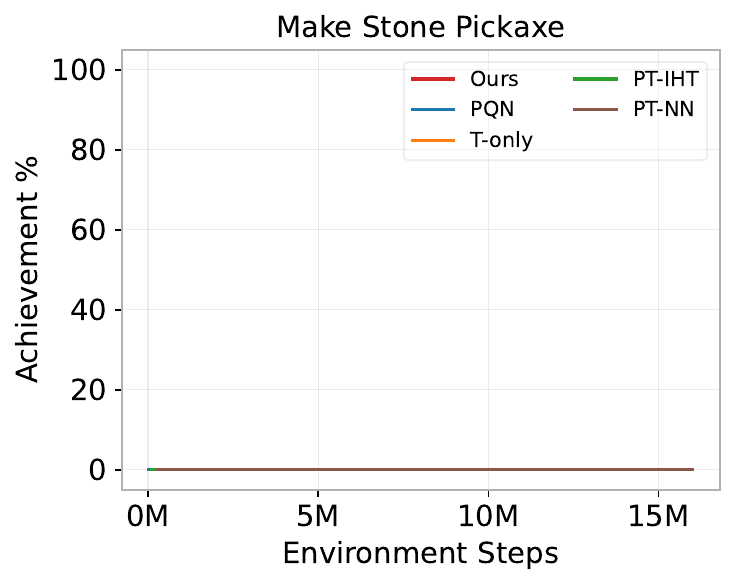}
    \caption{Make Stone Pickaxe.}
    \label{fig:online-make-stone-pickaxe}
  \end{subfigure}\hfill
  \begin{subfigure}[c]{0.24\textwidth}
    \centering
    \includegraphics[width=0.99\columnwidth]{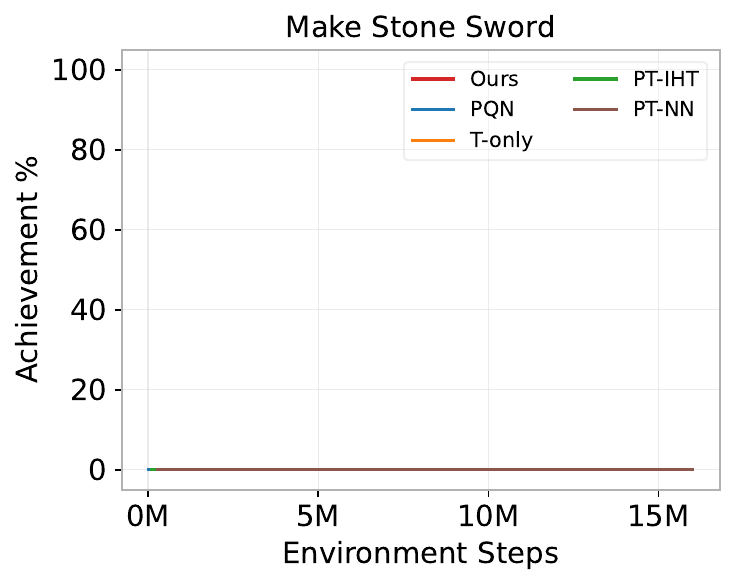}
    \caption{Make Stone Sword.}
    \label{fig:online-stone-sword}
  \end{subfigure}\hfill
  \begin{subfigure}[c]{0.24\textwidth}
    \centering
    \includegraphics[width=0.99\columnwidth]{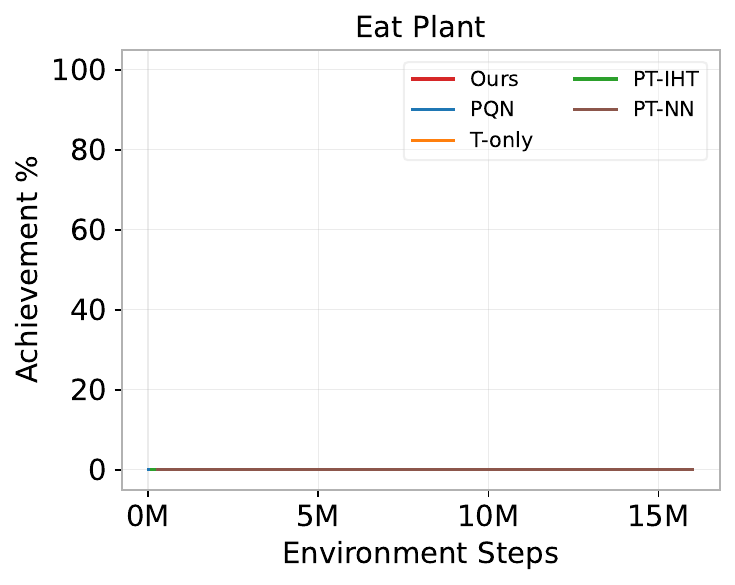}
    \caption{Eat Plant.}
    \label{fig:online-eat-plant}
  \end{subfigure}

    \begin{subfigure}[c]{0.24\textwidth}
    \centering
    \includegraphics[width=0.99\columnwidth]{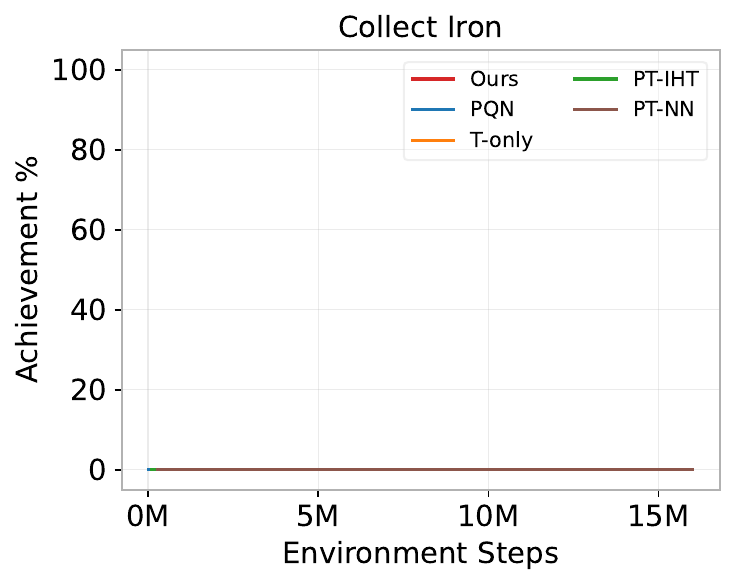}
    \caption{Collect Iron.}
    \label{fig:online-collect-iron}
  \end{subfigure}\hfill
  \begin{subfigure}[c]{0.24\textwidth}
    \centering
    \includegraphics[width=0.99\columnwidth]{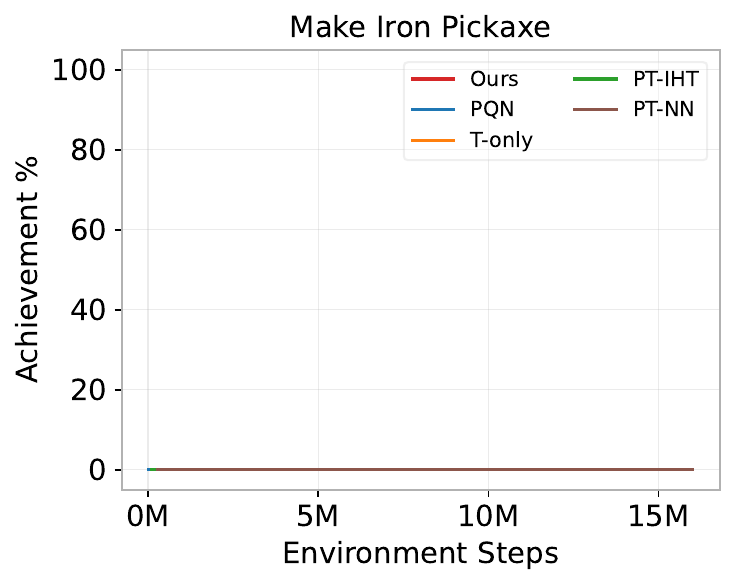}
    \caption{Make Iron Pickaxe.}
    \label{fig:online-make-iron-pickaxe}
  \end{subfigure}\hfill
  \begin{subfigure}[c]{0.24\textwidth}
    \centering
    \includegraphics[width=0.99\columnwidth]{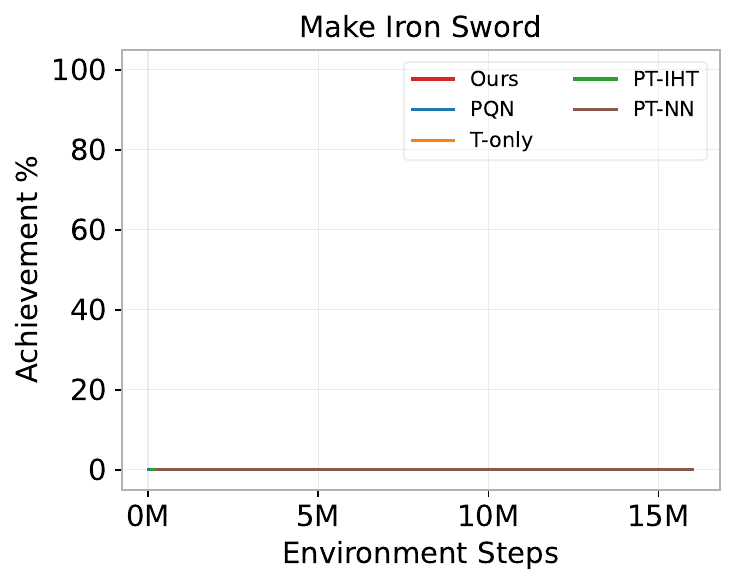}
    \caption{Make Iron Sword.}
    \label{fig:online-iron-sword}
  \end{subfigure}\hfill
  \begin{subfigure}[c]{0.24\textwidth}
    \centering
    \includegraphics[width=0.99\columnwidth]{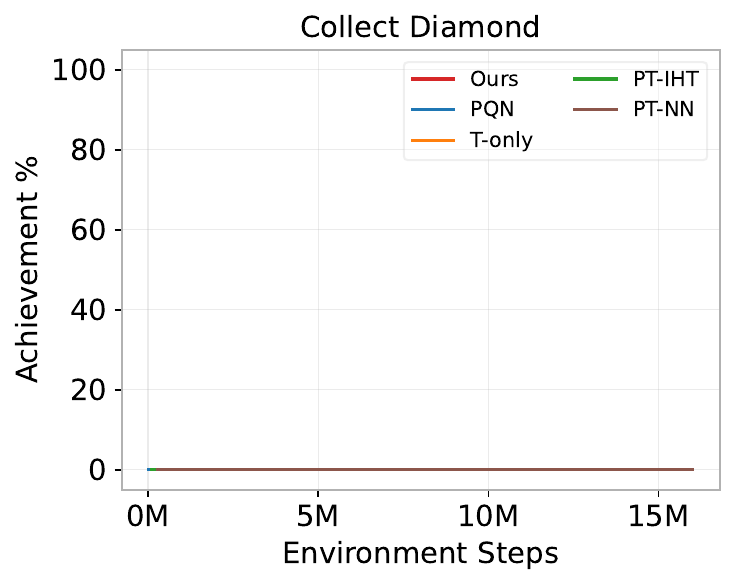}
    \caption{Collect Diamond.}
    \label{fig:online-collect-diamond}
  \end{subfigure}
  \caption{All achievements in the craftax online experiment.}
  \label{app-fig:online-achievements}
\end{figure}

\newpage
\section*{NeurIPS Paper Checklist}

\begin{enumerate}

\item {\bf Claims}
    \item[] Question: Do the main claims made in the abstract and introduction accurately reflect the paper's contributions and scope?
    \item[] Answer: \answerYes{} 
    \item[] Justification: Our claims are supported by illustrations in Sec.~\ref{sec:small-scale-experiments}, theoretical results in Sec.~\ref{sec:theory}, details of the non-parametric approximation in Sec.~\ref{sec:scaling-up-PT-memories}, and experiments on image tasks and Craftax in Sec.~\ref{sec:experiment}.

\item {\bf Limitations}
    \item[] Question: Does the paper discuss the limitations of the work performed by the authors?
    \item[] Answer: \answerYes{} 
    \item[] Justification: The limitations of our newly introduced non-parametric approximator are discussed in Sec.~\ref{sec:discussion}. Specifically, we address the use of a basic tokenization strategy in our image experiments and the large number of hyperparameters, both of which we highlight as avenues for future research.

\item {\bf Theory assumptions and proofs}
    \item[] Question: For each theoretical result, does the paper provide the full set of assumptions and a complete (and correct) proof?
    \item[] Answer: \answerYes{} 
    \item[] Justification: The standard theoretical assumptions are detailed in Sec.~\ref{sec:theory}, and the complete proofs are provided in Appendix~\ref{app-sec:proofs}.

    \item {\bf Experimental result reproducibility}
    \item[] Question: Does the paper fully disclose all the information needed to reproduce the main experimental results of the paper to the extent that it affects the main claims and/or conclusions of the paper (regardless of whether the code and data are provided or not)?
    \item[] Answer: \answerYes{} 
    \item[] Justification: Pseudocode for our algorithms is provided in Algorithms~\ref{algo:PT-linear-prediction}, \ref{algo:PT-linear-control}, \ref{algo:put-method}, \ref{algo:get-method}, \ref{algo:update-td-error-method}, and \ref{algo:ptq-craftax}. Complete experimental details, including baselines, hyperparameter tuning, and final values, are detailed in Appendices~\ref{app-sec-hp-tuning-small-scale}, \ref{app-sec:hp-strategy-transient}, and \ref{app-sec:large-scale-experiments}. Additionally, Appendix~\ref{app-sec:hp-strategy-transient} includes a practical guide for setting these hyperparameters.

\item {\bf Open access to data and code}
    \item[] Question: Does the paper provide open access to the data and code, with sufficient instructions to faithfully reproduce the main experimental results, as described in supplemental material?
    \item[] Answer: \answerYes{} 
    \item[] Justification: To ensure reproducibility, the complete codebase is open-sourced and is available at \url{https://github.com/NishanthVAnand/pt_representations_crl}.

\item {\bf Experimental setting/details}
    \item[] Question: Does the paper specify all the training and test details (e.g., data splits, hyperparameters, how they were chosen, type of optimizer) necessary to understand the results?
    \item[] Answer: \answerYes{} 
    \item[] Justification: Complete experimental details, including baselines, hyperparameter tuning, and final values, are detailed in Appendices~\ref{app-sec-hp-tuning-small-scale}, \ref{app-sec:hp-strategy-transient}, and \ref{app-sec:large-scale-experiments}. Additionally, Appendix~\ref{app-sec:hp-strategy-transient} includes a practical guide for setting these hyperparameters.

\item {\bf Experiment statistical significance}
    \item[] Question: Does the paper report error bars suitably and correctly defined or other appropriate information about the statistical significance of the experiments?
    \item[] Answer: \answerYes{} 
    \item[] Justification: All experiments are evaluated across multiple seeds, with confidence intervals reported directly in the corresponding plots. Full details are provided in Sec.~\ref{sec:small-scale-experiments} and \ref{sec:experiment}.

\item {\bf Experiments compute resources}
    \item[] Question: For each experiment, does the paper provide sufficient information on the computer resources (type of compute workers, memory, time of execution) needed to reproduce the experiments?
    \item[] Answer: \answerYes{} 
    \item[] Justification: While specific GPU and CPU configurations varied across cluster allocations, we provide an estimate of the wall-clock time for the large-scale Craftax experiments in Appendix~\ref{app-sec:craftax-details}.
    
\item {\bf Code of ethics}
    \item[] Question: Does the research conducted in the paper conform, in every respect, with the NeurIPS Code of Ethics \url{https://neurips.cc/public/EthicsGuidelines}?
    \item[] Answer: \answerYes{} 
    \item[] Justification: This work focuses on foundational algorithmic improvements and theoretical analysis. It does not involve human subjects, sensitive data, or applications that violate the NeurIPS Code of Ethics.

\item {\bf Broader impacts}
    \item[] Question: Does the paper discuss both potential positive societal impacts and negative societal impacts of the work performed?
    \item[] Answer: \answerNA{} 
    \item[] Justification: The scope of this paper is limited to foundational theoretical results and methodological evaluations in purely synthetic, simulated environments (e.g., Craftax); a discussion on societal impact was omitted.
    
\item {\bf Safeguards}
    \item[] Question: Does the paper describe safeguards that have been put in place for responsible release of data or models that have a high risk for misuse (e.g., pre-trained language models, image generators, or scraped datasets)?
    \item[] Answer: \answerNA{} 
    \item[] Justification: This paper focuses on foundational algorithmic theory and evaluations in simulated environments. It does not introduce, utilize, or release high-risk datasets or models that require safeguards against misuse.

\item {\bf Licenses for existing assets}
    \item[] Question: Are the creators or original owners of assets (e.g., code, data, models), used in the paper, properly credited and are the license and terms of use explicitly mentioned and properly respected?
    \item[] Answer: \answerYes{} 
    \item[] Justification: Our implementation builds upon the publicly available PureJaxQL library (\url{https://github.com/mttga/purejaxql}), which is released under the Apache-2.0 license. And, we use an environment from the Craftax library (\url{https://github.com/michaeltmatthews/craftax?tab=MIT-1-ov-file}), which is released under the MIT license. We explicitly acknowledge its creators and fully adhere to its terms of use.

\item {\bf New assets}
    \item[] Question: Are new assets introduced in the paper well documented and is the documentation provided alongside the assets?
    \item[] Answer: \answerNA{} 
    \item[] Justification: We do not introduce any new assets, such as datasets or environments, in this work. All experiments leverage existing, publicly available code and environments.

\item {\bf Crowdsourcing and research with human subjects}
    \item[] Question: For crowdsourcing experiments and research with human subjects, does the paper include the full text of instructions given to participants and screenshots, if applicable, as well as details about compensation (if any)? 
    \item[] Answer: \answerNA{} 
    \item[] Justification: We do not conduct any crowdsourcing experiments or research with human subjects in this work.

\item {\bf Institutional review board (IRB) approvals or equivalent for research with human subjects}
    \item[] Question: Does the paper describe potential risks incurred by study participants, whether such risks were disclosed to the subjects, and whether Institutional Review Board (IRB) approvals (or an equivalent approval/review based on the requirements of your country or institution) were obtained?
    \item[] Answer: \answerNA{} 
    \item[] Justification: Our work does not involve human subjects.

\item {\bf Declaration of LLM usage}
    \item[] Question: Does the paper describe the usage of LLMs if it is an important, original, or non-standard component of the core methods in this research? Note that if the LLM is used only for writing, editing, or formatting purposes and does \emph{not} impact the core methodology, scientific rigor, or originality of the research, declaration is not required.
    \item[] Answer: \answerNA{} 
    \item[] Justification: LLMs were not used as an important, original, or non-standard component of our core methodology. Their use was limited to standard assistance (e.g., code completion, grammar correction, and polishing), which does not require declaration in the paper.

\end{enumerate}

\end{document}